\documentclass[acmlarge]{acmart}

\usepackage[utf8]{inputenc}
\usepackage{graphicx}
\usepackage{booktabs}
\usepackage{amsmath}
\usepackage{array}
\usepackage{multirow}
\usepackage{bm}
\usepackage{subcaption}
\usepackage{url}
\usepackage{xcolor}
\usepackage{tikz}
\usepackage[edges]{forest}
\usepackage{bibunits}
\usepackage[capitalize]{cleveref}

\newcommand{\rev}[1]{#1}
\newcommand{\revnew}[1]{\textcolor{black}{#1}}

\definecolor{hidden-draw}{RGB}{20,68,106}
\definecolor{hidden-pink}{RGB}{255,245,247}

\DeclareUnicodeCharacter{2018}{'}
\DeclareUnicodeCharacter{2019}{'}
\DeclareUnicodeCharacter{2013}{--}
\DeclareUnicodeCharacter{2014}{---}
\DeclareUnicodeCharacter{2020}{\dag}
\DeclareUnicodeCharacter{2022}{\textbullet}
\DeclareUnicodeCharacter{2212}{-}

\renewcommand\footnotetextcopyrightpermission[1]{}
\defaultbibliographystyle{ACM-Reference-Format}
\defaultbibliography{mythesis}

\begin{document}

\title{Efficient Video Diffusion Models: Advancements and Challenges}

\author{Shitong Shao}
\affiliation{%
  \institution{The Hong Kong University of Science and Technology (Guangzhou)}
  \city{Guangzhou}
  \state{Guangdong}
  \country{China}
}
\email{sshao213@connect.hkust-gz.edu.cn}

\author{Lichen Bai}
\affiliation{%
  \institution{The Hong Kong University of Science and Technology (Guangzhou)}
  \city{Guangzhou}
  \state{Guangdong}
  \country{China}
}
\email{lbai292@connect.hkust-gz.edu.cn}

\author{Pengfei Wan}
\affiliation{%
  \institution{Kling Team, Kuaishou Technology}
   \city{Beijing}
  \country{China}
}
\email{wanpengfei@kuaishou.com}

\author{James Kwok}
\affiliation{%
  \institution{The Hong Kong University of Science and Technology}
  \country{Hong Kong}
}
\email{jamesk@cse.ust.hk}

\author{Zeke Xie}
\authornote{Corresponding author}
\affiliation{%
  \institution{The Hong Kong University of Science and Technology (Guangzhou)}
  \city{Guangzhou}
  \state{Guangdong}
  \country{China}
}
\email{zekexie@hkust-gz.edu.cn}

\renewcommand{\shortauthors}{Shitong Shao, James Kwok, Pengfei Wan, Zeke Xie}

\begin{abstract}
Video diffusion models have rapidly become the dominant paradigm for high-fidelity generative video synthesis, but their practical deployment remains constrained by severe inference costs. Compared with image generation, video synthesis compounds computation across spatial-temporal token growth and iterative denoising, making attention and memory traffic major bottlenecks in real-world settings. This survey provides a systematic and deployment-oriented review of efficient video diffusion models. We propose a unified categorization that organizes existing methods into four classes of main paradigms, including step distillation, efficient attention, model compression, and cache/trajectory optimization. Building on this categorization, we respectively analyze algorithmic trends of these four paradigms and examine how different design choices target two core objectives: reducing the number of function evaluations and minimizing per-step overhead. Finally, we discuss open challenges and future directions, including quality preservation under composite acceleration, hardware-software co-design, robust real-time long-horizon generation, and open infrastructure for standardized evaluation. To the best of our knowledge, our work is the first comprehensive survey on efficient video diffusion models, offering researchers and engineers a structured overview of the field and its emerging research directions.
\end{abstract}



\maketitle

\begin{bibunit}
\section{Introduction}\label{chap:introduction}
\begin{figure*}[t]
    \centering
    \includegraphics[width=\linewidth]{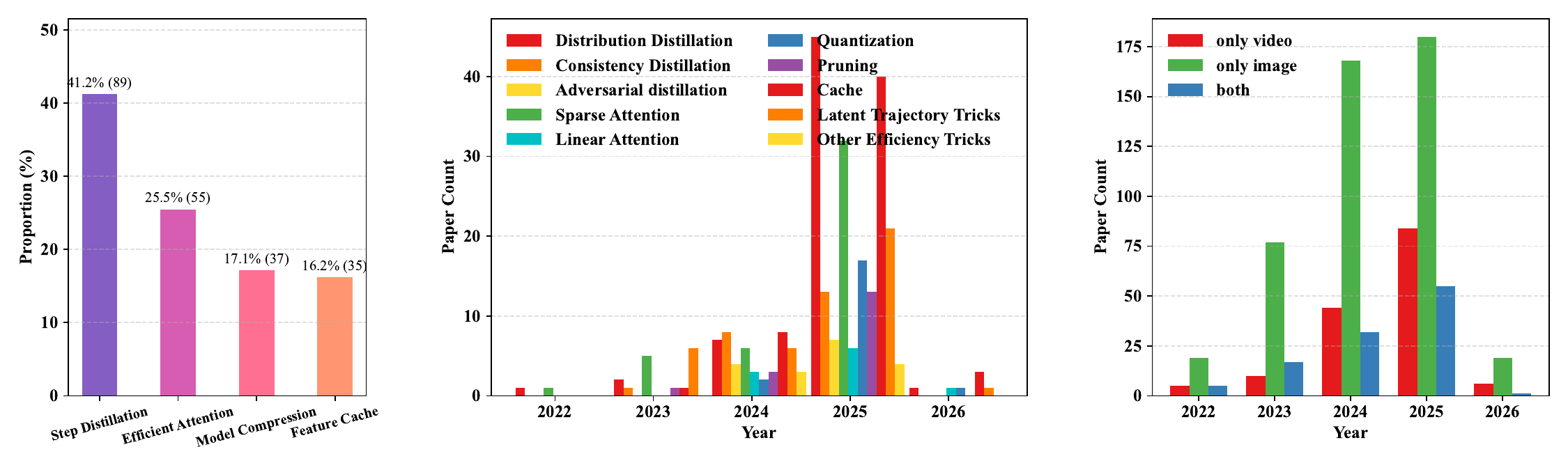}
    \vspace{-30pt}
    \caption{\textbf{Left:} Distribution of literature across various accelerated sampling algorithms for video diffusion models. \textbf{Middle:} Publication trends and adoption growth of accelerated sampling algorithms for video diffusion models (2022–2026). \textbf{Right:} Comparative growth trends of accelerated sampling algorithms in image versus video diffusion tasks (2022–2026).}
    \label{fig:dmd-survey-trend}
    \vspace{-15pt}
\end{figure*}
Video diffusion has transitioned toward Diffusion Transformer (DiT) based systems~\cite{DIT,VIT,wan2025,hyvideo1p5,hunyuanvideo,ltx2}, enabling increasingly coherent and high-fidelity generation. Driven by architectural scaling and long-context temporal modeling~\cite{kling15,seedance}, this trajectory has made video diffusion a dominant paradigm for generative media. However, these gains expose a fundamental systems bottleneck~\cite{fastvideo}: video generation compounds computation across spatial resolution, temporal length, and iterative denoising, causing attention and memory traffic to dominate runtime~\cite{sparge_attn,https://doi.org/10.48550/arxiv.2505.13389,dmd2,shao2025magicdistillationweaktostrongvideodistillation}. As a result, even state-of-the-art models face substantial deployment barriers in real-time and long-horizon settings~\cite{https://doi.org/10.48550/arxiv.2506.08009,Yin_2025}. Therefore, acceleration and efficiency must be treated as first-class objectives alongside generation quality.

Compared with image generation, efficient video generation is a harder systems problem because video synthesis must jointly handle high spatial resolution, long temporal horizon, and iterative denoising. This joint burden causes inference cost to grow along both token count and sampling depth, and directly determines whether interactive creation, real-time avatar generation, live streaming, and long-horizon world simulation are feasible under strict latency and memory constraints. Fig.~\ref{fig:dmd-survey-trend} shows a concrete research gap: in 2022--2026, image-only acceleration papers account for 
64.1\%
(463/722), while video-only papers only for 
20.6\%
(149/722) and cross-domain papers for 
15.2\%
(110/722). At the same time, video-only studies are accelerating quickly (from 5 in 2022 to 84 in 2025), indicating an early but fast-consolidating stage where step distillation and sparse-attention-style methods currently dominate. Existing surveys mainly focus on diffusion models in general or image-domain efficiency. To the best of our knowledge, this work is the first dedicated survey that systematically targets efficient generation for video diffusion models. Beyond transferring ideas from image diffusion, we emphasize why video must be treated as an independent efficiency research problem: temporal coherence constraints, long-context memory growth, and cross-step error accumulation make many image-efficient designs non-trivial to scale to video. In this survey, we provide a systematic and deployment-oriented review of accelerated sampling for video diffusion models, with the goal of consolidating a fast-growing but fragmented literature into a unified technical framework for both researchers and practitioners.

Concretely, these three axes interact multiplicatively rather than additively during deployment. Increasing resolution expands the number of latent tokens processed by each layer, extending duration enlarges the temporal context that must remain globally consistent, and multi-step denoising repeats this high-cost computation dozens of times within a single sample. In practice, this means that moving from short clips to minute-scale generation or from lower to higher resolution can quickly exceed the memory budget of a single accelerator and force unfavorable trade-offs in batch size, latency, or model capacity. The burden is even heavier for DiT-style backbones, where attention, KV-cache movement, and activation storage scale with both token volume and temporal span, making throughput highly sensitive to sequence length. As a result, efficient video diffusion is not only a matter of reducing FLOPs, but of controlling end-to-end system pressure so that generation remains feasible for production workloads such as interactive editing, live content pipelines, and persistent simulation.

The paper is organized as follows. Section~\ref{sec:notation} introduces preliminaries, including video diffusion foundations, metrics, and the categorization. Sections~\ref{sec:step_distillation}--\ref{sec:cache_and_trajectory} form the core technical review. Section~\ref{sec:step_distillation} covers consistency, distribution, and adversarial distillation. Section~\ref{sec:efficient_attention} reviews static sparse attention, dynamic sparse attention, and linear-hybrid attention designs. Section~\ref{sec:model_compression} discusses quantization and pruning strategies. Section~\ref{sec:cache_and_trajectory} analyzes feature and key-value (KV) caching, trajectory modification, and parallel execution optimizations. Section~\ref{sec:future_direction} discusses unresolved problems and practical research directions, including quality preservation under composite acceleration, hardware-aware algorithm design, and open-data infrastructure. Finally, Section~\ref{sec:conclusion} concludes with key takeaways.

\section{Preliminaries}\label{sec:notation}
\begin{table*}[t]
    \centering
    \small
    \caption{The list of symbols.}
    \label{tab:list_of_symbols}
    \begin{tabular}{@{}c p{0.34\textwidth} c p{0.34\textwidth}@{}}
        \toprule
        Symbols & Meanings & Symbols & Meanings \\
        \midrule
        \addlinespace[2pt]
        \multicolumn{4}{@{}l@{}}{\textbf{\textit{Diffusion Foundations}}} \\
        \cmidrule(lr){1-4}
        $x_0$ & clean data sample (video/image) & $x_t$ & noisy sample at time $t$ \\
        $z_0$ & clean latent & $z_t$ & noisy latent at time $t$ \\
        $x_t^i$ & $i$-th noisy chunk at time $t$ & $z_t^i$ & $i$-th noisy latent chunk at time $t$ \\
        $t$ & diffusion time index & $T$ & total diffusion time horizon in training \\
        $S$ & number of reverse sampling steps & $\Delta t$ & step size in the sampler \\
        $\alpha_t$ & signal scaling in the noise schedule & $\sigma_t$ & noise scaling in the noise schedule \\
        $\epsilon$ & Gaussian noise $\mathcal{N}(0, I)$ & $\epsilon_\theta$ & denoiser/noise predictor \\
        $s_\theta$ & score estimator $\nabla_{z_t}\log q(z_t)$ & $v_\theta$ & velocity predictor \\
        $q(z_0)$ & data distribution & $q(z_t|z_0)$ & forward diffusion process \\
        $p_\theta(z_0|z_t)$ & learned reverse process & $c$ & conditioning input (prompt/control) \\
        $w$ & classifier-free guidance weight & $M$ & memory/VRAM footprint \\
        \addlinespace[4pt]
        \multicolumn{4}{@{}l@{}}{\textbf{\textit{Step Distillation}}} \\
        \cmidrule(lr){1-4}
        $\theta$ & student model parameters & $\phi$ & teacher model parameters \\
        $\psi$ & discriminator/critic parameters & $G_\theta$ & distilled or few-step generator \\
        $D_\psi$ & discriminator for adversarial training & $f^{\text{pf}}$ & consistency function (PF-ODE form) \\
        $f^\phi_{\text{fake}}$ & critic/score term for fake samples & $f^\phi_{\text{real}}$ & critic/score term for real samples \\
        $\mathcal{L}_{\mathrm{reg}}$ & regression/distillation loss & $\mathcal{L}_{\mathrm{adv}}$ & adversarial loss \\
        \addlinespace[4pt]
        \multicolumn{4}{@{}l@{}}{\textbf{\textit{Efficient Attention}}} \\
        \cmidrule(lr){1-4}
        $Q,K,V$ & attention queries, keys, and values & $A$ & attention output \\
        $N$ & sequence length in tokens & $d$ & attention head dimension \\
        $\rho$ & sparsity keep ratio & $N_Q$/$N_K$ & query/key tile or block size \\
        \addlinespace[4pt]
        \multicolumn{4}{@{}l@{}}{\textbf{\textit{Model Compression \& Cache/Trajectory Optimization}}} \\
        \cmidrule(lr){1-4}
        $b_w, b_a$ & weight/activation bitwidths & $Q(\cdot; s)$ & quantizer with scale $s$ \\
        $r_\ell$ & layer keep ratio in pruning & $r_b$ & block keep ratio in model pruning \\
        $F_t^{\ell}$ & feature map at layer $\ell$ and timestep $t$ & $C_t^{\ell}$ & cached feature at layer $\ell$ and timestep $t$ \\
        $C^{\text{KV}}_\text{size}$ & KV-cache size & $k^{\text{refresh}}$ & cache refresh interval \\
        $L^{f}$ & number of blocks in the model & $\eta$ & speedup ratio \\
        \bottomrule
    \end{tabular}
\end{table*}

This section establishes formal notations for the model architectures and diffusion process in the context of video generation. We then categorize the prevailing techniques for accelerating sampling in video diffusion models, discussing the specific challenges each category addresses alongside their algorithmic foundations. All symbols used are defined in Table~\ref{tab:list_of_symbols}.

\subsection{Foundation of Video Diffusion Models}\label{sec:notation:vdm}

\textbf{VAE Latent Space.} In recent video diffusion frameworks~\cite{wan2025,hunyuanvideo,stepfunvideo}, the generative process operates within a lower-dimensional latent space to reduce computational overhead. An input video
$x_0 \in \mathbb{R}^{C_x \times F_x \times H_x \times W_x}$ is first mapped to a latent representation $z_0 \in \mathbb{R}^{C \times F \times H \times W}$ via a pretrained Video Variational Autoencoder (VAE). Here, $\{C_x, F_x, H_x, W_x\}$ and $\{C, F, H, W\}$ denote the channel count, frame count, height, and width of the pixel-space video and its latent counterpart, respectively. To balance visual fidelity with synthesis efficiency, mainstream architectures typically employ a temporal compression ratio of $F = \lfloor \frac{F_x - 1}{4} \rfloor + 1$, while the spatial downsampling factors, $\lfloor \frac{H_x}{H} \rfloor$ and $\lfloor \frac{W_x}{W} \rfloor$, are commonly fixed at 8~\cite{hunyuanvideo,wan2025}, 16~\cite{hyvideo1p5}, or 32~\cite{ltx2}. 

\hspace*{-\parindent}\textbf{Diffusion Process.} The diffusion process in latent video models is defined as follows: given a clean latent variable $z_0$ sampled from the data distribution $q(z_0)$, we define a forward diffusion process $\{z_t\}_{t \in [0,T]}$ over the interval $[0, T]$. This process starts at $z_0$ and progressively adds noise such that for any timestep $t \in (0,T]$, the distribution of the noisy latent $z_t$ is given by:
\begin{equation}
q(z_t|z_0) = \mathcal{N}(z_t|\alpha_t z_0, \sigma_t^2 \mathbf{I}).
\label{eq:discrete_forward}
\end{equation}
where $\alpha_t$ (signal scale) and $\sigma_t$ (noise scale) are differentiable functions of $t$ with bounded derivatives. The choice for $\alpha_t$ and $\sigma_t$ is referred to as the noise schedule of a diffusion model. Song et al.~\cite{sde} show that Eq.~\ref{eq:discrete_forward} can be written as the following stochastic differential equation (SDE):
\begin{equation}
dz_t = f(t)z_tdt + g(t)dw_t, \quad z_0 \sim q(z_0),
\label{eq:continuous_forward}
\end{equation}
where $f(t) = \frac{d\log \alpha_t}{dt}$, $g^2(t) = \frac{d\sigma^2_t}{dt} - 2 \frac{d\log \alpha_t}{dt}\sigma^2_t$, and $w_t \in \mathbb{R}^{B\times C \times F \times H \times W}$ is the standard Wiener process. The forward process has a corresponding reverse process, which can be expressed as
\begin{equation}
dz_t = [f(t)z_t- g^2(t)\nabla_z \log q(z_t)]dt + g(t)dw_t, \quad z_T \sim \mathcal{N}(\mathbf{0},\sigma_T\mathbf{I}).
\label{eq:continuous_reverse}
\end{equation}
The forward process facilitates noise injection into the latent variables during training, whereas the reverse process enables iterative multi-step sampling during inference. The training objectives for these diffusion models primarily focus on estimating the score function $-\sigma_t \nabla_{z} \log q(z_t)$ in Eq.~\ref{eq:continuous_reverse}. This optimization target is typically:
\begin{equation}
\mathcal{L}_{dm} =  \int_0^T \mathbb{E}_{q(z_t)}[\Vert \epsilon_\theta(z_t,t) + \sigma_t \nabla_{z} \log q(z_t) \Vert_2^2]dt,
\label{eq:old_diffusion_loss}
\end{equation}
where $z_t = \alpha_t z_0 + \sigma_t\epsilon$ and $\epsilon_\theta(\cdot,\cdot)$ denotes a score function estimator learned during training. 

Many recent diffusion/flow generative models, including SD3.5~\cite{SD35}, FLUX~\cite{FLUX}, and HunyuanVideo-1.5~\cite{kong2024hunyuanvideo}, adopt the flow matching paradigm~\cite{iclr22_rect}. The core principle involves employing a linear probability path where the noise schedules are defined as $\alpha_t = 1 - t$ and $\sigma_t = t$, respectively. Under this formulation, the training objective shifts from score matching to regressing the velocity field $v = z_1 - z_0$. Consequently, the model, parameterized as a velocity field $v_\theta(\cdot, \cdot)$, is optimized to predict this flow. The corresponding loss function is:
\begin{equation}
\mathcal{L}_{fm} =  \int_0^T \mathbb{E}_{q(z_t)}[\Vert v_\theta(z_t,t) - (z_1-z_0) \Vert_2^2]dt.
\label{eq:new_diffusion_loss}
\end{equation}

\hspace*{-\parindent}\textbf{Model Architecture.} Current video diffusion architectures are primarily categorized into two classes of vision transformers (ViTs)~\cite{VIT}: dual-stream designs based on multi-modal diffusion transformers (MMDiT) and single-stream DiT variants. The former facilitates modality fusion by concatenating text and visual tokens into a unified sequence for joint attention calculation. In contrast, the latter maintains separate streams and employs cross-attention mechanisms to inject textual information into the visual tokens. Architecturally, single-stream DiT designs are generally more computationally efficient and lightweight than their MMDiT counterparts. Within the open-source ecosystem, the HunyuanVideo (v1.0/1.5)~\cite{hunyuanvideo,hyvideo1p5} series adopts the MMDiT framework, whereas Wan (v2.1/2.2)~\cite{wan2025} utilizes the single-stream DiT architecture.

\hspace*{-\parindent}\textbf{Efficiency and Quality Metrics.} Accelerated video diffusion is typically evaluated with a combination of efficiency and quality metrics. On the efficiency side, commonly reported indicators include hardware-dependent throughput, inference latency, speedup, and peak VRAM usage, together with hardware-agnostic quantities such as NFE and FLOPs/MACs. On the quality side, widely used benchmarks include Fr\'echet Video Distance (FVD)~\cite{fvd}, frame-wise image metrics such as Fr\'echet Inception Distance (FID)~\cite{fid} and Learned Perceptual Image Patch Similarity (LPIPS)~\cite{LPIPS}, and holistic video evaluation suites such as VBench~\cite{vbench,vbench++}; motion-aware metrics including DOVER~\cite{dover_score} and FasterVQA~\cite{fastvqa} are additionally used when temporal fidelity is central.

\subsection{Categorization of Efficient Video Diffusion Models}\label{sec:notation:categorization}

\begin{figure*}[t]
    \centering
    \includegraphics[width=\linewidth]{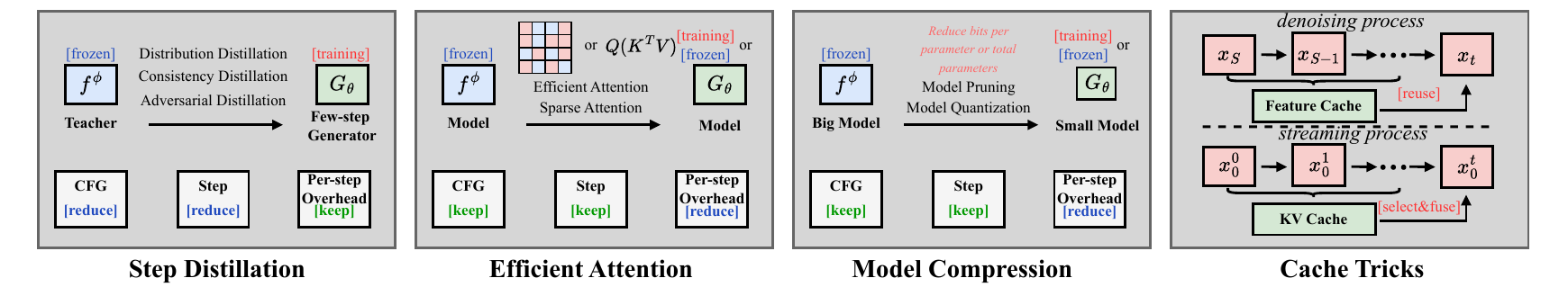}
    \caption{Conceptual illustration of efficient video diffusion generation. The main methods are organized into four major paradigms: step distillation, efficient attention, model compression, and cache/trajectory optimization.}
    \label{fig:dmd-survey-overview}
    \vspace{-15pt}
\end{figure*}

As summarized in Fig.~\ref{fig:dmd-survey-overview} and further detailed in Table~\ref{tab:video-accel-categorization},
we categorize accelerated video diffusion methods into four paradigms: (i) step distillation,
which reduces latency primarily by compressing denoising depth, i.e., NFE;
(ii)
efficient attention,
which reduces per-step complexity by sparsifying or linearizing expensive attention operations; (iii)
model compression,
which reduces compute and memory footprints through quantization and pruning; and (iv)
cache/trajectory optimization,
which
improves runtime by reusing intermediate states and redesigning denoising trajectories to avoid redundant computation.

\tikzset{
    taxRoot/.style={rectangle, draw=hidden-draw, rounded corners=3pt, text opacity=1, minimum height=2.2em, inner xsep=5pt, inner ysep=4pt, align=center, fill=hidden-draw, text=white, line width=1.0pt, font=\bfseries\small},
    taxL1/.style={rectangle, draw=hidden-draw, rounded corners=3pt, text opacity=1, minimum height=1.9em, inner xsep=4pt, inner ysep=4pt, align=center, fill=hidden-draw!92, text=white, line width=0.95pt, font=\bfseries\normalsize},
    taxL2/.style={rectangle, draw=hidden-draw!85, rounded corners=3pt, text opacity=1, minimum height=1.8em, inner xsep=4pt, inner ysep=4pt, align=center, fill=black!3, text=black, line width=0.85pt, font=\normalsize},
    taxLeaf/.style={rectangle, draw=hidden-draw!80, rounded corners=3pt, text opacity=1, minimum height=2.0em, inner xsep=4pt, inner ysep=4pt, align=left, fill=hidden-pink!55, text=black, line width=0.8pt, font=\footnotesize}
}

\providecommand{\taxleaf}[2]{#1\\[-1pt]{\scriptsize #2}}
\providecommand{\taxbadge}[1]{\textcolor{hidden-draw!95}{\raisebox{0.05ex}{\scriptsize #1}}}

\providecommand{\Description}[1]{}

\begin{table*}[h]
    \centering
    \resizebox{0.99\linewidth}{!}{
    \begin{forest}
        forked edges,
        for tree={
            grow=east,
            reversed=true,
            anchor=base west,
            parent anchor=east,
            child anchor=west,
            base=center,
            font=\footnotesize,
            edge+={darkgray!80, line width=0.9pt},
            s sep=4pt,
            l sep=7pt,
            inner xsep=2pt,
            inner ysep=3pt,
        },
        where level=0{taxRoot, text width=8.8em}{},
        where level=1{taxL1, text width=13.8em}{},
        where level=2{taxL2, text width=11.8em}{},
        where level=3{taxLeaf, text width=16.5em}{},
        [
            Accelerated Sampling for\\Video Diffusion Models
            [
                \taxbadge{\textbullet}~Step Distillation
                [
                    Distribution Distillation
                    [
    Streaming Distillation,
                        content={\taxleaf{Streaming Distillation}{[1--18]}}
                    ]
                    [
    Non-Streaming Distillation,
                        content={\taxleaf{Non-Streaming Distillation}{[19--27]}}
                    ]
                ]
                [
    Consistency Distillation,
                        content={\taxleaf{Consistency Distillation}{[28--35]}}
                ]
                [
                    Adversarial Distillation
                    [
    Combined Distillation,
                        content={\taxleaf{Combined Distillation}{[36]}}
                    ]
                    [
    Independent Distillation,
                        content={Independent Distillation}
                    ]
                ]
            ]
            [
                \taxbadge{\ensuremath{\square}}~Efficient Attention
                [
                    Sparse Attention
                    [
    Dynamic Sparsity,
                        content={\taxleaf{Dynamic Sparsity}{[37--57]}}
                    ]
                    [
    Static Sparsity,
                        content={\taxleaf{Static Sparsity}{[58--64]}}
                    ]
                ]
                [
                    Linear Attention
                    [
    Training-Based,
                        content={\taxleaf{Training-Based}{[65--73]}}
                    ]
                ]
            ]
            [
                \taxbadge{\ensuremath{\triangle}}~Model Compression
                [
                    Quantization
                    [
    Quantization-Aware Training,
                        content={\taxleaf{Quantization-Aware Training}{[74--79]}}
                    ]
                    [
    Post-Training Quantization,
                        content={\taxleaf{Post-Training Quantization}{[80--91]}}
                    ]
                ]
                [
    VAE Compression,
                        content={\taxleaf{VAE Compression}{[92--99]}}
                ]
                [
                    Pruning
                    [
    Token Pruning,
                        content={\taxleaf{Token Pruning}{[100--108]}}
                    ]
                    [
    Channel Pruning,
                        content={\taxleaf{Channel Pruning}{[109]}}
                    ]
                    [
    Model Pruning,
                        content={\taxleaf{Model Pruning}{[110--114]}}
                    ]
                ]
            ]
            [
                \taxbadge{\ensuremath{\diamond}}~Cache and Trajectory Optimization
                [
                    Cache
                    [
    Feature Cache,
                        content={\taxleaf{Feature Cache}{[115--131]}}
                    ]
                    [
    KV Cache,
                        content={\taxleaf{KV Cache}{[132--145]}}
                    ]
                ]
                [
                    Latent Trajectory Tricks
                    [
    Noise and State Modification,
                        content={\taxleaf{Noise and State Modification}{[146--148]}}
                    ]
                    [
    Trajectory Modification,
                        content={\taxleaf{Trajectory Modification}{[149--161]}}
                    ]
                    [
    Parallel Computation,
                        content={\taxleaf{Parallel Computation}{[162--164]}}
                    ]
                ]
                [
                    Other Efficiency Methods,
                    content={\taxleaf{Other Efficiency Methods}{[165--169]}}
                ]
            ]
        ]
    \end{forest}

    }
    \caption{A condensed categorization of accelerated sampling algorithms for video diffusion models used in the main manuscript. \textit{Note: Bracketed IDs inside nodes are local main manuscript indices.}}
    \label{tab:video-accel-categorization}
        \vspace{-15pt}
\end{table*}

\hspace*{-\parindent}\textbf{Step Distillation.} Step distillation accelerates video diffusion by compressing multi-step denoising into few-step or one-step generation, thereby directly reducing NFE. In this setting, the teacher is the original multi-step video diffusion model, while the student is the distilled few-step generator trained to reproduce the teacher with much lower sampling cost. Representative methods include consistency distillation~\cite{song2023consistency}, distribution distillation~\cite{dmd,dmd2}, and adversarial distillation~\cite{add,ladd}, which differ in their emphasis on trajectory consistency, output-distribution alignment, and perceptual refinement, respectively.

\hspace*{-\parindent}\textbf{Efficient Attention.} Efficient attention reduces the per-step cost of video diffusion by lowering the compute and memory overhead of attention, which is often the dominant bottleneck in DiT backbones. Representative directions include IO-aware attention kernels such as FlashAttention~\cite{flashattn3}, sparse attention that skips low-value interactions~\cite{sparge_attn,chen2025rainfusion20temporalspatialawarenesshardwareefficient}, and linear or hybrid attention that further reduces sequence-length scaling~\cite{ghafoorian2026rehyatrecurrenthybridattention,hu2024zigmaditstylezigzagmamba}. These methods improve throughput, but stronger approximations can weaken long-range temporal dependencies and global motion consistency.

\hspace*{-\parindent}\textbf{Model Compression.} Model compression reduces per-step latency and memory pressure by lowering arithmetic precision or removing redundant computation, mainly through quantization and pruning. Quantization maps weights and activations to low-bit representations and is currently more deployment-ready, while pruning removes low-utility tokens, channels, or blocks but often requires stronger recovery to offset quality loss~\cite{Chen_2025,liu2024taqdittimeawarequantizationdiffusion,kim2025vip,wu2025tamingdiffusiontransformerefficient}. In video diffusion, the practical challenge is not compression alone but preserving temporal consistency and perceptual quality after compression.

\hspace*{-\parindent}\textbf{Cache and Trajectory Optimization.} Cache and trajectory optimization accelerates video diffusion by reusing historical computation and reshaping denoising execution paths to reduce redundant work without always changing the nominal model depth. Representative families include feature cache~\cite{liu2025fastcachefastcachingdiffusion,ma2025magcachefastvideogeneration}, KV cache for causal or streaming generation~\cite{pfp,ji2025memflowflowingadaptivememory}, noise or state modification~\cite{liu2025free4dtuningfree4dscene}, trajectory redesign~\cite{sabour2024alignstepsoptimizingsampling}, and parallel execution~\cite{fang2024xditinferenceenginediffusion}. Their shared challenge is controlling long-horizon drift while maintaining throughput and memory scalability.

\section{Step Distillation}\label{sec:step_distillation}

\begin{figure*}[t]
    \centering
    \includegraphics[width=\linewidth]{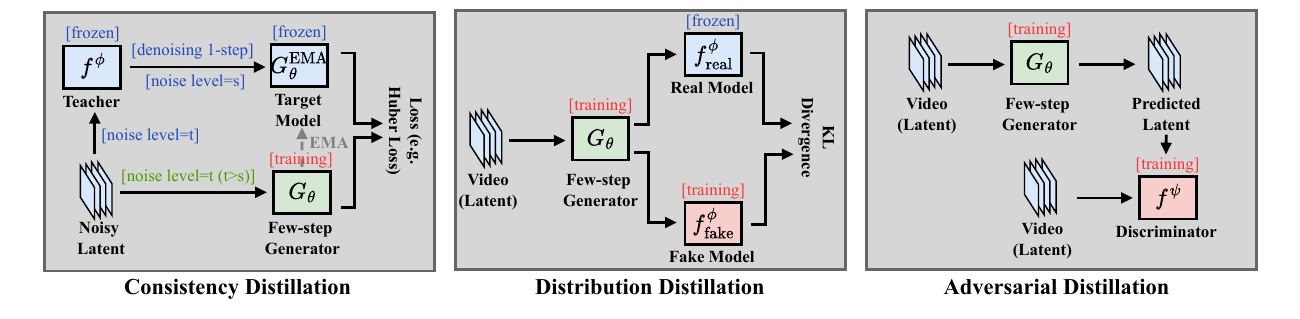}
    \caption{Overview of step distillation for accelerated video diffusion. The paradigm reduces NFE by distilling multi-step denoising trajectories into few-step or one-step generators, spanning consistency distillation, non-streaming/streaming distribution distillation, and adversarial variants.}
    \label{fig:dmd-survey-step-distillation-overview}
\end{figure*}

Step distillation is designed to accelerate video diffusion by reducing denoising steps, i.e., NFE, while preserving spatial fidelity and temporal coherence. As shown in Fig.~\ref{fig:dmd-survey-step-distillation-overview}, this section reviews three main branches: consistency distillation, distribution distillation, and adversarial distillation. Within distribution distillation, we further distinguish non-streaming and streaming settings. Together, these branches illustrate how step-distillation methods balance speed, robustness, and quality.

\subsection{Consistency Distillation}
Consistency distillation, as exemplified by LCD~\cite{song2023consistency}, distills a pretrained latent diffusion model through a consistency function
\begin{equation}
f^\text{pf}(z_t,t,G_\theta)=c_\text{skip}(t)z_t+c_\text{out}(t)G_\theta(z_t,t),
\end{equation}
where $c_\text{skip}(t)$ and $c_\text{out}(t)$ are timestep-dependent coefficients that balance how much of the noisy latent $z_t$ and model prediction $G_\theta(z_t,t)$ are retained in the consistency mapping. Here, $f^\text{pf}$ denotes the consistency transformation itself, and can be applied to either the teacher or student outputs. In Fig.~\ref{fig:dmd-survey-step-distillation-overview}(left), 
the teacher is denoted $f^\phi$, 
which is typically instantiated by an 
exponential moving average 
(EMA) model $G_\theta^{\mathrm{EMA}}$. This formulation encourages two noisy states on the same trajectory (e.g., $z_t$ and an earlier state $z_s^\phi$ with $s<t$) to map to the same clean target $z_0$. Intuitively, this lets a few-step student learn timestep-to-timestep self-consistency rather than reproduce every intermediate step. The online model is then trained against an 
EMA target model $G_\theta^{\mathrm{EMA}}$ by minimizing
\begin{equation}
\mathcal{L}_{\text{lcd}}=\mathbb{E}_{z_0,t,s,\text{ where }(s<t)}\!\left[d\!\left(f^\text{pf}(z_t,t,G_\theta),f^\text{pf}(z^\phi_s,s,G_\theta^{\mathrm{EMA}})\right)\right],
\end{equation}
\rev{with $z^\phi_s=z_t-(t-s)G_\theta^{\mathrm{EMA}}(z_t,t)$ denoting the EMA-refined earlier state and $d(\cdot,\cdot)$ denoting a discrepancy measure, typically the mean squared error.

In video generation, this paradigm is usually supplemented with 
(i) motion injection and alignment, and (ii)
preference optimization (described below).
rCM~\cite{rCM} further combines Meanflow~\cite{meanflow} and Distribution Matching Distillation (DMD)~\cite{dmd} to scale the framework to video diffusion. This design is attractive because it inherits the stability of image-domain consistency training, but the same conservative self-consistency objective also makes it harder to push to the most aggressive low-step regime.}

\hspace*{-\parindent}\textbf{Motion Injection and Alignment.} \rev{Motion-injected consistency distillation aligns not only denoised latents but also motion-sensitive representations extracted from the online prediction and EMA target. Concretely, the student prediction at noisy state $z_t$ is passed through a motion extractor $M(\cdot)$ and compared
against $z^\phi_s$
with $d(\cdot,\cdot)$.
The design of $M(\cdot)$ varies across methods: VideoLCM~\cite{https://doi.org/10.48550/arxiv.2312.09109} uses the identity mapping, MCM~\cite{https://doi.org/10.48550/arxiv.2406.06890} uses a learnable extractor, and DCM~\cite{https://doi.org/10.48550/arxiv.2506.03123} uses temporal latent differences. This variation suggests that the real bottleneck is not whether the motion cues are needed, but how explicitly they must be encoded before consistency supervision becomes useful for video.}



\hspace*{-\parindent}\textbf{Preference Optimization.} \rev{Integrating preference optimization with step distillation often improves few-step generators. In consistency distillation, the dominant choice is direct optimization with reward models applied to the predicted clean latent rather than a separate iterative preference-learning loop~\cite{liu2024alignment}. In practice, this component behaves more like a quality compensator than a core acceleration mechanism.}

\hspace*{-\parindent}\textbf{Specific Applications of Consistency Distillation.} \rev{Beyond general video generation, consistency distillation has been extended to specialized tasks that demand both efficiency and robustness. Representative examples include human-centric animation with LM2D~\cite{yin2024lm2dlyricsmusicdrivendance} and UniAnimate~\cite{wang2024unianimatetamingunifiedvideo}, high-resolution restoration with UltraVSR~\cite{Liu_2025_UltraVSR} and DOVE~\cite{https://doi.org/10.48550/arxiv.2505.16239}, video face enhancement with VividFace~\cite{zhang2025vividfacehighqualityefficientonestep}, and non-pixel decision domains such as ManiCM~\cite{lu2025manicmrealtime3ddiffusion}. This breadth is itself informative: consistency distillation is adopted most naturally in tasks where stability and controllability matter more than chasing the absolute minimum step count.}

\hspace*{-\parindent}\textbf{Discussion.} Consistency distillation remains the most stable route to few-step generation and is attractive when robustness and controllability matter more than extreme step compression. Its strength is broad task transferability, but \rev{its conservative matching objective also leaves a lower ceiling on aggressive NFE reduction than modern distribution-matching pipelines}.

\subsection{Distribution Distillation}

Distribution distillation, represented by methods such as DMD~\cite{dmd} and DMD2~\cite{dmd2}, matches the student-induced distribution to the teacher-induced distribution rather than enforcing strict timestep-by-timestep consistency. In DMD-style formulations, the teacher-side and student-side score estimators are denoted by $f^\phi_{\text{real}}$ and $f^\phi_{\text{fake}}$, respectively, and are both evaluated on a noisy latent $\hat{z}_t$ constructed from the current generator output. A standard DMD objective is
\begin{equation}
\mathcal{L}_{\text{dmd}}=D_{\mathrm{KL}}\!\left(\hat{z}_t - t f^\phi_{\text{fake}}(\hat{z}_t,t)\,\Vert\, \hat{z}_t - t f^\phi_{\text{real}}(\hat{z}_t,t)\right),\quad
\mathcal{L}_{\text{critic}}=\mathbb{E}_{t,\epsilon}\big\|\epsilon-\hat{z}_0-f^\phi_{\text{fake}}(\hat{z}_t,t)\big\|_2^2,
\end{equation}
where $\hat{z}_0=G_\theta(\epsilon,1)$ is the current generator output
with student generator
$G_\theta$,
$\hat{z}_t=(1-t)\hat{z}_0+t\epsilon$, 
and $f^\phi_{\text{fake}}$ is learned by the auxiliary critic while $f^\phi_{\text{real}}$ is derived from the teacher side. The generator $G_\theta$ and the critic that parameterizes $f^\phi_{\text{fake}}$ are optimized alternately to maintain stability. This makes distribution distillation the primary route for pushing video diffusion into very low-step regimes. This approach can be further divided into the non-streaming and streaming settings.
Here and throughout this subsection, we follow the common DMD literature shorthand of writing the KL divergence in terms of the corresponding teacher- and student-induced random variables, with the associated pushforward densities understood implicitly.

\subsubsection{Non-Streaming Distribution Distillation}\label{sec:non_streaming_distribution_distillation}

\rev{In video diffusion, most non-streaming distribution-distillation methods innovate around the distribution-matching objective itself rather than around new motion modules. This does not mean motion is unimportant; rather, motion-oriented adaptation was explored earlier, especially in LCM-style video distillation, whereas the DMD line has concentrated on making extreme low-step distillation stable enough to work in practice.}

\hspace*{-\parindent}\textbf{Improved Distribution Matching Distillation.} 
\rev{TDM~\cite{https://doi.org/10.48550/arxiv.2503.06674} and BLADE~\cite{gu2025blade} move toward trajectory matching, whereas AVDM2~\cite{https://doi.org/10.48550/arxiv.2412.05899} and MagicDistillation~\cite{shao2025magicdistillationweaktostrongvideodistillation} strengthen the DMD line. The trend is to retain DMD-level compression while reducing optimization brittleness.}

\hspace*{-\parindent}\textbf{Specific Applications of Non-Streaming Distribution Distillation.} \rev{Beyond generic video synthesis, non-streaming distribution distillation has been adapted to restoration and enhancement tasks such as FlashVSR~\cite{https://doi.org/10.48550/arxiv.2510.12747} and GFix~\cite{teng2025gfixperceptuallyenhancedgaussian}, human-centric animation with DiffusionTalker~\cite{chen2023diffusiontalkerpersonalizationaccelerationspeechdriven}, and controllable manipulation with EquiVDM~\cite{liu2025equivariancefastsamplingvideo}. This diversity suggests that once distribution matching is stabilized, it transfers naturally to deployment settings that demand very low-step generation.}

\subsubsection{Streaming Distribution Distillation}\label{sec:streaming_distribution_distillation}

Streaming video generation, characterized by the sequential synthesis of video chunks in a causal manner, is essential for interactive applications requiring low-latency and infinite-horizon output, such as live streaming and cloud gaming. As shown in Fig.~\ref{fig:dmd-survey-trend}, there
is an explosive surge in research on real-time video generation
in 2025--2026.
\rev{This literature can be read in two stages. First, representative frameworks such as CausVid~\cite{Yin_2025} and Self-Forcing~\cite{https://doi.org/10.48550/arxiv.2506.08009} establish the basic recipe for streaming distribution distillation. Second, later methods extend this recipe along several design axes, including causal attention design, rollout optimization, target engineering, and multi-stage training. This shift suggests that real-time video is constrained not only by NFE, but also by whether causal training remains stable over long horizons.}

\hspace*{-\parindent}\textbf{$\blacktriangleright$ CausVid.} CausVid~\cite{Yin_2025} initiates streaming distribution distillation by transferring knowledge from a bidirectional teacher to a causal student. The training process treats the video as a sequence of chunks and optimizes the causal student to predict the clean latent of each chunk conditioned on ground-truth history. Formally, for the $i$-th chunk $z_0^i$ sampled from real data, the training flow injects noise to state $t$ and performs causal denoising as follows:
\begin{equation}
z_{t}^i = (1-t) z_{0}^i + t \epsilon, \quad \epsilon \sim \mathcal{N}(0, \mathbf{I}), \quad \hat{z}_{0}^i = G_\theta(z_{t}^i, t \mid z_{\text{data}}^{<i}),
\label{eq:causvid_flow}
\end{equation}
where 
$G_\theta$ 
is the student generator, and
$z_{\text{data}}^{<i}$ represents the ground-truth history enforced via block-wise causal masking. 
$G_\theta$ 
is then optimized using the DMD objective to align its predicted distribution with that of the bidirectional teacher through the teacher-side term $t f^\phi_{\text{real}}(\hat{z}_t,t)$:
\begin{equation}
\mathcal{L}_{\text{dmd}} = D_{\mathrm{KL}}\!\left(\hat{z}_t - t f^\phi_{\text{fake}}(\hat{z}_t,t)\,\Vert\, \hat{z}_t - t f^\phi_{\text{real}}(\hat{z}_t,t)\right),
\label{eq:causvid_loss}
\end{equation}
where $\hat{z}_t = (1-t)\text{cat}([\hat{z}_0^0,\hat{z}_0^1,\cdots,\hat{z}_0^{F-1}])+t\epsilon$, 
$\text{cat}(\cdot)$ 
is the concatenation operator, and 
$\epsilon$ is the Gaussian noise, respectively. \rev{This asymmetric formulation matters because it keeps access to data, but the history used for conditioning is still the ground-truth past context $z_{\text{data}}^{<i}$ rather than the model's own generated history. As a result, the train-time conditioning context remains cleaner than the test-time context, so exposure errors can still accumulate during autoregressive rollout. This limitation is precisely why later work moved beyond pure teacher-forced conditioning.}

\begin{figure*}[h]
    \centering
    \includegraphics[width=0.65\linewidth]{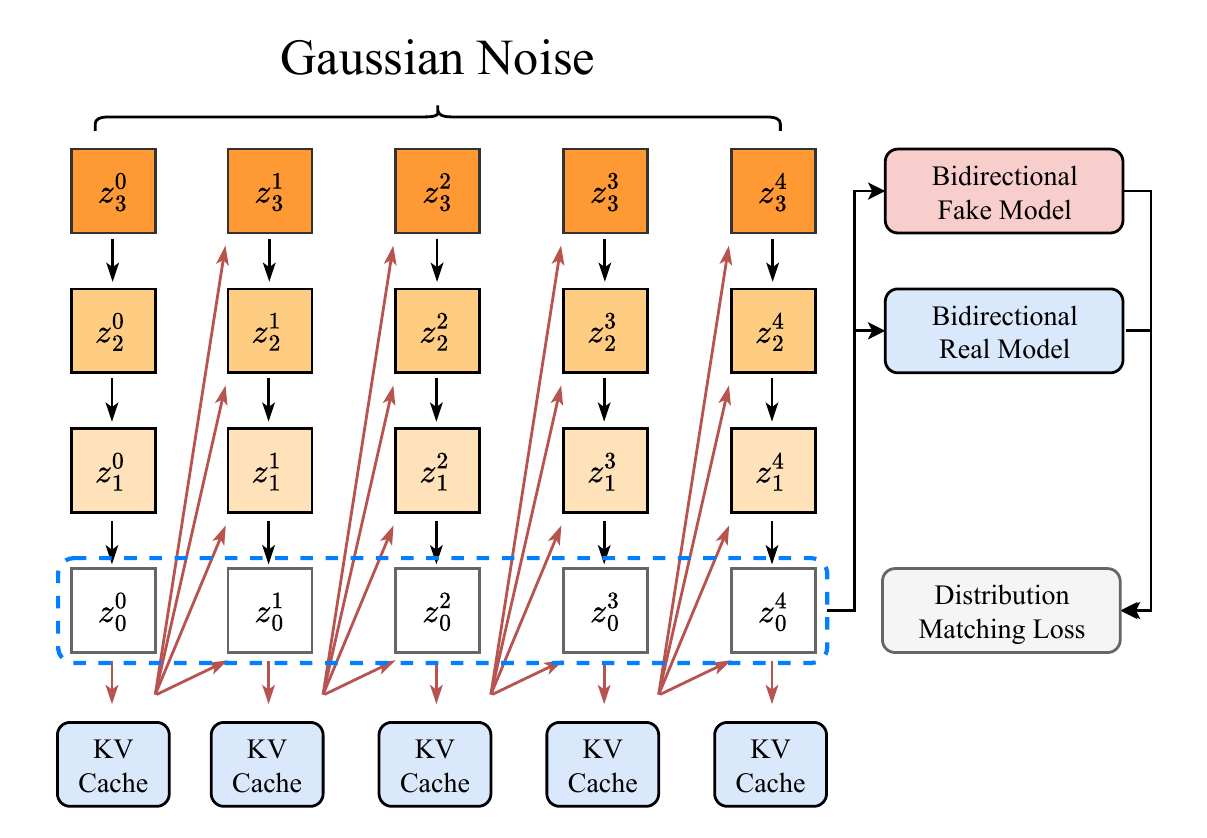}
        \vspace{-15pt}
    \caption{Overview of the Self-Forcing algorithm. This framework serves as the foundation for various real-time video generation methods. At its core, the algorithm leverages a causal model for video synthesis and subsequently employs standard DMD to facilitate few-step generation, thereby enabling the real-time, streaming output of video chunks.}
    \label{fig:self-forcing}
        \vspace{-15pt}
\end{figure*}

\hspace*{-\parindent}\textbf{$\blacktriangleright$ Self-Forcing.} Self-Forcing~\cite{https://doi.org/10.48550/arxiv.2506.08009} marks the first successful scaling of causal models to robust real-time generation. To mitigate the exposure bias that hinders prior approaches, it bridges the train-test discrepancy by incorporating autoregressive rollouts during training. The name ``Self-Forcing'' emphasizes that the model is trained to condition on its own previously generated history, rather than relying only on ground-truth past chunks as in teacher-forced training; the empirical results in Self-Forcing~\cite{https://doi.org/10.48550/arxiv.2506.08009} show that this self-conditioning strategy substantially improves long-horizon robustness.
Formally, let $z^i$ denote the $i$-th latent chunk and let $\mathcal{C}_{<i}$ denote the KV cache accumulated from all previously generated chunks before chunk $i$ is synthesized. The generation process therefore follows a causal chain in which the current chunk is conditioned on this historical cache. For consistency with Fig.~\ref{fig:self-forcing}, we denote the denoised prediction for chunk $i$ by $z_0^i$ in the rollout below. For a standard 4-step distillation process (steps $t \in \{3, 2, 1, 0\}$), the causal flow for the $i$-th chunk and subsequent cache update can be described as
\begin{equation}
z_{3}^i \xrightarrow{G_\theta(\cdot|\mathcal{C}_{<i})} z_2^i \xrightarrow{G_\theta(\cdot|\mathcal{C}_{<i})} z_{1}^i \xrightarrow{G_\theta(\cdot|\mathcal{C}_{<i})} z_{0}^i.
\label{eq:self_forcing_flow}
\end{equation}
The cache is then updated as $\mathcal{C}_{i} \leftarrow \text{Update}(\mathcal{C}_{<i}, \text{Proj}(z_{0}^i))$, where $\text{Proj}(z_{0}^i)$ denotes the projected cache features extracted from the newly generated clean chunk $z_{0}^i$, and the next chunk $i+1$ is generated using the updated cache $\mathcal{C}_{i}$. This process is illustrated in Fig.~\ref{fig:self-forcing}.
By unfolding this sequence during training, the generator $G_\theta$ develops robustness to its own historical predictions in $\mathcal{C}_{<i}$. In practice, the target in Eq.~\ref{eq:self_forcing_loss} is built from teacher-forced chunk latents to stabilize optimization, while autoregressive rollouts close the train--test exposure gap. Training is subsequently governed by a DMD objective applied to the noisy latent
$\hat{z}_t = (1-t)\text{cat}([z_0^0,z_0^1,\cdots,z_0^{F-1}])+t\epsilon$
used for target construction:
\begin{equation}
\mathcal{L}_{\text{dmd}} = D_{\mathrm{KL}}\!\left(\hat{z}_t - t f^\phi_{\text{fake}}(\hat{z}_t,t)\,\Vert\, \hat{z}_t - t f^\phi_{\text{real}}(\hat{z}_t,t)\right),
\label{eq:self_forcing_loss}
\end{equation}
\rev{This design alleviates long-horizon error accumulation, but does not eliminate it. More importantly, its core robustness mechanism is still rollout self-conditioning rather than a pretraining-style recipe that keeps absorbing supervision from high-quality video data. Hence, it still falls short of a clearly scaling recipe.}

\rev{Building on these representative foundations, later streaming-distillation methods mainly evolve along four relatively orthogonal directions: causal attention design, causal rollout optimization, DMD-target engineering, and multi-stage training. The remainder of this subsection follows this organization.}

\hspace*{-\parindent}\textbf{Causal Attention Design.} This direction focuses on how streaming models restrict memory growth while preserving temporal consistency under causal attention. Most frameworks therefore use attention sinks and windowing to preserve temporal consistency while controlling KV-cache cost, as in Rolling Forcing~\cite{https://doi.org/10.48550/arxiv.2509.25161}, Reward Forcing~\cite{lu2025reward}, LongLive~\cite{https://doi.org/10.48550/arxiv.2509.22622}, and VideoSSM~\cite{yu2025videossmautoregressivelongvideo}. \rev{Thus, full-history attention is no longer viable, but windowing alone still cannot guarantee long-range recall.}

\hspace*{-\parindent}\textbf{Causal Rollout Optimization.} This direction focuses on improving the autoregressive rollout itself so that exposure drift accumulates more slowly over time. Rolling Forcing~\cite{https://doi.org/10.48550/arxiv.2509.25161} introduces a rolling-window joint denoising scheme with progressively increasing noise levels within the window, thereby adding a form of local non-causality inside the rolling window to suppress long-horizon error accumulation. Self-Forcing++~\cite{Sun_2025} instead injects noise into self-generated long rollouts, and Live Avatar~\cite{huang2025live} perturbs historical KV states during distillation; both can be viewed as controlled-corruption strategies that improve robustness to accumulated errors. \rev{These follow-ups indicate that plain causal rollout is not enough: later methods increasingly rely either on controlled corruption or on limited local non-causality to keep errors from compounding.}

\hspace*{-\parindent}\textbf{DMD-Target Engineering.} This direction focuses on modifying the distribution-matching target so that DMD better reflects streaming-specific objectives and failure modes. The motivation is that plain DMD aligns distributions in a generic sense, but does not explicitly encode streaming-specific concerns such as motion saliency, cross-chunk consistency, or reward-guided prioritization. Reward Forcing~\cite{lu2025reward} introduces Re-DMD, which scales distribution-matching gradients using motion-based VLM rewards to prioritize high-dynamic regions. TalkingMachines~\cite{https://doi.org/10.48550/arxiv.2506.03099} integrates a flow-matching objective with the distribution matching loss for real-time audio-driven animation, and LongLive~\cite{https://doi.org/10.48550/arxiv.2509.22622} applies distillation only to newly generated clips during rolling extension. \rev{The common trend is task-aware target engineering, but these methods also show that better targets alone cannot fully solve causal instability.}

\hspace*{-\parindent}\textbf{Multi-Stage Training.} This direction focuses on bridging the gap between bidirectional pretraining and causal deployment through staged curricula. A common assumption in streaming distillation is to initialize from a bidirectional pretrained model, since starting from a purely causal model is usually much weaker. This creates a bidirectional-to-causal gap: the two architectures expose different attention patterns and conditioning histories during generation, so transferring a bidirectional model into causal deployment is not seamless. To mitigate this gap, Live Avatar~\cite{huang2025live} and HY-WorldPlay~\cite{sun2025worldplay} use a sequential curriculum with Diffusion Forcing~\cite{chen2024diffusion} as an intermediate stage before Self-Forcing. In contrast, Causal Forcing~\cite{zhu2026causal} argues that Diffusion Forcing is less stable than Teacher Forcing~\cite{zhu2026causal}, and therefore adopts a pipeline of Teacher Forcing, ODE initialization~\cite{iclr22_rect}, and then Self-Forcing. \rev{The broader message is that this bidirectional-to-causal gap is still unresolved, and current success often depends on carefully staged training rather than a single inherently stable objective.}

\textbf{Discussion.} \rev{Distribution distillation is the only step-distillation approach that has reliably entered the extremely low-step regime of one-step to four-step generation, which explains its central role in practical video acceleration~\cite{fastvideo,ltx2}. Its main weakness is not expressivity but fragility: once causal rollout and memory constraints enter the loop, stability engineering becomes as important as the distillation loss itself.}

\subsection{Adversarial Distillation}

Adversarial distillation focuses on improving perceptual realism and motion sharpness in few-step generation, either as a companion objective or as a standalone optimization target. Representative methods such as eADD~\cite{add} inject GAN-style supervision through
\begin{equation}
\mathcal{L}_{\text{adv}}^{G}=-\mathbb{E}_{\epsilon}\big[\log D_{\psi}(G_\theta(\epsilon,1))\big],\quad
\mathcal{L}_{\text{adv}}^{D}=-\mathbb{E}_{z_0}\big[\log D_{\psi}(z_0)\big]-\mathbb{E}_{\epsilon}\big[\log(1-D_{\psi}(G_\theta(\epsilon,1)))\big].
\label{eq:baseline_gan_loss}
\end{equation}
where $G_\theta(\epsilon,1)$ denotes the generated sample obtained from Gaussian noise $\epsilon$, $z_0$ denotes a real clean sample, and $D_\psi$ is the discriminator parameterized by $\psi$. This objective sharpens details and is commonly used either jointly with consistency or distribution distillation or as a later refinement stage. This subsection includes both combined and independent adversarial designs.

\subsubsection{Combined Adversarial Distillation}\label{sec:combined_adversarial_distillation}

While the conceptual roots of adversarial distillation lie in GANs~\cite{StyleGAN}, the inherent instability of GAN training remains a significant challenge when applied to video diffusion distillation. In the combined setting, adversarial design is most often used as an auxiliary objective to enhance consistency- or distribution-based distillation rather than as a standalone distillation route. Recent advancements over established methods, such as ADD~\cite{add}, LADD~\cite{ladd}, and DMD2~\cite{dmd2}, therefore focus on improving how this auxiliary adversarial component is integrated, including strategies such as scaling the discriminator capacity.

\subsubsection{Independent Adversarial Distillation}\label{sec:independent_adversarial_distillation}

Independent Adversarial Distillation refers to the exclusive use of adversarial objectives, rather than integrating them with other distillation paradigms. In this setting, Adversarial Distillation transitions from an auxiliary role to the primary distillation objective, a configuration that frequently suffers from training instability. To ensure convergence, certain frameworks, such as Seaweed-APT~\cite{https://doi.org/10.48550/arxiv.2501.08316}, perform Adversarial Distillation on models already optimized via Consistency Distillation. Their objective is to further refine the generative quality of an existing few-step model, rather than executing a full-step distillation from scratch.

\textbf{Discussion.} \rev{Adversarial distillation is most useful as a quality compensation mechanism for few-step generators, especially for motion realism and texture sharpness. As an independent objective it remains too unstable for large-scale video models, so the credible path is to use it on top of distribution- or consistency-based training.}

\subsection{Summary and Outlook}

Step distillation remains the strongest acceleration lever because it directly reduces the denoising depth that dominates diffusion inference. However, the central question is no longer whether few-step video generation is possible, but whether it remains stable as compression becomes more aggressive and training scales up. In practice, current DMD- and LCM-style pipelines for video often require substantial hyperparameter tuning, can converge too quickly, and do not yet scale gracefully as training continues. An important future direction is therefore to make step distillation interact more naturally with large-scale video-model training and adaptation pipelines, so that it behaves less like a fragile post hoc compression stage and more like a stable optimization component whose quality can continue improving with further training.

A second direction is to study step distillation as part of a broader acceleration stack rather than as an isolated objective. Reducing NFE alone does not remove the dominant per-step costs of attention kernels, memory traffic, and backbone redundancy, so practical systems will increasingly need joint designs that combine step distillation with sparse attention, pruning, quantization, or cache reuse. The challenge is not merely stacking modules: each approximation perturbs denoising dynamics, and naive combinations can amplify temporal drift, quality loss, or error accumulation. Future work should therefore emphasize co-designed hybrid pipelines in which step reduction and per-step efficiency are optimized together, with explicit recovery mechanisms for motion fidelity and temporal consistency.

A third direction is the shift from offline short-video acceleration to streaming and real-time generation. Recent causal and chunk-wise pipelines show that step distillation can become the backbone of low-latency video synthesis, but they also expose new bottlenecks such as exposure bias, history drift, KV-cache growth, and the bidirectional-to-causal transition gap. For this setting, the next generation of methods will likely need causal-aware distillation objectives, more robust rollout training, and multi-stage curricula that better align pretraining, distillation, and long-horizon deployment. If these issues are addressed, step distillation may evolve from a few-step sampling technique into a general foundation for real-time and long-horizon video generation.

\section{Efficient Attention}\label{sec:efficient_attention}

\begin{figure*}[t]
    \centering
    \includegraphics[width=\linewidth]{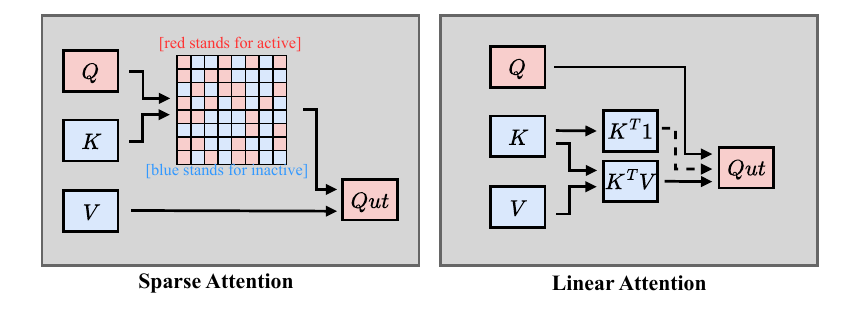}
        \vspace{-30pt}
    \caption{Overview of efficient attention for video diffusion acceleration. The methods reduce per-step overhead via dynamic/static sparsity and linear or hybrid attention designs while balancing approximation error and kernel-level efficiency.}
    \label{fig:dmd-survey-efficient-attention-overview}
    \vspace{-15pt}
\end{figure*}

Efficient attention is designed to accelerate video diffusion by reducing per-step compute and memory traffic in attention kernels, which are often the dominant runtime bottleneck in DiT-based systems. Fig.~\ref{fig:dmd-survey-efficient-attention-overview} summarizes this landscape. This section follows a progression from the most regular and deployment-friendly designs to the most adaptive and model-altering ones: we begin with static sparse attention, then move to dynamic sparse attention, and finally review training-based linear or hybrid alternatives under the dual criteria of approximation quality and hardware executability.

At the systems level, many later efficient-attention designs build on FlashAttention~\cite{flashattn3}, which accelerates exact attention through IO-aware execution rather than a new attention formula. It exploits the GPU memory hierarchy by tiling $Q$, $K$, and $V$ into SRAM-sized blocks, loading them from HBM, performing blockwise attention updates on chip, and writing partial results back only after sufficient local computation. This tiling strategy avoids materializing the full $N\times N$ attention matrix and substantially reduces HBM traffic, which is why it serves as a practical foundation for later sparse and hybrid designs. Building on this systems baseline, we begin with static sparse attention, the most regular sparse regime, before moving to dynamic sparse attention and then to training-based linear or hybrid alternatives.

\subsection{Static Sparse Attention}\label{sec:static_sparse_attention}

\begin{figure*}[h]
    \centering
    \includegraphics[width=\linewidth]{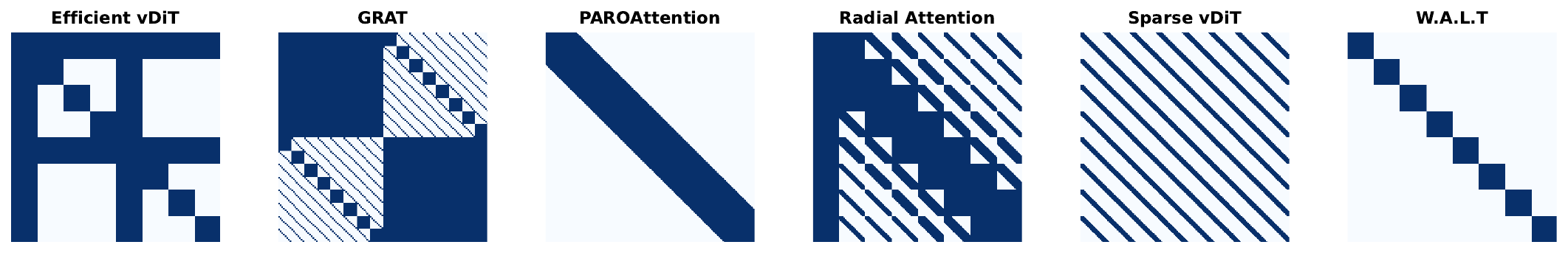}
        \vspace{-15pt}
    \caption{Overview of representative static attention masks used by different methods under a common illustrative setup with sequence length $N_Q = N_K = 40960$ across 8 frames. For clarity, each mask is visualized using the default hyperparameters reported in the original publication.}
    \label{fig:static_mask_grid_2x3}
        \vspace{-8pt}
\end{figure*}

\rev{Sparse attention mechanisms predicated on static attention masks do not adjust the selected $K$ and $V$ tokens according to the input. Typically, static sparse attention uses generalized backends that accept an attention mask together with $Q$, $K$, and $V$, avoiding custom forward or backward kernels. Its main advantage is compatibility with widely adopted backends such as SageAttention~\cite{sageattention3}, FlexAttention~\cite{flexattention}, and Block Sparse Attention~\cite{blocksparseattention}; its drawback is that it inevitably discards some $K$ and $V$ values relative to full attention. This makes static sparsity appealing when deployment regularity matters more than adaptivity, but it also means mask design quality largely determines its ceiling.}

\hspace*{-\parindent}\textbf{Static Mask Patterns.} \rev{Static masks usually encode spatiotemporal locality together with hardware-friendly block structure. Figure~\ref{fig:static_mask_grid_2x3} compares six representative layouts, spanning the tiled locality of Efficient-vDiT~\cite{https://doi.org/10.48550/arxiv.2502.06155}, the grouped block structure of GRAT~\cite{ren2025groupingfirstattendingsmartly}, the reordered diagonal band of PAROAttention~\cite{https://doi.org/10.48550/arxiv.2506.16054}, the radial sparsity of Radial Attention~\cite{li2025radialattentiononlogn}, the uniform striped pattern of Sparse-vDiT~\cite{https://doi.org/10.48550/arxiv.2506.03065}, and the coarse block-diagonal schedule used by W.A.L.T~\cite{https://doi.org/10.48550/arxiv.2412.14169}. These methods differ mainly in how wide the locality prior is, how much global structure they retain, and how directly the mask maps to efficient block execution. The practical lesson is that mask design is the real bottleneck: once a fixed prior misses the interactions a video actually needs, static sparsity has no adaptive mechanism to recover them.}

\textbf{Discussion.} \rev{Static sparse attention remains attractive because regular masks are easy to implement, benchmark, and deploy. Its ceiling is lower, however, because any mismatch between the fixed mask prior and the actual video content becomes irreversible information loss. This rigidity is precisely why later methods moved toward dynamic sparse attention, which adapts the active token or KV selection pattern to the current video content at the cost of extra routing and execution complexity.}

\subsection{Dynamic Sparse Attention}\label{sec:dynamic_sparse_attention}

\rev{Compared with static masks, dynamic sparse attention adapts the active token or KV pattern to the current input and can therefore preserve important interactions more flexibly. The trade-off is that the system must first estimate which query, key, or value tokens should remain active, so routing itself consumes compute and often requires more specialized kernels or runtime support before the algorithmic gain can translate into end-to-end speedup. We organize this space along two orthogonal axes: the routing policy used to select active tokens, and the sparsity objective that specifies whether sparsity is applied only to $\{K, V\}$ or across the full $\{Q, K, V\}$ set. In practice, the main difference is where the routing cost is paid and how much structure the hardware can still exploit.}

\subsubsection{Dynamic Routing Policy}

Dynamic routing is categorized into three primary methodologies, which include \textbf{(1)} computation-based routing that utilizes efficient attention or similarity calculations across the current $Q, K$, and $V$ tensors to determine the routing policy~\cite{zhang2025trainingfreeefficientvideogeneration,sparge_attn,qiao2025flashomniunifiedsparseattention,chen2025rainfusion20temporalspatialawarenesshardwareefficient,xu2025xattentionblocksparseattention,https://doi.org/10.48550/arxiv.2505.18809,https://doi.org/10.48550/arxiv.2506.23858,https://doi.org/10.48550/arxiv.2502.01776,https://doi.org/10.48550/arxiv.2509.01085,https://doi.org/10.48550/arxiv.2505.14708}, \textbf{(2)} prior-driven routing that leverages external domain knowledge such as human pose keypoints for token selection or object-centric preservation in video editing~\cite{kahatapitiya2024objectcentricdiffusionefficientvideo,gao2025ditvrzeroshotdiffusiontransformer,https://doi.org/10.48550/arxiv.2510.02617}, and \textbf{(3)} statistic-based routing that utilizes historical statistics to estimate the optimal sparsity configuration for the current attention operation~\cite{https://doi.org/10.48550/arxiv.2502.21079}.

\hspace*{-\parindent}\textbf{Computation-based Routing.} \rev{Computation-based routing compresses $Q$, $K$, and $V$ to estimate token saliency cheaply before sparse attention is applied. Representative examples include DraftAttention~\cite{https://doi.org/10.48550/arxiv.2505.14708}, RainFusion2.0~\cite{chen2025rainfusion20temporalspatialawarenesshardwareefficient}, VSA~\cite{https://doi.org/10.48550/arxiv.2505.13389}, and SpargeAttention~\cite{sparge_attn}; other methods instead use low-bit score estimation, as in FlashOmni~\cite{qiao2025flashomniunifiedsparseattention}. The practical difficulty is that the proxy used for routing must be both cheap and faithful: once routing cost or approximation error becomes too large, the theoretical sparsity gain quickly erodes.}

\hspace*{-\parindent}\textbf{Prior‑Driven Routing.} \rev{Prior‑driven routing uses external cues to decide which tokens should be preserved. Object‑Centric Diffusion~\cite{kahatapitiya2024objectcentricdiffusionefficientvideo} biases token merging toward background regions, DiTVR~\cite{gao2025ditvrzeroshotdiffusiontransformer} uses optical-flow trajectories, and Input‑Aware Sparse Attention~\cite{https://doi.org/10.48550/arxiv.2510.02617} uses human-pose keypoints. These priors can be powerful in editing or human-centric settings, but their advantage weakens when the scene structure is open-ended and no strong prior is available.}

\hspace*{-\parindent}\textbf{Statistic‑Based Routing.} \rev{Statistic‑based routing reuses historical statistics, such as cached log‑sum‑exp values or prior attention distributions, to approximate the current sparse mask. AdaSpa~\cite{https://doi.org/10.48550/arxiv.2502.21079} exemplifies this idea by exploiting the empirical stability of sparse indices across denoising steps and recovering them through an online LSE-cached search. The appeal is lower routing cost, but the assumption of cross-step stability can break when content changes sharply, which limits how aggressively these methods can prune.}

\subsubsection{Sparse Objective}
Sparse attention mechanisms can be categorized into two primary groups according to their sparsity objectives, which include \textbf{(1)} methods that sparsify only the $K$ and $V$ tensors to reduce the memory footprint of the KV cache~\cite{https://doi.org/10.48550/arxiv.2505.18809,https://doi.org/10.48550/arxiv.2506.23858,https://doi.org/10.48550/arxiv.2502.01776,chen2025rainfusion20temporalspatialawarenesshardwareefficient,https://doi.org/10.48550/arxiv.2505.14708,xu2025xattentionblocksparseattention}, and \textbf{(2)} approaches that apply sparsity across the complete set of $\{Q, K, V\}$ components to further optimize computational throughput~\cite{kahatapitiya2024objectcentricdiffusionefficientvideo,https://doi.org/10.48550/arxiv.2509.01085}. For sparse attention, most methods emphasize KV-only sparsification to preserve generative quality while reducing KV storage. By contrast, QKV sparsification appears in token-merging or query-pruning paradigms, such as object-centric token merging and bidirectional sparse attention.

\hspace*{-\parindent}\textbf{Sparsity of Key and Value.} \rev{Most dynamic sparse attention methods preserve dense queries and sparsify only keys and values, since the query represents the token currently being updated. Representative KV-only designs include VORTA~\cite{https://doi.org/10.48550/arxiv.2505.18809}, VMoBA~\cite{https://doi.org/10.48550/arxiv.2506.23858}, Sparse VideoGen~\cite{https://doi.org/10.48550/arxiv.2502.01776}, RainFusion2.0~\cite{chen2025rainfusion20temporalspatialawarenesshardwareefficient}, DraftAttention~\cite{https://doi.org/10.48550/arxiv.2505.14708}, and Xattention~\cite{xu2025xattentionblocksparseattention}. With $N_Q$ query tokens, $N_K$ key tokens, attention feature dimension $D$, and keep ratio $\rho \in (0,1]$, the cost of dense attention $\mathcal{O}(N_Q N_K D)$ becomes $\mathcal{O}(\rho N_Q N_K D)$. This is the safer sparsity regime, but also the less aggressive one: quality is easier to preserve precisely because the query side is left untouched.}

\hspace*{-\parindent}\textbf{Sparsity of Query, Key and Value.} \rev{A more aggressive line also sparsifies queries, either by pruning them directly or by merging tokens before attention so that the whole $\{Q,K,V\}$ set becomes shorter. BSA~\cite{https://doi.org/10.48550/arxiv.2509.01085} prunes locally redundant queries within 3D blocks, whereas Object‑Centric Diffusion~\cite{kahatapitiya2024objectcentricdiffusionefficientvideo} merges background tokens more aggressively than object tokens. This raises the speed ceiling, but it also turns routing mistakes into irreversible information loss because the model no longer keeps a dense set of active queries.}

\textbf{Discussion.} \rev{Dynamic sparse attention is more flexible than static sparsity because routing adapts to content instead of committing to a fixed mask. Its main weakness is that the savings are fragile: routing itself costs computation, and highly irregular masks are hard to execute efficiently on real kernels. As a result, many methods look strong at the algorithmic level but underdeliver end-to-end unless they co-design sparsity with executability.}

\subsection{Training-Based Linear Attention}\label{sec:training_based_linear_attention}

Linear attention~\cite{ghafoorian2025attention} rewrites the order of matrix multiplication as
\begin{equation}
A=(QK^{\top})V=Q(K^{\top}V),
\end{equation}
\rev{so that the $D\times D$ matrix $K^{\top}V$ is formed first and then left-multiplied by $Q$. This changes the dominant cost from sequence-length quadratic $\mathcal{O}(N_Q N_K D)$ to approximately $\mathcal{O}((N_Q + N_K) D^2)$, which is beyond what most sparse attention methods can reach asymptotically. The trade-off is that this departs substantially from softmax attention, so training-free conversion is usually ineffective and video diffusion methods must retrain or redesign the backbone. We therefore categorize such algorithms into three groups: \textbf{(1)} full replacement with linear attention~\cite{ma2025consistentcontrollableimageanimation,zhang2025efficientlongdurationtalkingvideo}, \textbf{(2)} hybrid linear-sparse configurations~\cite{https://doi.org/10.48550/arxiv.2509.24006,ghafoorian2026rehyatrecurrenthybridattention}, and \textbf{(3)} the Mamba series~\cite{hu2024zigmaditstylezigzagmamba,wang2024unianimatetamingunifiedvideo,wang2024opticalflowrepresentationalignment,zatsarynna2025mantadiffusionmambaefficient}. The practical question is therefore not whether linear attention is cheaper, but whether the resulting approximation remains faithful enough to justify retraining the whole video stack.}

\hspace*{-\parindent}\textbf{Full Replacement with Linear Attention.} \rev{Full-replacement methods substitute quadratic self-attention with linear attention throughout the backbone, so the savings are realized end-to-end rather than only in selected blocks. In practice, this usually means either training a linear-attention backbone from scratch or converting a pretrained DiT and then retraining it. The difficulty is not the asymptotic gain but the approximation gap: once all attention layers depart from softmax behavior, motion-critical interactions become harder to preserve.}

\hspace*{-\parindent}\textbf{Hybrid Configurations of Linear and Sparse Attention.} \rev{Hybrid designs use linear attention for the bulk of the sequence while reserving softmax attention for critical regions. SLA~\cite{https://doi.org/10.48550/arxiv.2509.24006} and ReHyAt~\cite{ghafoorian2026rehyatrecurrenthybridattention} exemplify this direction. This is more credible than full replacement because it accepts that linear attention alone is not lossless for video, but it still depends heavily on whether the model can correctly identify which blocks deserve exact attention and whether the resulting pattern remains kernel-efficient.}

\hspace*{-\parindent}\textbf{Mamba Series.} \rev{The Mamba line replaces attention with state space models to obtain linear scaling through recurrent-style sequence processing. Zigma~\cite{hu2024zigmaditstylezigzagmamba}, UniAnimate~\cite{wang2024unianimatetamingunifiedvideo}, MedSora~\cite{wang2024opticalflowrepresentationalignment}, and MANTA~\cite{zatsarynna2025mantadiffusionmambaefficient} represent this direction in different video domains. Their value is clear for long-context modeling, but they should be viewed as a backbone redesign rather than a drop-in acceleration trick, since they no longer inherit the same inductive bias as softmax attention.}

\textbf{Discussion.} \rev{Training-based linear attention is attractive when very long contexts dominate and retraining is acceptable. In current video diffusion systems, hybrid sparse-linear designs are more credible than full replacement because pure linearization still struggles to preserve motion-critical interactions.}

\subsection{Summary and Outlook}

Efficient attention has shifted from optional optimization to core systems infrastructure for deployable video diffusion. Sparse attention currently provides the best practical quality-speed trade-off because it preserves critical long-range interactions while still reducing dominant attention costs, whereas pure linearization often requires stronger approximation assumptions. The central bottleneck is hardware coupling: gains from routing and mask design are realized only when they map cleanly to kernel-friendly, balanced execution. Near-term progress will depend on co-design across algorithm, mask structure, quantization format, and runtime scheduling.

\section{Model Compression}\label{sec:model_compression}

\begin{figure*}[t]
    \centering
    \includegraphics[width=\linewidth]{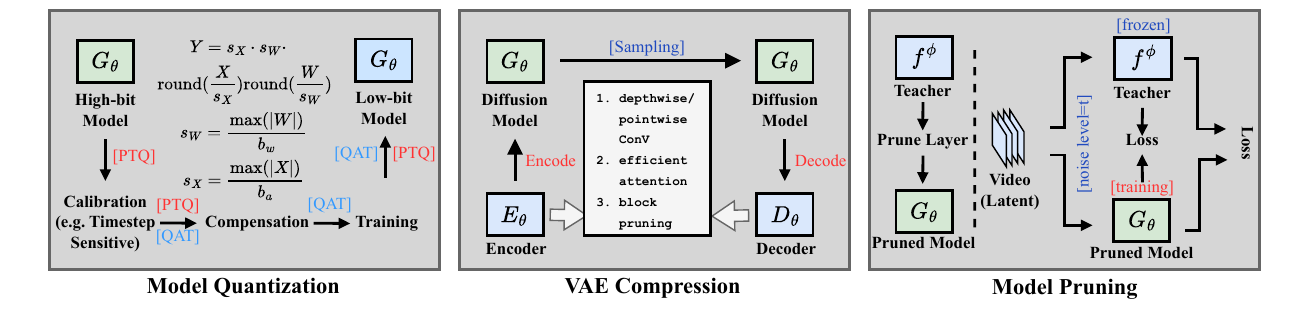}
        \vspace{-30pt}
    \caption{Overview of model compression for accelerated video diffusion. The figure highlights quantization-aware training (QAT), post-training quantization (PTQ), VAE compression, and multi-granularity pruning. Specifically, VAE compression reduces video autoencoding overhead and the length of the latent sequence, while the remaining branches employ compensation mechanisms to mitigate quality degradation.}
    \label{fig:dmd-survey-model-compression-overview}
        \vspace{-10pt}
\end{figure*}

Model compression is designed to accelerate video diffusion by reducing arithmetic workload, memory bandwidth pressure, and model footprint at inference time. As shown in Fig.~\ref{fig:dmd-survey-model-compression-overview}, this section reviews quantization-aware training (QAT), post-training quantization (PTQ), \revnew{VAE compression,} and pruning under a shared question: how to maximize efficiency while minimizing temporal and perceptual degradation.

\subsection{Quantization-Aware Training}\label{sec:quantization_aware_training}
Quantization-aware training (QAT)~\cite{https://doi.org/10.48550/arxiv.2509.23681,https://doi.org/10.48550/arxiv.2506.04648,https://doi.org/10.48550/arxiv.2505.22167,feng2025s} optimizes a low-bit student model under quantization perturbations throughout the training phase, as opposed to applying quantization post-optimization. A basic quantization view maps a tensor $X$ to low-bit integers through
\begin{equation}
\bar{X}=\text{round}\left(\frac{X}{s}\right),\quad s=\frac{\max|X|}{2^{b-1}-1},
\end{equation}
where $b$ denotes the target bitwidth and $\bar{X}$ is the quantized integer tensor associated with the floating-point tensor $X$. Under the same notation, $\bar{W}$ denotes the quantized integer version of the weight matrix $W$, so that a linear layer is approximated by
\begin{equation}
Y=XW\approx (s_x\cdot\bar{X})(s_w\cdot\bar{W})=(s_xs_w)\cdot(\bar{X}\bar{W}).
\end{equation}
This is commonly summarized by the W$b_w$A$b_a$ notation, and modern video diffusion quantizers often use group-wise scaling to balance speed and fidelity. For video diffusion models, let $G_{\theta}$ denote the frozen full-precision teacher and $G_{\theta}^{(b_w,b_a)}$ represent the quantized student with weight and activation bit-widths $b_w$ and $b_a$, respectively. Utilizing a quantizer $Q(\cdot; s)$, a standard training objective is formulated as:
\begin{equation}
\min_{\theta,s} \mathbb{E}_{z_0,z_1,t}\left[\left\Vert G_{\theta}^{(b_w,b_a)}(z_t,t)-(z_1-z_0)\right\Vert_2^2\right]
+\lambda\,\mathcal{L}_{\mathrm{reg}}\left(G_{\theta}^{(b_w,b_a)},G_{\theta}\right),
\end{equation}
where $\theta$ are the student network parameters, $s$ denotes quantization scales (or step sizes) used by $Q(\cdot;s)$, and $\mathcal{L}_{\mathrm{reg}}$ denotes a generic teacher-alignment regularizer that encourages the quantized student to remain close to the frozen full-precision model, for example through output-, feature-, or distribution-level consistency constraints. Concretely, $s$ is usually defined per channel, per group, or per tile, and may be static, layer-wise updated, or directly learned. Typical examples include clipping bounds and activation/weight scaling factors in low-bit integer mapping. Under this view, PTQ mainly calibrates $s$ with limited optimization, while QAT jointly updates $\theta$ and $s$ to absorb quantization-induced bias during training.

\rev{Recent video quantization methods largely converge on compensation of quantization-induced bias. QuantSparse~\cite{https://doi.org/10.48550/arxiv.2509.23681} and FPSAttention~\cite{https://doi.org/10.48550/arxiv.2506.04648} compensate within attention, Q-VDiT~\cite{https://doi.org/10.48550/arxiv.2505.22167} compensates in parameter and feature space, and S$^2$Q-VDiT~\cite{feng2025s} emphasizes data- and token-level compensation. The key lesson is that low-bit video diffusion succeeds only when quantization is treated as an error-control problem rather than a pure numeric compression problem.}

\textbf{Discussion.} \rev{QAT is the right tool when the target bit-width is so low that calibration alone cannot absorb error. Its weakness is cost: the fidelity gain is real, but the extra training burden is not always justified outside deployment settings.}

\subsection{Post-Training Quantization}\label{sec:post_training_quantization}

Post-training quantization (PTQ)~\cite{liu2024taqdittimeawarequantizationdiffusion,https://doi.org/10.48550/arxiv.2406.02540,Chen_2025,https://doi.org/10.48550/arxiv.2505.18663,yang2025lrqditlogrotationposttrainingquantization,liu2025clqcrosslayerguidedorthogonalbased} maintains a fixed pretrained diffusion backbone and optimizes quantization parameters through calibration, as opposed to updating model weights in an end-to-end manner. Utilizing the previously established notation, a generic PTQ objective is formulated as follows:
\begin{equation}
\min_{s,\xi}\; \mathbb{E}_{z_0,z_1,t}\!\left[\left\Vert G_{\theta}^{(b_w,b_a)}(z_t,t; s,\xi)-(z_1-z_0)\right\Vert_2^2\right],
\end{equation}
where $\theta$ remains constant, while $s$ and $\xi$ are made explicit as calibration-dependent quantization variables in $G_{\theta}^{(b_w,b_a)}(z_t,t; s,\xi)$; specifically, $s$ denotes the quantization scale in $Q(\cdot;s)$, and $\xi$ represents auxiliary variables utilized during calibration, such as pruning factors, grouping rules, and bit allocation strategies. A unifying design principle in recent video PTQ methodologies is systematic compensation, which seeks to minimize quantization error through refined calibration, optimized transformations, and hardware-aware quantization pathways.

\hspace*{-\parindent}\textbf{Outlier and Distribution-Shift Handling.} Although outliers and distribution shifts are not unique to PTQ, they become a central design consideration in PTQ because the pretrained weights are frozen and quantization error must be absorbed through calibration and lightweight transformations alone. For diffusion and video diffusion models, these effects are especially visible in activations and attention statistics, where heavy tails, channel imbalance, and timestep-dependent variation can sharply degrade low-bit mapping. Accordingly, a substantial subset of PTQ methods focuses on reshaping activation or weight distributions before quantization rather than treating quantization as a purely static rounding problem. LRQ-DiT~\cite{yang2025lrqditlogrotationposttrainingquantization} employs log-domain quantization with adaptive rotation, whereas TR-DQ~\cite{https://doi.org/10.48550/arxiv.2503.06564} utilizes timestep-aware smoothing alongside rotation and permutation transforms. Similarly, CLQ~\cite{liu2025clqcrosslayerguidedorthogonalbased} leverages orthogonal transforms with cross-layer guidance, and SageAttention2/3~\cite{zhang2025sageattention2efficientattentionthorough,zhang2026sageattention3microscalingfp4attention} mitigates attention outliers within INT4 and FP4-style execution paths. These approaches share a common ``transform-then-quantize'' paradigm, formulated as
\begin{equation}
\bar{X}=Q\!\left(T_x(X);s_a\right),\quad
\bar{W}=Q\!\left(T_w(W);s_w\right),\quad
\hat{Y}=\bar{X}\bar{W},\quad
T_xT_w \approx I,
\end{equation}
where $T_x(\cdot)$ and $T_w(\cdot)$ are paired transforms chosen to suppress heavy tails or channel imbalance while approximately preserving the original linear mapping.

\hspace*{-\parindent}\textbf{Timestep-Aware Quantization.} The central premise of timestep-aware PTQ is that diffusion statistics fluctuate significantly across different values of $t$, rendering a single static quantizer suboptimal. Consequently, methodologies such as TaQ-DiT~\cite{liu2024taqdittimeawarequantizationdiffusion}, DVD-Quant~\cite{https://doi.org/10.48550/arxiv.2505.18663}, Q-DiT~\cite{Chen_2025}, ViDiT-Q~\cite{https://doi.org/10.48550/arxiv.2406.02540}, S$^2$Q-VDiT~\cite{feng2025s2qvditaccuratequantizedvideo}, and TR-DQ~\cite{https://doi.org/10.48550/arxiv.2503.06564} adopt timestep-dependent calibration or bit-allocation policies. A generic formulation for this paradigm is
\begin{equation} s_t=\Pi_s(t),\qquad b_a(t)=\Pi_b(t),\qquad \min_{\Pi_s,\Pi_b}\sum_{t=1}^{S}\mathbb{E}_{z_0,z_1,t}\left[\left\Vert G_{\theta}^{(b_w,b_a(t))}(z_t,t)-(z_1-z_0)\right\Vert_2^2\right], 
\end{equation}
where $\Pi_s$ and $\Pi_b$ map each timestep to a corresponding quantization scale and activation precision, respectively. This design specifically addresses the diffusion-specific temporal distribution drift, which is a primary factor enabling PTQ to maintain efficacy for video diffusion models at low bitwidths.

\textbf{Discussion.} \rev{PTQ remains the default deployment path because it respects pretrained backbones and minimizes retraining cost. Its ceiling is set by calibration quality; without timestep-aware compensation, temporal artifacts can grow quickly in long-horizon generation.}

\subsection{VAE Compression}\label{sec:vae_compression}

\revnew{VAE compression targets a bottleneck that quantization and pruning do not directly address: the cost of video encoding/decoding itself and the latent sequence length subsequently seen by the diffusion model. In the center branch of Fig.~\ref{fig:dmd-survey-model-compression-overview}, a video encoder $E_\theta$ maps inputs into a latent representation, the diffusion generator $G_\theta$ operates on that latent sequence, and a decoder $D_\theta$ reconstructs the video afterward. Recent work follows three recurring directions. The first is compatibility-preserving video VAE design, where CV-VAE~\cite{zhao2024cvvaecompatiblevideovae} and IV-VAE~\cite{wu2025improvedvideovaelatentvideo} retain useful image-VAE priors while improving temporal compression through latent regularization, keyframe-based branching, and group-causal convolutions. The second is stronger spatiotemporal compression through architectural redesign, as in OD-VAE~\cite{chen2024odvaeomnidimensionalvideocompressor}, WF-VAE~\cite{li2024wfvaeenhancingvideovae}, and LeanVAE~\cite{cheng2025leanvaeultraefficientreconstructionvae}, which respectively emphasize omni-dimensional compression, wavelet-driven energy concentration, and lightweight reconstruction modules. The third is structure-dynamics decomposition, where VidTwin~\cite{wang2024vidtwinvideovaedecoupled}, Hi-VAE~\cite{liu2025hivaeefficientvideoautoencoding}, and DC-VideoGen~\cite{chen2025dcvideogenefficientvideogeneration} separate static structure from motion or adapt pretrained generators into a more aggressively compressed latent space.}

\textbf{Discussion.} \revnew{VAE compression is attractive because every gain is paid twice: it reduces VAE-side overhead and also shortens the latent sequence processed by the downstream generator. Its weakness is that compression error becomes upstream error for the entire pipeline; once motion, identity, or high-frequency detail is discarded in the latent space, later modules cannot easily recover it. Current evidence therefore favors methods that explicitly preserve temporal structure or adapt pretrained generators to the new latent space, rather than relying on brute-force downsampling alone.}

\subsection{Token Pruning}\label{sec:token_pruning}

\rev{Token pruning methods for video generation converge on four shared principles: importance scoring from saliency, motion, or density cues in Astraea and FastVID~\cite{liu2025astraeatokenwiseaccelerationframework,shen2025fastviddynamicdensitypruning}; dynamic rather than static retention in FullDiT2 and VGDFR~\cite{he2025fulldit2efficientincontextconditioning,yuan2025vgdfrdiffusionbasedvideogeneration}; recency-aware asymmetric compression that preserves recent or high-motion tokens in AsymRnR and long-context next-frame prediction~\cite{https://doi.org/10.48550/arxiv.2412.11706,gu2025longcontextautoregressivevideomodeling}; and merge-style compaction rather than hard dropping alone in frame context packing~\cite{zhang2025framecontextpackingdrift}. Taken together, these works suggest that token pruning is no longer just about removing tokens; the real design problem is deciding which information can be merged, delayed, or discarded without breaking motion continuity.}

\textbf{Discussion.} \rev{Token pruning remains a relatively weak direction for video diffusion. Conceptually it is close to a restrictive form of static sparse attention, so its quality ceiling is limited once important motion or identity cues are removed. In practice, gather and scatter overhead further reduces its value, which is why token pruning has so far remained less convincing than sparse attention or step distillation.}


\subsection{Model Pruning}\label{sec:model_pruning}

\rev{Model pruning in video diffusion primarily targets inference bottlenecks by removing redundant structure from a backbone $G_\theta$. As illustrated in Fig.~\ref{fig:dmd-survey-model-compression-overview} (right), this often couples pruning with a frozen teacher branch $f^\phi$ for quality recovery. In practice, pruning mainly operates at DiT-block granularity: methods rank block importance, retain a target block keep ratio $r_\ell$, and then recover quality on latent-video inputs at noise level $t$ through compensation or distillation. Representative strategies include depth-aware shallow-block pruning that preserves deeper motion-sensitive layers~\cite{wu2025individualcontentmotiondynamics}, iterative block removal with online distillation recovery~\cite{kim2025vip}, and sensitivity-guided tri-level compression where block pruning is the primary structural axis and head or feed-forward reduction is secondary~\cite{wu2025tamingdiffusiontransformerefficient}; some approaches also compress auxiliary modules such as text encoders~\cite{yan2024animatedstickersbringingstickers}. The common pattern is revealing: pruning itself is rarely the hard part, whereas identifying dispensable capacity and then repairing the resulting temporal damage is what determines whether the method is usable.}

\textbf{Discussion.} \rev{Model pruning can provide real structural speedups, but in video diffusion the main bottleneck is recovering the original generation performance and perceptual quality after pruning, rather than removing blocks itself; current evidence suggests it is most worthwhile when sensitivity-guided pruning is paired with distillation-based quality repair~\cite{kim2025vip,wu2025tamingdiffusiontransformerefficient}.}

\subsection{Summary and Outlook}

Model compression is indispensable for deployment, but the key issue is preserving temporal consistency and perceptual quality rather than maximizing compression ratio. In video diffusion, quantization is the most deployment-ready backbone-level strategy because it usually needs only light calibration or compensation, whereas \revnew{VAE compression attacks system cost earlier by reducing encode/decode overhead and latent sequence length.} By contrast, pruning often requires expensive post-training recovery to offset capacity loss and motion degradation. Accordingly, pruning remains less common than quantization or well-designed latent compression.

Across these branches, the recurring success pattern is compensation or adaptation rather than compression alone. Future progress depends on more stable and targeted recovery, including timestep-aware calibration for PTQ, training-time noise absorption for QAT, \revnew{latent-space compatibility or motion-preserving reconstruction for VAE compression,} and distillation- or sensitivity-guided recovery for pruning. The goal is to make compression a controlled optimization process with predictable quality loss.

Compression also should not be optimized in isolation. In video diffusion, realized gains depend jointly on memory traffic, kernel support, attention cost, \revnew{latent sequence length,} and cross-module error accumulation, so quantization, \revnew{VAE compression,} and pruning will be most useful when co-designed with sparse attention, step distillation, and cache reuse. A practical direction is deployment-oriented joint design under explicit runtime, memory, and quality constraints.

\section{Cache and Trajectory Optimization}\label{sec:cache_and_trajectory}

\begin{figure*}[t]
    \centering
    \includegraphics[width=\linewidth]{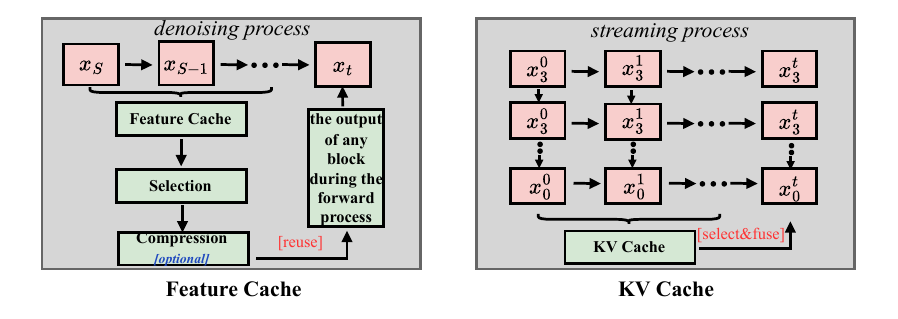}
        \vspace{-30pt}
    \caption{Overview of cache and trajectory optimization methods for video diffusion acceleration. The framework integrates feature and KV reuse, denoising-state and trajectory redesign, and system-level execution strategies to improve long-horizon efficiency.}
    \label{fig:dmd-survey-cache-trajectory-overview}
        \vspace{-10pt}
\end{figure*}

Cache and trajectory optimization are grouped together because both intervene directly in denoising execution logic rather than changing model parameters. Cache methods reuse features, KV states, or intermediate activations computed at previous denoising steps, effectively carrying historical computation forward into the current update as a momentum-like state reuse mechanism. Trajectory methods instead alter how denoising proceeds, for example through zigzag rollouts, coarse-to-fine schedules, or explicit noise/state corrections that change the latent evolution path itself. Fig.~\ref{fig:dmd-survey-cache-trajectory-overview} summarizes this shared paradigm across feature cache, KV cache, noise/state modification, trajectory modification, parallel computation, and other system-level methods, each targeting throughput and memory scalability by reducing redundant denoising work from a different interface.

\subsection{Feature Cache}\label{sec:feature_cache}
Feature cache addresses a fundamental inefficiency in diffusion inference: as illustrated in the left panel of Fig.~\ref{fig:dmd-survey-cache-trajectory-overview}, intermediate features computed at earlier denoising steps can be stored and reused at later steps when adjacent steps remain sufficiently similar, instead of being recomputed end-to-end every time. This redundancy is particularly costly in long-video and high-resolution generation tasks, where repeated per-step evaluation quickly increases latency and memory-bandwidth pressure. A generic reuse rule can be written as
\begin{equation}
\tilde{F}_t^{\ell}=g_t^{\ell}C_{t-\Delta}^{\ell}+(1-g_t^{\ell})F_t^{\ell},\quad
g_t^{\ell}=\mathbb{I}\!\left(d(z_t,z_{t-\Delta})<\tau\right)\ \text{or}\ \mathbb{I}(t\bmod k\neq 0),
\end{equation}
\rev{where $F_t^{\ell}$ denotes the newly computed feature at layer $\ell$ and denoising step $t$, $C_{t-\Delta}^{\ell}$ is the cached feature from the same layer at an earlier step $t-\Delta$, $\Delta$ is the cache look-back interval, and $\tilde{F}_t^{\ell}$ is the feature actually fed to the subsequent computation. Here, $g_t^{\ell}\in\{0,1\}$ is a binary reuse gate, $z_t$ and $z_{t-\Delta}$ are the latent states at steps $t$ and $t-\Delta$, $d(\cdot,\cdot)$ is a similarity metric, $\tau$ is a reuse threshold, and $k$ is a refresh interval. When $g_t^{\ell}=1$, the cached feature is reused and the layer is skipped; otherwise the feature is recomputed and the cache refreshed. Because naive skipping often introduces temporal drift, recent methods instead treat feature cache as a controlled optimization problem that maximizes reuse while preserving temporal coherence and visual fidelity~\cite{sun2024unveilingredundancydiffusiontransformers,liu2025fastcachefastcachingdiffusion,zheng2025letfeaturesdecidesolvers,zheng2025forecastcalibratefeaturecaching,peng2025ertacacheerrorrectificationtimesteps,cui2025bwcacheacceleratingvideodiffusion,ma2025magcachefastvideogeneration}. Their pipelines are largely organized around two questions: how to score reuse versus refresh, and how to reuse cached states once a hit is declared. This already indicates the main trend of the area: the field has moved away from naive skipping and toward explicit drift management.}

\hspace*{-\parindent}\textbf{Importance Scoring for Reuse and Refresh.} Feature cache scoring mechanisms can be grouped into three approaches: similarity-based scoring, which measures inter-step feature stability using indicators such as cosine/L1 distance, residual change, or adjacent-step consistency~\cite{sun2024unveilingredundancydiffusiontransformers,https://doi.org/10.48550/arxiv.2410.19355,https://doi.org/10.48550/arxiv.2411.02397,wei2025mixcachemixtureofcachevideodiffusion,chen2026sortblocksimilarityawarefeaturereuse}; forecast-error-based scoring, which predicts future features and triggers refresh when the prediction deviates from the realized state~\cite{liu2025reusingforecastingacceleratingdiffusion,zheng2025forecastcalibratefeaturecaching,peng2025ertacacheerrorrectificationtimesteps,bu2025dicacheletdiffusionmodel}; and schedule/prior-based scoring, which relies on timestep priors, fixed refresh intervals, stage-wise heuristics, or predefined cache schedules rather than expensive online search~\cite{https://doi.org/10.48550/arxiv.2510.05367,https://doi.org/10.48550/arxiv.2408.12588,liu2025fastcachefastcachingdiffusion}.

\hspace*{-\parindent}\textbf{Cache Reuse and Refresh Mechanisms.} Feature cache operates predominantly across three execution modes. \textbf{(1)} Full reuse after a cache hit directly applies cached states to bypass skipped steps, reverting to full recalculation only when explicit trigger conditions are met~\cite{ma2025magcachefastvideogeneration,https://doi.org/10.48550/arxiv.2510.05367,https://doi.org/10.48550/arxiv.2410.19355,cui2025bwcacheacceleratingvideodiffusion,peng2025ertacacheerrorrectificationtimesteps}. \textbf{(2)} Partial refresh selectively updates specific layers, blocks, or modules while retaining the remaining cached states, thereby circumventing the overhead of complete recomputation~\cite{https://doi.org/10.48550/arxiv.2510.05367,https://doi.org/10.48550/arxiv.2408.12588,https://doi.org/10.48550/arxiv.2411.02397,cui2025bwcacheacceleratingvideodiffusion,chen2026sortblocksimilarityawarefeaturereuse,bu2025dicacheletdiffusionmodel}. \textbf{(3)} Error-compensated reuse initially deploys cached features, either entirely or partially, and subsequently applies explicit correction, calibration, or trajectory alignment to mitigate accumulated temporal drift~\cite{https://doi.org/10.48550/arxiv.2410.19355,liu2025reusingforecastingacceleratingdiffusion,zheng2025forecastcalibratefeaturecaching,peng2025ertacacheerrorrectificationtimesteps,bu2025dicacheletdiffusionmodel,chen2026sortblocksimilarityawarefeaturereuse}. Notably, recent frameworks increasingly gravitate towards hybrid approaches, integrating selective refreshing with error compensation to simultaneously optimize inference throughput and visual generation quality.

Feature cache is practical because it exploits inter-step redundancy with limited architectural change, but \rev{naive reuse is brittle. Adaptive refresh and explicit correction are therefore essential if it is to remain reliable in long-horizon generation.}

\subsection{KV Cache}\label{sec:kv_cache}

KV cache stores historical key-value (KV) tensors $\{K,V\}$ from previous generation steps and reuses them during subsequent inference, eliminating the need for the model to repeatedly calculate attention states already captured in the context. In streaming or autoregressive video diffusion, the current chunk $z_t^i$ attends to historical keys and values through
\begin{equation}
O_i=\text{Softmax}\left(\frac{Q_i[K_{<i};K_i]^\top}{\sqrt{d}}\right)[V_{<i};V_i].
\end{equation}
As history grows, KV memory and attention cost increase rapidly, so practical systems often compress the full history $\{K_{0:t},V_{0:t}\}$ into a smaller set $\{\tilde{K},\tilde{V}\}$. A common strategy is token selection based on an importance score $\mathcal{I}$,
\begin{equation}
\mathcal{S}_{\text{KV}}=\text{Top}_{C^{\text{KV}}_\text{size}}(\mathcal{I}),\quad
\tilde{K}=\{K_i\mid i\in\mathcal{S}_{\text{KV}}\},\quad
\tilde{V}=\{V_i\mid i\in\mathcal{S}_{\text{KV}}\}.
\end{equation}
In long-video generation, this directly addresses a critical bottleneck: the rapid escalation of latency and memory consumption that occurs when each new data block requires a complete historical record for attention computation. Most KV cache methods are built upon causal video generation pipelines, leveraging autoregressive temporal dependencies to reuse cached representations and curtail memory overhead~\cite{https://doi.org/10.48550/arxiv.2411.16375,https://doi.org/10.48550/arxiv.2406.10981,xing2024live2difflivestreamtranslation,ye2025yanfoundationalinteractivevideo,feng2025vidarcembodiedvideodiffusion,kong2025causnvsautoregressivemultiviewdiffusion,yu2025videomarautoregressivevideogeneratio}. A notable exception is PoseGen~\cite{he2025posegenincontextlorafinetuning}, which employs cross-segment background key-value sharing to ensure stitching consistency, as opposed to standard sliding-window caching mechanisms.

\rev{Advancements in KV cache converge on a unified objective: reducing effective cache length and attention cost while preserving long-range consistency. These methods can be grouped into three paradigms: budget windows and eviction, which constrain cache size through sliding windows or stale-state removal~\cite{https://doi.org/10.48550/arxiv.2411.16375,https://doi.org/10.48550/arxiv.2406.10981,xing2024live2difflivestreamtranslation,ye2025yanfoundationalinteractivevideo,feng2025vidarcembodiedvideodiffusion,kong2025causnvsautoregressivemultiviewdiffusion,wang2025liftvsrliftingimagediffusion}; compressed memory KV, which summarizes history into compact memory tokens or latent summaries~\cite{jiang2025lovicefficientlongvideo,zhang2026pretrainingframepreservationautoregressive,pfp,li2026packcachetrainingfreeaccelerationmethod,ji2025memflowflowingadaptivememory}; and selective KV, which activates historical states by relevance or importance~\cite{zhang2025egolcdegocentricvideogeneration,ji2025memflowflowingadaptivememory,li2026packcachetrainingfreeaccelerationmethod}. The trend is from storing everything toward selective memory, but the hard part remains deciding what can be forgotten without causing identity drift later.}

\textbf{Discussion.} \rev{In video diffusion, KV cache is most valuable when generation proceeds chunk by chunk or frame block by frame block, because each new chunk must attend to an expanding temporal history whose cost grows with video length. The core difficulty is therefore video-specific: compressing or evicting historical KV states without losing motion continuity, object identity, scene layout, or cross-chunk stitching consistency over long videos. Poor cache decisions may not fail immediately, but can accumulate into delayed drift or visible discontinuities many chunks later.}

\paragraph{Noise and State Modification}\label{sec:noise_state_modification}
\rev{As a complementary direction, noise and state modification covers methods that intervene locally on the denoising trajectory rather than redesigning the full path. The modified object is typically the starting noise, an intermediate latent/state, or a local state-transition operator at particular steps, while the global timestep schedule remains largely unchanged. Under this definition, FreeNoise~\cite{https://doi.org/10.48550/arxiv.2310.15169} reshapes the initial noise assignment and applies window-based fusion for long-video consistency, TokenFlow~\cite{https://doi.org/10.48550/arxiv.2307.10373} propagates edited diffusion states across frames via correspondence-guided token replacement, Latent-Shift~\cite{https://doi.org/10.48550/arxiv.2304.08477} injects temporal coupling through parameter-free channel shifts inside latent residual branches, and Free4D~\cite{liu2025free4dtuningfree4dscene} performs latent fusion and reference-state replacement during denoising. These methods are therefore best understood as pointwise or local interventions on the trajectory: they edit what the denoiser starts from or carries at selected positions, rather than replanning the whole denoising route.}

\textbf{Discussion.} \rev{This category is attractive because local trajectory interventions are often lightweight and training-free. Its limitation is narrow operating range: once the handcrafted noise or state edit no longer matches the scene dynamics or task setting, quality can degrade quickly.}

\paragraph{Trajectory Modification}\label{sec:trajectory_modification}
\rev{Trajectory modification is another complementary route, but it acts at the global-path level. Instead of editing local states at isolated points, these methods reallocate where computation is spent along the sampling route, either by changing timestep placement or by decomposing generation into stages with different costs. Current video methods mainly fall into two groups: step/schedule trajectory designs, which optimize timestep placement, leap-style skipping, local transition approximation, or adaptive sub-stepping~\cite{sabour2024alignstepsoptimizingsampling,zhu2025acceleratingdiffusionsamplingexploiting,kim2025ondevicesoraenablingtrainingfree,cheng2025adaptivebeginofvideotokensautoregressive,liu2025multimotion,zhuang2025timestepmasterasymmetricalmixture}; and multi-stage path redesigns, which replace a single-path rollout with hierarchical stages such as content-motion decomposition, high-low-high resolution sampling, or coarse-to-fine refinement~\cite{https://doi.org/10.48550/arxiv.2403.14148,https://doi.org/10.48550/arxiv.2503.18940,https://doi.org/10.48550/arxiv.2410.05954,https://doi.org/10.48550/arxiv.2502.05179,li2025arlonboostingdiffusiontransformers,cheng2025srdiffusionacceleratevideodiffusion}. These methods are most convincing when they expose genuine denoising redundancy at the path level rather than merely handcrafting a shorter schedule.}

\textbf{Discussion.} \rev{Trajectory modification can deliver larger gains than local state edits because it changes how the whole denoising budget is allocated. Its risk is global miscalibration: once the redesigned path becomes too aggressive, temporal artifacts and quality loss can accumulate across the entire rollout rather than only at one local intervention point.}

\paragraph{Parallel Computation}\label{sec:parallel_computation}
\rev{Parallel computation is a further complementary system route that replaces strictly serial execution with concurrent execution. In video generation, representative implementations include parallel denoising across temporal chunks, multi-GPU inference with communication-computation overlap, and decomposition of generation into concurrent subproblems~\cite{Zhang_2023,https://doi.org/10.48550/arxiv.2305.13077,bandyopadhyay2025blockcascadingtrainingfree,chen2025dbspacceleratingsparseattention,wang2025pipeditacceleratingdiffusiontransformers,fang2024xditinferenceenginediffusion}. DiffCollage~\cite{Zhang_2023} factorizes large content into overlapping local factors, ControlVideo~\cite{https://doi.org/10.48550/arxiv.2305.13077} combines cross-frame parallel attention with hierarchical sampling, Block Cascading~\cite{bandyopadhyay2025blockcascadingtrainingfree} parallelizes temporal blocks, db-SP~\cite{chen2025dbspacceleratingsparseattention} parallelizes sparse attention, PipeDiT~\cite{wang2025pipeditacceleratingdiffusiontransformers} pipelines denoising and decoding across GPUs, and xDiT~\cite{fang2024xditinferenceenginediffusion} unifies sequence, patch, and CFG parallelism. The field has learned that parallelism is easy to claim but hard to realize; real speedup depends on overlap and scheduling rather than concurrency alone.}

\textbf{Discussion.} \rev{Parallel computation turns latent concurrency into throughput gains, especially in multi-GPU settings. In practice, communication overlap and scheduling quality matter more than parallelism, which is why many designs underdeliver.}

\paragraph{Other Efficiency Methods}\label{sec:other_efficiency_methods}
\rev{Other efficiency methods collect residual but still relevant system bottlenecks outside the main cache-centric thread. They target auxiliary overheads such as long-context memory management and VAE-side reconstruction cost~\cite{Shrivastava_2024,cheng2025leanvaeultraefficientreconstructionvae,chen2025dcvideogenefficientvideogeneration}. Shrivastava et al.~\cite{Shrivastava_2024} optimize long-horizon generation through differential temporal-cache compression, LeanVAE~\cite{cheng2025leanvaeultraefficientreconstructionvae} redesigns the reconstruction VAE to reduce encoding and decoding cost, and DC-VideoGen~\cite{chen2025dcvideogenefficientvideogeneration} accelerates inference with a deep-compression video autoencoder that lowers latent token count. These methods can still affect end-to-end performance because systems are often bottlenecked outside the sampler itself, so ignoring auxiliary modules can offset part of the gain from core acceleration.}

\textbf{Discussion.} \rev{Other efficiency methods can still affect end-to-end latency when bottlenecks lie outside the core sampler. Their impact is strongest when jointly tuned with mainstream acceleration methods rather than treated as isolated optimizations.}

\subsection{Summary and Outlook}

Viewed from this perspective, the main thread of this section is cache management, especially feature reuse and KV reuse, while the remaining paragraph-level directions provide complementary ways to intervene in denoising execution. Cache methods decide what past computation is carried forward into later updates, whereas trajectory edits, parallel execution, and auxiliary system optimizations adjust how subsequent denoising work is scheduled or implemented. Their shared strength is that they reduce redundant work by intervening in the execution path itself rather than changing model weights. Their shared risk is long-horizon error accumulation, where stale cache states or miscalibrated execution heuristics gradually propagate through temporal dependencies.

Future work should focus on closed-loop cache management rather than static reuse rules. For video diffusion, refresh and eviction decisions should be tied to the denoising trajectory itself: later steps may tolerate aggressive reuse, while high-motion segments, scene cuts, or rapid viewpoint changes may require earlier refresh to prevent stale features from being propagated across frames. A promising direction is to estimate cache reliability online and couple it with drift control, so the system can decide when to reuse, partially refresh, or fully recompute states before temporal errors accumulate into visible flicker, identity shift, or motion inconsistency.

A second direction is tighter cache-trajectory co-design. Rollout scheduling, step skipping, and state extrapolation are usually treated separately from cache policy, but in practice they interact strongly: a more aggressive trajectory schedule changes how long cached features remain valid, and cache reuse in turn affects which rollout patterns are safe. Future methods should therefore optimize cache lifetime, refresh points, and rollout schedules jointly, under explicit latency-memory-quality budgets, instead of tuning them as isolated heuristics. This is especially important for long videos, where poor cache-trajectory interaction can turn small local approximation errors into persistent temporal drift.

\section{Open Problems and Future Directions}\label{sec:future_direction}

In this section, we separate current unresolved issues from longer-term future directions. The first subsection focuses on bottlenecks that are already exposed by existing acceleration methods, such as unclear regime-dependent trade-offs, composite-acceleration instability, and hardware-execution mismatch. The second subsection then discusses forward-looking directions, including real-time generation, open training datasets for acceleration, and the limits of efficient attention.

\subsection{Current Technical Issues in Accelerated Video Diffusion}

\subsubsection{Trade-offs of Acceleration Schemes}

The first unresolved issue is that the field still lacks a regime-aware understanding of acceleration trade-offs. Existing results suggest that acceleration in video diffusion is not governed by a single Pareto frontier, but by several deployment-dependent trade-offs across latency, memory, implementation complexity, and output fidelity. Step distillation delivers the largest reduction in sampling steps, but often leaves less room for later reuse-based methods; sparse attention improves the dominant attention bottleneck, but its realized gains depend heavily on token patterns and kernel efficiency; quantization and caching can be effective, yet their benefits are often shaped by hardware support and temporal error accumulation. The current problem is therefore not simply to rank methods in isolation, but to characterize their comparative advantages under concrete regimes such as offline generation, few-step high-quality synthesis, and real-time streaming.

Regarding efficient attention mechanisms, sparse attention currently dominates over linear attention. Most algorithms claiming to employ linear attention fail to achieve true linear complexity in practice. This limitation arises because video tokens cannot be compressed to extreme degrees; retaining certain important tokens remains necessary to ensure generation quality. Consequently, linear attention is frequently utilized as an error compensation mechanism for tokens discarded by sparse attention, as exemplified by SLA~\cite{https://doi.org/10.48550/arxiv.2509.24006}.

Token pruning and model pruning remain relatively unpopular compared to step distillation and efficient attention. Post-training pruning incurs prohibitive computational costs, while training-free approaches result in severe degradation of generated video quality. Furthermore, their speedup benefits generally fall short of those achieved by attention optimization and step distillation, particularly for token pruning. Despite substantial theoretical speedup potential, practical gains remain limited due to the overhead of token splitting and merging operations and the non-proportional relationship between token reduction and speedup in MLP (multilayer perceptron) and attention computations.

Model quantization remains technically attractive, but its unresolved difficulty lies in turning nominal low-bit compression into reliable end-to-end speedup. In practice, its focus often centers on attention acceleration, given that attention constitutes the most computationally expensive module in video DiTs. Merely accelerating the linear layers for computing $Q$, $K$, $V$ projections and MLP layers yields limited practical benefits. Moreover, this domain presents substantial technical barriers; achieving tangible acceleration typically requires expertise in Triton~\cite{triton}, TileLang~\cite{tilelang}, and even CuDNN for implementing low-level custom operators.

Feature cache also provides substantial speedup potential, though its objective conflicts with step distillation. While it can be applied to few-step generators following step distillation for additional acceleration, many feature cache algorithms induce performance degradation in such scenarios. This exposes a concrete unresolved issue: current cache policies are still poorly matched to the low-step, high-sensitivity setting created by step distillation.

KV cache addresses the challenges of excessive memory consumption and the large number of tokens involved in attention calculations during streaming generation. This area holds particular importance given the critical role of real-time generation in video synthesis. Well-designed KV cache compression or selection algorithms significantly advance the development of real-time generation capabilities.

\subsubsection{Quality Preservation in Composite Acceleration}

While composite acceleration achieves superior speedup ratios compared to single-method approaches, it often induces a more pronounced degradation in model performance. The issue is not only that each component introduces approximation error, but that these errors interact across modules in ways that are still poorly understood. Common paradigms for these integrated approaches include pairing step distillation with sparse attention, quantization with sparse attention, step distillation with feature caching, or model pruning with feature caching.

Representative methods reveal several recurring design patterns for addressing this challenge. In step-distillation-plus-sparsity pipelines, BLADE~\cite{gu2025blade} integrates adaptive block-sparse attention directly into a sparsity-aware step-distillation objective; this ensures the student is optimized under the same sparse computation regime used at inference, rather than inheriting a mismatch between dense training and sparse deployment. Similarly, Efficient-vDiT~\cite{https://doi.org/10.48550/arxiv.2502.06155} and FlashVSR~\cite{https://doi.org/10.48550/arxiv.2510.12747} employ staged distillation pipelines alongside sparse attention modules to progressively absorb the approximation errors introduced by sparsification, avoiding the instability of applying all compression operations simultaneously.

For quantization-plus-sparsity co-acceleration, FPSAttention~\cite{https://doi.org/10.48550/arxiv.2506.04648} and QuantSparse~\cite{https://doi.org/10.48550/arxiv.2509.23681} circumvent naive stacking through joint co-design mechanisms: FPSAttention aligns FP8 quantization and sparsity under a unified tile-wise granularity and denoising-step-aware strategy, while QuantSparse~\cite{https://doi.org/10.48550/arxiv.2509.23681} combines low-bit compression with attention sparsification via training-time compensation, such as distillation-style saliency preservation and sparse-aware reparameterization, to reduce compounded approximation bias.

In caching-plus-pruning pipelines, UniCP~\cite{Sun_2025} treats reuse and token reduction as a unified error-management problem rather than as two independent heuristics, introducing error-aware cache-window control and dynamic correction mechanisms to suppress temporal drift caused by stale cached states and aggressive pruning. More generally, these methods suggest that quality preservation in composite acceleration depends less on the individual efficacy of each component and more on the presence of a coupling-aware compensation strategy, such as joint optimization, staged adaptation, or explicit error correction, to regulate the accumulation of cross-module approximation errors across denoising steps and temporal dependencies.

The unresolved issue behind these results is the absence of principled composite-acceleration design grounded in explicit error budgeting. Current hybrid methods indicate that the dominant failure mode is not the isolated approximation error of a single component, but rather the interaction terms across modules, such as sparsity-induced feature distortion amplified by low-bit quantization, or cache staleness magnified by pruning. What is still missing is a framework that exposes intermediate error signals and allocates tolerances across components, potentially by jointly scheduling sparsity ratios, bit-widths, cache refresh rates, and pruning strengths according to timestep sensitivity and feature dynamics. Likewise, compensation remains under-formalized: cross-module consistency distillation, residual error prediction, and timestep-aware correction branches exist as promising design motifs, but not yet as a unified principle for controlling coupled approximation bias.

\subsubsection{Hardware Co-design in Acceleration Algorithms}

Hardware co-design remains a central open challenge in accelerated sampling for video diffusion models, particularly for attention- and quantization-based methods. Unlike trajectory scheduling or high-level caching policies, the practical gains of efficient attention and low-precision computation are highly sensitive to kernel implementation, memory layout, synchronization overhead, and hardware-specific operator support. Consequently, a method reporting significant algorithmic savings—such as reduced attention density or lower bit-widths—may still yield limited end-to-end speedup if its execution pattern is irregular, its data movement dominates runtime, or its precision format lacks efficient hardware support. This gap is especially pronounced in video diffusion, where long sequences, spatiotemporal attention, and iterative denoising amplify memory-bandwidth pressure and kernel launch overhead.

For efficient attention, the primary challenge is the mismatch between algorithmic sparsity and hardware execution. While semantic-based routing often yields irregular, input-dependent masks, hardware efficiency requires regular block structures and contiguous memory access. Consequently, sparse attention must be evaluated not only by approximation error but by its kernel-level feasibility and scheduling overhead~\cite{sparge_attn,qiao2025flashomniunifiedsparseattention,chen2025dbspacceleratingsparseattention,wang2025pipeditacceleratingdiffusiontransformers,fang2024xditinferenceenginediffusion}. The unresolved issue is that current routing objectives are usually optimized for semantic fidelity alone, without an equally explicit constraint on hardware-amenable execution patterns such as regular block structure and stable throughput.

Regarding quantization, realized speedups depend more on precision formats and memory alignment than nominal bit-widths. In video diffusion, timestep-dependent statistics often turn scale loading and format conversion into primary bottlenecks. Current hardware-level tensions, such as the trade-off between fine-grained scaling accuracy and metadata traffic, remain largely unresolved~\cite{zhang2025sageattentionaccurate8bitattention,zhang2025sageattention2efficientattentionthorough,zhang2026sageattention3microscalingfp4attention,https://doi.org/10.48550/arxiv.2506.04648}. The issue is that bit allocation, scaling granularity, and kernel fusion are still rarely optimized jointly under explicit runtime and memory constraints, which leaves portability and realized performance unstable across accelerators.

\subsection{Future Directions of Accelerated Sampling of Video Diffusion Models}

\subsubsection{The Development Prospects of Real-Time Generation}

Real-time generation is becoming the most consequential direction for accelerated video diffusion, because practical deployment requires not only high visual fidelity but also interactive latency and sustained throughput. As shown in Fig.~\ref{fig:dmd-survey-trend}, real-time-oriented acceleration has exhibited the fastest growth in the most recent year among major acceleration branches, indicating a clear shift from offline synthesis toward online, user-in-the-loop generation. This trend is further reflected in emerging real-time systems such as Krea Realtime 14B~\cite{krea_realtime_14b} and Live Avatar~\cite{huang2025live}. Together, these developments show that real-time capability is not a marginal optimization target but a core requirement for broad applications, including interactive creation, digital humans, live broadcasting, telepresence, and embodied/world-model interfaces.

Real-time video diffusion still faces three interconnected bottlenecks. First, error accumulation remains the primary failure mode in long-term causal generation, where biases from local chunk synthesis propagate over time; representative solutions include proactive error injection, online correction paradigms, and revising positional encodings~\cite{li2025stable,huang2025live,https://doi.org/10.48550/arxiv.2506.08009,lu2025reward,Yin_2025,yin2024slow,yesiltepe2025infinity}. Second, multi-shot transitions are difficult to maintain as scene boundaries exacerbate temporal discontinuities and identity drift; practical remedies include KV cache refreshing, agent-based dynamic shot splitting, and cross-shot feature anchoring~\cite{https://doi.org/10.48550/arxiv.2509.22622,liang2025lookingbackwardstreamingvideotovideo,yi2025magicinfinitegeneratinginfinitetalking,yu2025autorefinerimprovingautoregressivevideo}. Third, an infinitely growing KV cache threatens memory scalability during long-form generation, necessitating bounded memory designs such as rolling-window eviction, global memory compression, receiver-tag aggregation, and memory-generator decoupling~\cite{zhang2026pretrainingframepreservationautoregressive,pfp,yu2025videossmautoregressivelongvideo,zhu2025memorizeandgeneratelongtermconsistencyrealtime,jiang2025lovicefficientlongvideo,li2026packcachetrainingfreeaccelerationmethod}. Ultimately, future real-time systems will require the joint optimization of these dimensions, as addressing any single bottleneck in isolation is insufficient for robust deployment.

In conclusion, addressing the core challenges of error accumulation, multi-view consistency, and latency in real-time interaction is essential for advancing downstream applications, such as world models for game engines or personalized interactive avatars. Consequently, exploring alternatives and extensions beyond the current self-forcing paradigm may facilitate more robust causal video generation and efficient distillation algorithms, especially under long-horizon or highly dynamic scenarios.

\subsubsection{Open Training Datasets for Acceleration}

The scarcity of high-quality open training datasets remains a structural bottleneck for acceleration research in video diffusion. While contemporary methods increasingly optimize sampling, attention, quantization, and caching, their empirical gains remain inherently tied to the data quality during fine-tuning or distillation. Prevalent open datasets, such as WebVid-10M~\cite{webvid10m}, OpenVid-1M~\cite{nan2024openvid}, UCF101~\cite{soomro2012ucf101}, Mixkit~\cite{Mixkit}, and SkyTimelapse~\cite{SkyTimelapse}, exhibit critical deficiencies in at least one of four dimensions: native resolution, motion fidelity, shot-transition continuity, or taxonomic diversity. These shortcomings disproportionately affect accelerated models, as few-step or compressed pipelines are significantly more sensitive to temporal artifacts and supervisory noise than their full-step counterparts.

A secondary challenge lies in dataset privatization. Our analysis reveals a heavy reliance on proprietary datasets; the most competitive industrial systems are typically trained on non-public corpora, which hinders reproducibility and complicates fair benchmarking. Concurrently, large-scale web crawling introduces unresolved copyright, privacy, and compliance risks, particularly regarding identity-centric or user-generated content. Consequently, the field must pivot from merely seeking ``larger open data'' toward developing acceleration-oriented open data infrastructure. This includes standardized curation protocols for motion and shot continuity, transparent provenance and licensing, and privacy-preserving release mechanisms. Furthermore, synthetic trajectory datasets and data distillation pipelines offer a controllable and legally robust alternative to raw web-scale scraping, providing a viable path for training next-generation accelerated video diffusion models.

\subsubsection{The Upper Limits of Efficient Sparse Attention Mechanisms}

Sparse attention is likely the dominant path for practical acceleration in video diffusion because it provides a better quality-speed trade-off than pure linearization. In long video generation, most informative interactions are concentrated on a small subset of tokens, so structured sparsification can remove redundant computation while preserving critical motion and identity cues~\cite{sparge_attn,https://doi.org/10.48550/arxiv.2505.18809,li2025radialattentiononlogn}. By contrast, fully linear attention often relies on aggressive low-rank or kernelized compression that can over-smooth long-range dependencies and weaken temporal fidelity in high-motion scenes. A more reliable direction is to treat linear attention as a secondary compensation term after sparse routing, for example as a cheap residual branch for pruned low-importance tokens, rather than as the primary backbone for all token interactions.

The ecosystem evidence is also moving in this direction. On the open-source and research side, sparse-attention-driven solutions are now widespread across training-free and training-based pipelines, including SpargeAttention~\cite{sparge_attn}, FlashOmni~\cite{qiao2025flashomniunifiedsparseattention}, VORTA~\cite{https://doi.org/10.48550/arxiv.2505.18809}, Efficient-vDiT~\cite{https://doi.org/10.48550/arxiv.2502.06155}, and block-sparse distillation frameworks such as BLADE~\cite{gu2025blade}. This suggests that sparse attention is evolving from an isolated optimization trick toward a core systems component for deployable video generation.

For future progress, three technical fronts are most important. \textbf{First}, routing quality and hardware executability must be co-optimized, so semantic token selection is aligned with tile-friendly kernels and balanced GPU workloads. \textbf{Second}, sparsity must be long-horizon-aware, with explicit anti-drift mechanisms across timesteps and shots, instead of static masks optimized only for short clips. \textbf{Third}, sparse attention should be integrated with complementary modules, including cache management, quantization, and lightweight residual compensation for dropped-token reconstruction. If these problems are solved, sparse attention can become the foundation of minute-scale real-time generation, enabling lower latency, lower memory, and higher temporal consistency at the same time, while keeping quality close to dense baselines.

\subsubsection{The Potential of MoE Acceleration in Video Generation}

\revnew{A further future direction is the possible transition from dense video generators to mixture-of-experts (MoE) video backbones as model scale continues to grow. If video generation follows the same route as LLMs, acceleration will need to address not only sampling and attention, but also MoE-specific bottlenecks such as expert routing overhead, all-to-all communication, load imbalance across experts, expert placement under hybrid parallelism, and the memory cost of large expert pools. In that regime, several acceleration strategies developed for LLM MoE systems may become relevant to video generation, including hybrid dense-MoE parallelism, communication-computation overlap, load-balanced routing, expert prefetch and caching, and expert-aware quantization or pruning. This would introduce a new systems frontier for video diffusion acceleration, where the central challenge is no longer only how to accelerate a dense generator, but how to preserve video quality while making sparse expert activation efficient at scale.}

\section{Summary}\label{sec:conclusion}

In this survey, we presented a comprehensive review of efficient video diffusion models, covering the full technical landscape from step distillation and efficient attention to model compression and cache/trajectory optimization. We formalized the core principles, analyzed method evolution and trend shift, and highlighted how acceleration methods differ in reducing NFE versus lowering per-step overhead. Looking ahead, we explicitly distinguish between the critical challenges and promising future directions. We identify key challenges centered on unresolved issues, such as the compounding quality degradation in composite-acceleration pipelines, the persistent gap between algorithmic sparsity and hardware execution efficiency, and the exposure bias that compromises robustness in real-time streaming. We hope this work provides both a structured foundation and a practical roadmap for future research toward efficient, reliable, and deployable video diffusion generation.


\putbib
\end{bibunit}

\clearpage
\begin{bibunit}
\appendix

\section{Supplementary Method Profiles}\label{app:method_profiles}

This appendix consolidates representative method-level profiles that were moved out of the main manuscript to improve readability. Each subsection below corresponds to one main manuscript technical section and preserves the original grouping logic.

\subsection{Step Distillation: Representative Methods}\label{app:method_step_distillation}

This subsection introduces representative methods under step distillation in Section~\ref{sec:step_distillation}.

\subsubsection{Consistency Distillation}

This subsubsection highlights representative methods under consistency distillation.

\hspace*{-\parindent}\textbf{$\blacktriangleright$ VideoLCM.} VideoLCM~\cite{https://doi.org/10.48550/arxiv.2312.09109} adapts LCM to the video domain by inserting temporal mixing layers into the pre-trained image backbone. In this framework, $M(\cdot)$ acts as the identity function $\mathbb{I}$ applied to the video latents, treating the entire spatiotemporal volume uniformly.

\hspace*{-\parindent}\textbf{$\blacktriangleright$ MCM.} MCM~\cite{https://doi.org/10.48550/arxiv.2406.06890} addresses the trade-off between motion smoothness and spatial quality by disentangling these attributes. Here, $M(\cdot)$ serves as a motion extractor, ensuring that the consistency loss specifically constrains the motion representation. To mitigate spatial degradation stemming from low-quality video data, MCM introduces an independent image adversarial loss utilizing high-quality images, ensuring that synthesized videos retain spatial fidelity during motion distillation:
\begin{equation}
\begin{split}
\mathcal{L}_{\text{MCM}}^{G} &= -\mathbb{E}_{\epsilon}\frac{1}{F}\sum_i\big[\log D_{\psi}(G_\theta(\epsilon,1)_i)\big], \\
\mathcal{L}_{\text{MCM}}^{D} &= -\mathbb{E}_{z_0}\frac{1}{F}\sum_i\big[\log D_{\psi}((z_0)_i)\big] - \mathbb{E}_{\epsilon}\frac{1}{F}\sum_i\big[\log(1-D_{\psi}(G_\theta(\epsilon,1))_i)\big],
\end{split}
\end{equation}
where $F$ denotes the number of video frames. In contrast to Eq.~\ref{eq:baseline_gan_loss}, which treats the entire video volume as the unit for adversarial learning, this formulation applies adversarial supervision at the individual frame level.

\hspace*{-\parindent}\textbf{$\blacktriangleright$ DCM.} DCM~\cite{https://doi.org/10.48550/arxiv.2506.03123} mitigates optimization interference by decoupling the generation process. It partitions the ordinary differential equation (ODE) trajectory into two stages managed by distinct experts: a semantic expert responsible for layout and motion, and a detail expert dedicated to refinement. In this framework, $M(\cdot)$ denotes the temporal difference of video latent variables (i.e., $G_\theta(\epsilon,1)_{[1:L]}-G_\theta(\epsilon,1)_{[0:L-1]}$), constraining the student to capture motion amplitudes identical to those of the teacher.

\hspace*{-\parindent}\textbf{$\blacktriangleright$ DanceLCM.} DanceLCM~\cite{https://doi.org/10.48550/arxiv.2504.11143} focuses on human animation and employs a segmented trajectory distillation strategy to alleviate optimization difficulty. It introduces a motion-focused loss function that assigns higher weights to dynamic regions, such as limbs. The criterion for identifying these dynamic regions is defined as
\begin{equation}
\text{Mask}_{i,j} = \begin{cases} 1, & \text{if } |G_\theta(\epsilon,1)_{[1:L,i,j]}-G_\theta(\epsilon,1)_{[0:L-1,i,j]}| > 0.2, \\ 0, & \text{otherwise}, \end{cases}
\end{equation}
where $i$ and $j$ represent the indices for the height and width of $G_\theta(\epsilon,1)$. This approach mitigates the inherent ambiguity in the distillation process stemming from an imperfect teacher model.

\hspace*{-\parindent}\textbf{$\blacktriangleright$ MotionLCM.} MotionLCM~\cite{dai2024motionlcmrealtimecontrollablemotion} targets controllable motion generation by integrating a Motion ControlNet into the latent space. In this framework, $M(\cdot)$ incorporates explicit supervision from the motion space to ensure faithful adherence to control signals. Since the latent space lacks explicit geometric interpretation, making direct supervision of joint positions infeasible, the predicted latent variable $\hat{z}_0$ is decoded back into the motion space $\hat{x}_0$ for supervision, thereby enhancing motion consistency.

\hspace*{-\parindent}\textbf{$\blacktriangleright$ MotionPCM.} To effectively model complex human motion dynamics, MotionPCM~\cite{jiang2025motionpcmrealtimemotionsynthesis} partitions the diffusion trajectory into $K$ sub-intervals $[t_i, t_{i+1}]$. In contrast to standard global origin mapping, MotionPCM enforces local consistency by mapping $z_t$ to the interval start point $z_{t_i}$. This piecewise alignment strategy, formulated as $\min \Vert G_\theta(z_t,t)- z_{t_i} \Vert_2^2$, mitigates the accumulation of discretization errors. Furthermore, the method incorporates a distribution-matching adversarial loss to facilitate high-fidelity motion synthesis suitable for real-time applications.

\hspace*{-\parindent}\textbf{$\blacktriangleright$ T2V-Turbo-v2.} T2V-Turbo-v2~\cite{https://doi.org/10.48550/arxiv.2410.05677} integrates a diverse mixture of differentiable reward models into the consistency distillation framework to enhance both visual quality and semantic alignment. Recognizing that image-based feedback alone is insufficient for video generation, it employs a combination of image-text reward models, specifically HPSv2.1~\cite{HPSV2} and CLIPScore, to ensure frame-level visual fidelity, alongside a video-text reward model, InternVideo2~\cite{wang2024internvideo2}, to capture temporal dynamics and semantic consistency. The model is supervised through a direct optimization strategy, where gradients are backpropagated from these weighted reward models to the student model, ensuring the generated videos are effectively aligned with human preferences across multiple dimensions.

\hspace*{-\parindent}\textbf{$\blacktriangleright$ AdaDiff.} For quality supervision, AdaDiff~\cite{Zhang_2025_AdaDiff} employs a specialized evaluation model that concurrently assesses image-text alignment and perceptual fidelity. To ensure robust guidance, AdaDiff eschews absolute score thresholds in favor of a relative ranking mechanism. Specifically, for a given prompt, the model evaluates the quality of videos generated across all candidate step counts; a positive reward is assigned only if the video produced by the current step count ranks within the top-$k$ candidates.

\hspace*{-\parindent}\textbf{$\blacktriangleright$ DOLLAR.} DOLLAR~\cite{https://doi.org/10.48550/arxiv.2412.15689} introduces a latent reward model (LRM) strategy to facilitate efficient preference optimization for video generation, thereby bypassing the prohibitive memory overhead associated with standard pixel-space fine-tuning. Rather than computing rewards on decoded videos, DOLLAR trains a lightweight proxy model that operates directly within the latent space. This LRM is supervised to approximate the outputs of established pixel-space preference models; specifically, the authors employ HPSv2 and PickScore~\cite{kirstain2023pickapicopendatasetuser} as teacher models to provide the ground-truth signals. For generator supervision, the model is fine-tuned by maximizing the scores predicted by this differentiable LRM, enabling efficient gradient updates without the need for high-resolution video decoding.

\hspace*{-\parindent}\textbf{$\blacktriangleright$ SwiftVideo.} SwiftVideo~\cite{https://doi.org/10.48550/arxiv.2508.06082} employs a trajectory alignment strategy based on DPO rather than the aforementioned direct optimization, enhancing the generation quality of few-step videos without necessitating external reward models. It constructs a self-synthetic preference dataset where videos generated with a larger NFE (e.g., $S=8$) are designated as winners ($z_0^w$), and those with a smaller NFE (e.g., $S=4$) serve as losers ($z_0^l$). The DPO loss for the velocity-based student model $G_\theta$ is formulated as
\begin{equation} \begin{split}
& \mathcal{L}_{\text{dpo}}(G_\theta; G_{\text{ref}})  =  -\mathbb{E}_{z_t, t, c} \Big[ \log \sigma^\text{dpo}\Big( \\
 &\beta^\text{dpo} (\Vert z_t - tG_{\text{ref}}(z_t, t) - z_0^w \Vert_2^2 - \Vert z_t -tG_\theta(z_t, t) - z_0^w \Vert_2^2 ) \\
&- \beta^\text{dpo} ( \Vert z_t -tG_{\text{ref}}(z_t, t) - z_0^l \Vert_2^2 - \Vert z_t -tG_\theta(z_t, t) - z_0^l \Vert_2^2 ) \Big) \Big].\\ \end{split} 
\label{eq:swiftvideo_dpo_final}
\end{equation}
where $\sigma^\text{dpo}$ is the sigmoid function, $\beta^\text{dpo}$ is a temperature hyperparameter, and $G_{\text{ref}}$ represents the frozen reference model. By minimizing this objective, the student is incentivized to decrease the reconstruction error for the high-quality winner $z_0^w$ more aggressively than for the low-quality loser $z_0^l$, thereby aligning the few-step generative trajectory with high-NFE teacher outputs.

\subsubsection{Distribution Distillation}

This subsubsection highlights representative methods under distribution distillation.

\paragraph{Non-Streaming Distribution Distillation}

This paragraph highlights representative methods under non-streaming distribution distillation.

\hspace*{-\parindent}\textbf{$\blacktriangleright$ TDM.} TDM~\cite{https://doi.org/10.48550/arxiv.2503.06674} mitigates the accumulation of numerical errors inherent in conventional trajectory distillation by incorporating distribution matching during trajectory generation. Rather than enforcing strict point-wise alignment with the teacher's ODE solver, TDM employs a distribution matching objective to align student trajectory segments with the teacher's distribution. Specifically, let $z_S$ denote Gaussian noise, which is iteratively denoised to yield the sequence $\{z_i\}_{i=0}^S$. Each state $z_i$ facilitates the prediction of a clean latent $\hat{z}_{0,i}$ via Tweedie’s formula. Distribution matching is subsequently performed for each $\hat{z}_{0,i}$ as follows:
\begin{equation}
    \begin{split}
\mathcal{L}_{\text{tdm}} &= \sum_{i=0}^{S-1} D_{\mathrm{KL}}\!\left(\hat{z}_{t,i}-tf^\phi_{\text{fake}}(\hat{z}_{t,i},t)\Vert \hat{z}_{t,i} -tf^\phi_{\text{real}}(\hat{z}_{t,i},t)\right),\\
&\text{ where }\hat{z}_{t,i} = (1-t)\hat{z}_{0,i} + tz_S,\text{ }t\in\mathcal{U}[i/S,(i+1)/S). \\
    \end{split}
\end{equation}
Unlike standard DMD, TDM performs distribution matching across the entire trajectory. Furthermore, TDM is a data-free algorithm.

\hspace*{-\parindent}\textbf{$\blacktriangleright$ BLADE.} BLADE~\cite{gu2025blade} addresses the sub-optimality of treating step distillation and architectural compression as decoupled optimization phases, a practice that frequently results in degraded performance or prohibitive computational overhead. To this end, BLADE introduces a data-free joint training paradigm that integrates an adaptive block sparse attention mechanism with TDM. This sparsity-aware distillation significantly enhances model fidelity.

\hspace*{-\parindent}\textbf{$\blacktriangleright$ AVDM2.} AVDM2~\cite{https://doi.org/10.48550/arxiv.2412.05899} introduces a hybrid framework that couples adversarial learning with DMD to distill few-step video generators. To align the student model with the underlying data distribution, AVDM2 employs a video adversarial loss, utilizing a discriminator initialized from the teacher's encoder to distinguish between synthesized and ground-truth videos. This approach notably enhances visual quality and represents a pioneer effort in scaling DMD to video diffusion models.

\hspace*{-\parindent}\textbf{$\blacktriangleright$ MagicDistillation.} MagicDistillation~\cite{shao2025magicdistillationweaktostrongvideodistillation} addresses the prohibitive memory overhead incurred by concurrently loading three models, namely the few-step generator, the fake model, and the real model, during distribution distillation. To alleviate this bottleneck, MagicDistillation facilitates parameter sharing between the fake and real models, where the learnable components of the fake model are implemented via low-rank adaptation (LoRA) modules integrated into the original backbone. This strategy significantly reduces GPU memory consumption, enabling the successful training of large-scale models such as HunyuanVideo-14B~\cite{hunyuanvideo} and Wan2.1-13B~\cite{wan2025}.

\subsubsection{Adversarial Distillation}

This subsubsection highlights representative methods under adversarial distillation.

\paragraph{Combined Adversarial Distillation}

This paragraph highlights representative methods under combined adversarial distillation.

\hspace*{-\parindent}\textbf{$\blacktriangleright$ AnimateDiff-Lightning.} AnimateDiff-Lightning~\cite{https://doi.org/10.48550/arxiv.2403.12706} integrates progressive distillation~\cite{iclr22_progressive} with a non-saturating adversarial loss, leveraging both conditional and unconditional discriminators to enhance generative quality. Although progressive distillation established the precedent for step distillation, it was subsequently surpassed by the emergence of LCM and DMD. Notably, apart from AnimateDiff-Lightning, this distillation approach has not been further extended to video diffusion models. Ultimately, this joint training paradigm enables high-fidelity video synthesis with fewer sampling steps and superior motion realism compared to standalone progressive trajectory distillation.

\hspace*{-\parindent}\textbf{$\blacktriangleright$ POSE.} POSE~\cite{cheng2025pose} integrates adversarial distillation with score-based distillation. During the training process, the student model undergoes an initial warmup phase utilizing VSD~\cite{vsd}, followed by a joint training stage that couples adversarial distillation with VSD. This multi-stage optimization strategy mitigates the inherent instability associated with standalone adversarial distillation, thereby facilitating single-step video generation for large-scale DiT architectures.

\hspace*{-\parindent}\textbf{$\blacktriangleright$ MoGAN.} MoGAN~\cite{xue2025moganimprovingmotionquality} integrates a motion-centric adversarial objective in optical flow space with a distribution matching loss to preserve appearance fidelity and ensure structural alignment. While the adversarial branch specifically targets temporal dynamics, the distribution matching objective anchors the model to a reduced number of sampling steps. This synergistic approach yields superior motion quality without compromising visual consistency.

\paragraph{Independent Adversarial Distillation}

This paragraph highlights representative methods under independent adversarial distillation.

\hspace*{-\parindent}\textbf{$\blacktriangleright$ SF-V.} SF-V~\cite{https://doi.org/10.48550/arxiv.2406.04324} proposes a latent adversarial training framework designed to distill pre-trained video diffusion models into a single-step generator. Unlike contemporary hybrid approaches, SF-V operates as an independent adversarial distillation paradigm that does not rely on integration with consistency distillation or multi-step distribution matching targets. To ensure high-fidelity synthesis and motion diversity within a single forward pass, the framework introduces a discriminator architecture featuring separate spatial and temporal heads. These heads are attached to a frozen pre-trained UNet backbone to supervise frame-level quality and inter-frame consistency respectively, preventing the generated sequences from collapsing into static images. By combining a Pseudo-Huber reconstruction loss with a discretized lognormal noise distribution for the discriminator, SF-V achieves stable convergence and high-quality motion synthesis using exclusively adversarial objectives.

\hspace*{-\parindent}\textbf{$\blacktriangleright$ Seaweed-APT.} Seaweed-APT~\cite{https://doi.org/10.48550/arxiv.2501.08316} differs from most distillation methods that rely on teacher-generated targets; it utilizes the pre-trained weights solely for initialization and subsequently performs adversarial optimization directly against real data, thereby bypassing the quality upper bound imposed by a teacher model. To stabilize the generative adversarial system, the framework introduces several architectural refinements, such as a discriminator design based on multi-layer Transformer feature fusion. Empirical results demonstrate that this approach significantly enhances visual fidelity and realism for one-step generation, enabling real-time, high-resolution video synthesis on H100 clusters.

\subsection{Efficient Attention: Representative Methods}\label{app:method_efficient_attention}

This subsection introduces representative methods under efficient attention in Section~\ref{sec:efficient_attention}.

\subsubsection{Dynamic Sparse Attention}

This subsubsection highlights representative methods under dynamic sparse attention.

\hspace*{-\parindent}\textbf{$\blacktriangleright$ SpargeAttention.} SpargeAttention~\cite{sparge_attn} is a universal, training-free sparse attention mechanism that utilizes a two-stage online filtering process to accelerate diverse generative models without compromising accuracy. To handle the varying attention patterns across tasks, it introduces a selective token compression algorithm for sparse mask prediction. For each block $Q_i$ and $K_j$, it computes mean-pooled representative tokens $q_i = \mathrm{mean}(Q_i)$ and $k_j = \mathrm{mean}(K_j)$. The block-level sparse mask $M_g$ is then determined by calculating a compressed attention map $\hat{P}_{ij} = \mathrm{softmax}(q_i k_j^\top / \sqrt{D})$ and applying a Top-Cdf selection: $M_g[i,j] = 1$ if $\hat{P}_{ij}$ is among the top values whose cumulative sum reaches a threshold $\tau$, and $M_g[i,j] = 0$ otherwise. To ensure robustness, blocks with low internal cosine similarity are always computed. During the FlashAttention inner loop, if $M_g[i,j] = 1$, the scores $S_{ij}$ are computed, and a second-stage online softmax-aware filter further skips the $P_{ij}V_j$ product if the local maximum $m_{\text{local}} = \max_{\text{row}}(S_{ij})$ is significantly smaller than the global running maximum $m_i$. This condition is formulated as:
\begin{equation}
\begin{split}
\text{Skip } \tilde{P}_{ij}V_j \text{ if } \max(m_{\text{local}} - m_i) < \lambda,
\end{split}
\end{equation}
where $\lambda$ is a hyper-parameter controlling the approximation error. By integrating these filters into the 8-bit quantized SageAttention framework and utilizing a HilbertCurve permutation to enhance spatial locality for visual tokens, SpargeAttention achieves substantial speedup over dense kernels while maintaining end-to-end performance in long-context language and high-resolution video synthesis.

\hspace*{-\parindent}\textbf{$\blacktriangleright$ Sparse VideoGen2.} Sparse VideoGen2~\cite{https://doi.org/10.48550/arxiv.2505.18875} addresses the limitations of position-based clustering and the computational waste caused by scattered critical tokens through a training-free semantic-aware permutation framework. Instead of grouping spatially adjacent tokens, SVG2 applies k-means clustering to $Q \in \mathbb{R}^{N_q \times D}$ and $K \in \mathbb{R}^{N_k \times D}$ to aggregate semantically similar tokens, subsequently reordering them into a contiguous layout via permutation matrices $\pi_q \in \mathbb{R}^{N_q \times N_q}$ and $\pi_k \in \mathbb{R}^{N_k \times N_k}$. The attention operation is then performed on these densified, permuted tensors to ensure hardware efficiency:
\begin{equation}
\begin{split}
A = \pi_q^\top \left( \mathrm{Softmax}\left(\frac{(\pi_q Q)(\pi_k K)^{\top} \odot \mathcal{M} + (1-\mathcal{M}) \odot (-\infty)}{\sqrt{D}}\right)(\pi_k V) \right),
\end{split}
\end{equation}
where $\mathcal{M}$ represents the sparse mask derived from a centroid-based top-$p$ selection strategy. Specifically, SVG2 approximates attention scores using cluster centroids to identify critical blocks dynamically, thereby maximizing critical token recall while minimizing the padding overhead typically associated with non-contiguous sparse patterns.

\hspace*{-\parindent}\textbf{$\blacktriangleright$ FlashOmni.} FlashOmni~\cite{qiao2025flashomniunifiedsparseattention} compresses the decision logic for feature cache and block skipping into 8-bit sparse symbols, denoted as $\mathcal{S}_c$ and $\mathcal{S}_s$. The process begins by computing a low-resolution proxy attention map $\tilde{P} = \mathrm{softmax}(\text{mean}(Q)\text{mean}(K)^\top / \sqrt{d})$ to estimate block importance. For feature cache, FlashOmni explicitly defines two metrics: vision-to-text contribution $\mathcal{C}_{v\to t}$ and text-to-vision guidance $\mathcal{G}_{t\to v}$, derived from specific regions of $\tilde{P}$. The logical binary masks $M_c$ (where 0 indicates reuse) are generated by prioritizing blocks with lower combined importance scores until a cumulative threshold $\tau_c$ is reached. $M_s$ follows a dynamic selection strategy (e.g., SpargeAttention~\cite{sparge_attn}) on $\tilde{P}$. These boolean masks are then quantized into compact integer symbols through Big-Endian bitwise packing:
\begin{equation}
\begin{aligned}
M_c[i] &= \mathbb{I}\left(\text{CumSum}(\mathcal{C}_{v\to t} + \mathcal{G}_{t\to v})_i > \tau_c\right), \quad \mathcal{S}_c[k] = \sum_{b=0}^{7} M_c[8k + b] \cdot 2^{7-b}, \\
M_s[i,j] &= \mathbb{I}(\tilde{P}_{ij} \in \text{Top}_{\tau_s}), \quad \quad \quad \quad \quad \quad \quad \mathcal{S}_s[i, k] = \sum_{b=0}^{7} M_s[i, 8k + b] \cdot 2^{7-b}, \\
\end{aligned}
\end{equation}
where $\mathbb{I}(\cdot)$ is the indicator function. The summation represents the Big-Endian bitwise packing of 8 consecutive boolean flags into a single uint8 symbol, enabling efficient decoding within the unified attention kernel.

\subsubsection{Training-Based Linear Attention}

This subsubsection highlights representative methods under training-based linear attention.

\hspace*{-\parindent}\textbf{$\blacktriangleright$ SANA-Video.} SANA-Video~\cite{https://doi.org/10.48550/arxiv.2509.24695} extends the linear DiT architecture to the video domain by completely replacing quadratic attention modules with an efficient linear attention mechanism. To effectively incorporate positional information while maintaining training stability, it applies Rotary Position Embeddings ($\text{RoPE}(\cdot)$) after the ReLU activation function $\phi(\cdot)$. Crucially, to prevent numerical instability caused by the potential negativity of rotated features, RoPE is excluded from the query term in the denominator's normalization factor. The output $O_i$ for the $i$-th token is thus computed as
\begin{equation}
\begin{split}
O_i = \frac{\mathrm{RoPE}(\phi(Q_i)) \left(\sum_{j=1}^{N_K} \mathrm{RoPE}(\phi(K_j))^{\top} V_j\right)}{\phi(Q_i) \left(\sum_{j=1}^{N_K} \mathrm{RoPE}(\phi(K_j))^{\top}\right)},
\end{split}
\end{equation}
where the numerator utilizes fully rotated queries and keys to capture relative positions, while the denominator relies on the unrotated query to ensure a positive normalization term. Furthermore, SANA-Video leverages the cumulative property of this linear formulation to enable block-wise autoregressive generation. By maintaining a recurrent state of fixed size $\mathcal{O}(D^2)$, it achieves a constant-memory KV cache that is independent of the sequence length $N_K$, effectively mitigating the memory bottlenecks typically associated with long video synthesis.

\hspace*{-\parindent}\textbf{$\blacktriangleright$ SLA.} SLA~\cite{https://doi.org/10.48550/arxiv.2509.24006} introduces a fine-tunable hybrid mechanism that dynamically routes computation between precise sparse attention and approximate linear attention based on lexical saliency. The framework utilizes a compressed predictor, consistent with the methodology for computing compressed attention maps in SpargeAttention~\cite{sparge_attn}, to generate a block-level mask $M_c$. Under this routing policy, negligible blocks are discarded, whereas critical blocks ($M_c[i,j]=1$) undergo standard softmax attention. Concurrently, marginal blocks ($M_c[i,j]=0$) are processed via linear attention to recover global semantic information. The fused output $O_i$ for the $i$-th query block is formulated as follows:
\begin{equation}
\begin{split}
O_i = \sum_{j \in \mathcal{N}_{\text{crit}}} \mathrm{Softmax}\left(\frac{Q_i K_j^\top}{\sqrt{D}}\right) V_j + \mathrm{Proj}\left( \frac{\phi(Q_i) \sum_{j \in \mathcal{N}_{\text{marg}}} \phi(K_j)^\top V_j}{\phi(Q_i) \sum_{j \in \mathcal{N}_{\text{marg}}} \phi(K_j)^\top} \right),
\end{split}
\end{equation}
where $\mathcal{N}_{\text{crit}}$ and $\mathcal{N}_{\text{marg}}$ denote the sets of key-value blocks identified as critical ($M_c[i,j]=1$) and marginal ($M_c[i,j]=0$), respectively. The function $\phi(\cdot)$ represents an activation function that linearizes the attention computation for the marginal component, and $\mathrm{Proj}(\cdot)$ is a learnable linear projection designed to align the distribution of the linear approximation with the sparse attention output.

\subsection{Model Compression: Representative Methods}\label{app:method_model_compression}

This subsection introduces representative methods under model compression in Section~\ref{sec:model_compression}.

\subsubsection{Quantization-Aware Training}

This subsubsection highlights representative methods under quantization-aware training.

\hspace*{-\parindent}\textbf{$\blacktriangleright$ FPSAttention.} FPSAttention~\cite{https://doi.org/10.48550/arxiv.2506.04648} is a training-aware method that integrates FP8 quantization and structured sparsity directly into the attention mechanism. Instead of learning quantization parameters, it employs a {denoising step-aware schedule} to dynamically adjust the sparsity window $W_t$. The attention output is computed as
\begin{equation}
A_t = \mathrm{Dequant}\!\left(\mathrm{FP8}\!\left(\mathrm{Softmax}\!\left(\frac{Q_t^{\mathrm{fp8}}(K_t^{\mathrm{fp8}})^\top \odot M(W_t)}{\sqrt{d}}\right)\right) \cdot V_t^{\mathrm{fp8}}\right),
\end{equation}
where $Q_t^{\mathrm{fp8}}, K_t^{\mathrm{fp8}}$ are quantized tile-wise, and $V_t^{\mathrm{fp8}}$ is quantized channel-wise using analytically computed scales $s_t$. The training objective remains the standard diffusion loss, but the model parameters $\theta$ are finetuned to adapt to these structured approximations:
\begin{equation}
\min_{\theta}\;\mathbb{E}_{z_0,z_1,t}\left[\left\Vert G_{\theta}(z_t,t, s_t, M_t)-(z_1-z_0)\right\Vert_2^2\right].
\end{equation}
Here, the explicit regularization is replaced by the architectural constraints imposed by the FPSAttention kernel during the forward pass.

\subsubsection{Post-Training Quantization}

This subsubsection highlights representative methods under post-training quantization.

\hspace*{-\parindent}\textbf{$\blacktriangleright$ SageAttention2.} SageAttention2~\cite{zhang2025sageattention2efficientattentionthorough} optimizes the attention mechanism by quantizing $Q$ and $K$ to INT4 with hardware-friendly per-thread granularity, while maintaining $P$ and $V$ in FP8. To address the precision loss from outliers in INT4, it applies a dual smoothing technique by subtracting channel-wise means, denoted as $\gamma(Q_i) = Q_i - \bar{q}_i$ and $\gamma(K_j) = K_j - \bar{k}$. Exploiting the translation invariance of Softmax, the effective attention scores are computed by decomposing $Q_i K_j^\top$ and discarding the bias term constant to each row:
\begin{equation}
S_{ij} = \text{Dequant}\big( \text{INT4}(\gamma(Q_i)) \cdot \text{INT4}(\gamma(K_j))^\top \big) + \bar{q}_i \gamma(K_j)^\top.
\end{equation}
This approximates the centered product term while the correction $\bar{q}_i \gamma(K_j)^\top$ restores the necessary rank information.

\hspace*{-\parindent}\textbf{$\blacktriangleright$ Q-DiT.} Q-DiT~\cite{Chen_2025} addresses the significant spatial variance across input channels and the temporal shifts in activations inherent to Diffusion Transformers. To optimize the trade-off between quantization granularity and generation quality, it employs an evolutionary search algorithm to determine the optimal layer-wise group size configuration $g = \{g_1, \dots, g_L\}$, where $g_l$ denotes the number of elements sharing quantization parameters in layer $l$. Unlike standard loss minimization, Q-DiT directly minimizes the generation distance metric (FID or FVD) $\mathcal{D}_{\text{FID/FVD}}$ subject to a BitOps computational budget $N_{\text{ops}}$:
\begin{equation}
g^* = \operatorname*{arg\,min}_g \mathcal{D}_{\text{FID/FVD}}(R, G_{\theta}^{(g)}) \quad \text{s.t.} \quad \mathcal{B}(g) \le N_{\text{ops}},
\end{equation}
where $R$ represents real samples, $G_{\theta}^{(g)}$ denotes samples generated by the model $\theta$ quantized with configuration $g$, and $\mathcal{B}(\cdot)$ calculates the total bit-operations. While $g$ determines the granularity, the quantization parameters are adaptive. Q-DiT implements sample-wise dynamic activation quantization where the scale $s_{i,t}$ and zero-point $z_{i,t}$ for a sample $i$ at timestep $t$ are computed on-the-fly:
\begin{equation}
\begin{split}
s_{i,t} &= \frac{\max(x_{i,t}) - \min(x_{i,t})}{2^b - 1}, \quad z_{i,t} = \left\lfloor \frac{\min(x_{i,t})}{s_{i,t}} \right\rfloor,\\
\hat{x}_{i,t} &= s_{i,t} \cdot \left(\operatorname{clip}\left(\left\lfloor \frac{x_{i,t}}{s_{i,t}} \right\rfloor + z_{i,t}, 0, 2^b - 1\right) - z_{i,t}\right).
\end{split}
\end{equation}
This combination of evolutionary granularity search (i.e., $g$) and dynamic parameter adaptation (i.e., $s$ and $z$) allows Q-DiT to achieve W4A8 quantization with minimal degradation in visual fidelity.

\hspace*{-\parindent}\textbf{$\blacktriangleright$ S$^2$Q-VDiT.} S$^2$Q-VDiT~\cite{feng2025s2qvditaccuratequantizedvideo} mitigates the high variance inherent in video calibration by replacing random timestep sampling with a Hessian-aware salient data selection strategy. Recognizing that different timesteps contribute unequally to the denoising trajectory and quantization sensitivity, it identifies the optimal calibration subset through a dual-metric assessment. Specifically, it quantifies the diffusion informativeness $C_{\text{diff}}$ via latent transition magnitude and the quantization sensitivity $C_{\text{quant}}$ using a Hessian-based approximation:
\begin{equation}
C_{\text{diff}}(z_t) = \frac{\|z_t - z_{t-1}\|_2^2}{\|z_t\|_2^2}, \quad C_{\text{quant}}(z_t) = \|z_t^\top z_t\|_2,
\end{equation}
where $z_t$ represents the video latent representation at timestep $t$. To select the most representative temporal data, S$^2$Q-VDiT constructs the calibration set $D_{\text{calib}}$ by maximizing a unified salience score derived from the normalized product of these metrics:
\begin{equation}
C_{\text{sample}}(z_t) = {C}_{\text{diff}}(z_t) \cdot {C}_{\text{quant}}(z_t), \quad D_{\text{calib}} = \operatorname*{Top-K}_{t \in [1, S]} \{ C_{\text{sample}}(z_t) \}.
\end{equation}
This approach ensures that the quantization parameters are calibrated on timesteps that embody the most significant distribution shifts and error sensitivities, thereby stabilizing the post-training optimization process.

\subsubsection{Token Pruning}

This subsubsection highlights representative methods under token pruning.

\hspace*{-\parindent}\textbf{$\blacktriangleright$ AsymRnR.} AsymRnR~\cite{https://doi.org/10.48550/arxiv.2412.11706} introduces a training-free acceleration strategy that decouples the token reduction of Query ($Q$) and Key-Value ($K, V$) sequences. To select important tokens, AsymRnR employs a clustering-based approach using negative Euclidean distance rather than cosine similarity, ensuring sensitivity to feature magnitude. Redundant tokens are explicitly discarded, rather than averaged, to preserve the original feature distribution. This asymmetric design allows for aggressive compression of robust features (typically $V$) while preserving sensitive ones (typically $Q$ in early layers).

\hspace*{-\parindent}\textbf{$\blacktriangleright$ FastVID.} FastVID~\cite{shen2025fastviddynamicdensitypruning} proposes a dynamic inference-time acceleration framework that systematically addresses video redundancy through two distinct perspectives: temporal and visual context. To preserve temporal structure, the method employs dynamic temporal segmentation, which adaptively partitions videos into temporally ordered segments based on frame transition similarity rather than fixed intervals. Within these high-redundancy segments, density spatiotemporal pruning consolidates tokens using a hybrid strategy. Specifically, density-based token merging selects distinctive anchor tokens via density peak clustering and aggregates neighbors to preserve global visual context, while attention-based token selection retains salient tokens identified by attention scores to capture fine-grained details.

\subsubsection{Channel Pruning}

This subsubsection highlights representative methods under channel pruning.

\hspace*{-\parindent}\textbf{$\blacktriangleright$ Mobile Video Diffusion.} Mobile Video Diffusion~\cite{https://doi.org/10.48550/arxiv.2412.07583} addresses channel-dimension compression to facilitate efficient on-device video generation. Its core mechanism involves a channel funnel inserted between affine layers to constrain intermediate channel dimensionality during optimization. These components are subsequently merged into the original weights at deployment, ensuring that no additional inference branches are introduced. Formally, consider an affine transformation $y=W_2\,\mathrm{GELU}(W_1x)$. The channel funneling operation is expressed as
\begin{equation}
\hat{y}=W_2F\ \mathrm{GELU}\!\left(F^{\top}W_1x\right),\qquad F\in\mathbb{R}^{C\times C^\prime},\qquad C^\prime <C,
\end{equation}
where $C^\prime$ denotes the pruned channel width. Post-training, $F$ is absorbed into the adjacent weight matrices to produce a compact equivalent mapping. This approach yields significant efficiency gains with minimal quality degradation, positioning it as a representative channel-oriented compression strategy for mobile video diffusion.

\subsubsection{Model Pruning}

This subsubsection highlights representative methods under model pruning.

\hspace*{-\parindent}\textbf{$\blacktriangleright$ ICMD.} ICMD~\cite{wu2025individualcontentmotiondynamics} introduces a structural pruning framework grounded in the observation that deeper network layers govern motion dynamics while shallower layers determine individual content. To quantify redundancy, the method evaluates the sensitivity of each block $b_l$ within the pre-trained teacher $G_{\theta}$ by measuring the generation quality degradation, specifically the generation distance metric $\mathcal{D}_{\text{FID/FVD}}$, when the block is bypassed:
\begin{equation}
\mathcal{I}_l = \mathcal{D}_{\text{FID/FVD}}(R, G_{\theta \setminus b_l}),
\end{equation}
where $R$ represents real samples and $G_{\theta \setminus b_l}$ denotes the model with block $b_l$ replaced by an identity mapping or a projection layer. Blocks with low sensitivity $\mathcal{I}_l$, primarily located in shallower layers, are pruned to yield a lightweight student architecture $G_{\theta}^\text{pruning}$. To compensate for the capacity reduction, the framework utilizes a dual-objective consistency loss, combining spatial feature distillation with a temporal adversarial objective:
\begin{equation}
\mathcal{L}_{\text{total}} = \lambda_{\text{ICD}} \sum_{k=1}^{K} \| f_k^\text{pruning}(z_t) - f_k^{\theta}(z_t) \|_2^2 + \lambda_{\text{MCA}} \mathcal{L}_{\text{adv}}(G_{\theta}^\text{pruning}, D),
\end{equation}
where $\lambda_{\text{ICD}}$ and $\lambda_{\text{MCA}}$ are hyperparameters balancing the spatial and temporal objectives. The index $k$ iterates over $K$ (typically $K=4$) selected intermediate feature maps to enforce frame-level spatial fidelity. $\mathcal{L}_{\text{adv}}$ represents the multi-frame adversarial loss computed via a discriminator $D$, which employs spatial and temporal heads to distinguish between the video latents of the student and teacher models, thereby preserving global motion dynamics.

\hspace*{-\parindent}\textbf{$\blacktriangleright$ Taming DiT.} Taming DiT~\cite{wu2025tamingdiffusiontransformerefficient} introduces a sensitivity-aware tri-level pruning strategy designed to optimize DiT architectures for resource-constrained mobile devices. In contrast to coarse-grained methods that rely solely on removing transformer blocks, this approach operates across three distinct levels of granularity: blocks, multi-head attention heads, and FFN dimensions. The selection of redundant components is governed by learnable binary masks, which are optimized to balance model fidelity against specific hardware constraints. This process is driven by a sensitivity analysis establishing that FFN dimensions contribute more significantly to performance than attention heads, thereby prioritizing the aggressive pruning of the latter. To navigate the vast search space efficiently, the method utilizes Gumbel-Softmax sampling for differentiable mask selection and incorporates a knowledge distillation framework with feature alignment to recover the generative capacity of the pruned sub-network.

\subsection{Cache and Trajectory Optimization: Representative Methods}\label{app:method_cache_trajectory}

This subsection introduces representative methods under cache and trajectory optimization in Section~\ref{sec:cache_and_trajectory}.

\subsubsection{Feature Cache}

This subsubsection highlights representative methods under feature cache.

\hspace*{-\parindent}\textbf{$\blacktriangleright$ PAB.} PAB~\cite{https://doi.org/10.48550/arxiv.2408.12588} accelerates DiT inference by exploiting the U-shaped redundancy pattern where feature differences are minimal during the middle sampling steps. The method implements a Pyramid Attention Broadcast strategy that directly reuses the feature output from a source timestep $t$ for a subsequent window of steps $t-k$. Crucially, the broadcast range $R_{\ell}$ is adaptively tailored to the specific type of block $\ell$ (e.g., spatial, temporal, or cross-modal) based on its sensitivity to temporal changes. The reuse mechanism is formulated as
\begin{equation}
{F}_{t-k}^{\ell} = F_t^{\ell}, \quad \text{where } 1 \le k \le R_{\ell},
\end{equation}
where $R_{\ell}$ denotes the broadcast range determined by the variance of block $\ell$. Additionally, PAB introduces Broadcast Sequence Parallelism, which exploits the broadcasted temporal attention to bypass inter-GPU communication during distributed inference.

\hspace*{-\parindent}\textbf{$\blacktriangleright$ FasterCache.} FasterCache~\cite{https://doi.org/10.48550/arxiv.2410.19355} addresses the detail degradation inherent in static feature repetition by modeling inter-step feature evolution as a continuous trajectory. Recognizing that simple caching ignores the subtle gradients essential for denoising, it employs a dynamic feature reuse strategy that approximates the feature map at a skipped timestep $t-\Delta t$ via a first-order Taylor expansion based on adjacent cached states. Concurrently, to alleviate the computational burden of CFG, the method introduces CFG-Cache, which exploits the spectral redundancy between conditional and unconditional velocity predictions. By decomposing the residual $\delta_t = G_\theta(x_t,t, \emptyset) - G_\theta(x_t,t,c)$ into frequency components, the unconditional branch is synthesized rather than computed. The joint acceleration mechanism is formulated as
\begin{equation}
\begin{split}
\text{Feature}: \quad & \tilde{F}_{t-\Delta t}^{\ell} = C_{t}^{\ell} + \gamma_t \left(C_{t}^{\ell} - C_{t+\Delta t}^{\ell}\right), \\
\text{CFG}: \quad & G_\theta(x_{t-\Delta t}, t, \emptyset) = G_\theta(x_{t-\Delta t},t, c) + \mathcal{F}^{-1}\left( \Lambda_t \odot \mathcal{F}(\delta_t) \right), \\
\end{split}
\label{eq:faster_cache}
\end{equation}
where $\gamma_t$ is a time-dependent scalar approximating the velocity of feature drift, $\mathcal{F}$ denotes the Fourier transform, and $\Lambda_t$ represents an adaptive scaling function that modulates the low- and high-frequency components of the cached residual $\delta_t$ to rectify spectral shifts across the sampling schedule.

\hspace*{-\parindent}\textbf{$\blacktriangleright$ TaylorSeer.} TaylorSeer~\cite{liu2025reusingforecastingacceleratingdiffusion} advances beyond the static ``cache-then-reuse'' paradigm by adopting a ``cache-then-forecast'' strategy to mitigate the feature divergence that typically restricts acceleration ratios in standard caching frameworks. Motivated by the observation that feature trajectories in diffusion models exhibit smoothness and differentiability, the method synthesizes intermediate features at skipped timesteps via truncated Taylor series expansion. Rather than computing explicit derivatives, it approximates higher-order temporal variations using finite differences derived from features cached at a fixed refresh interval $k^{\text{refresh}}$. For an $m$-th order approximation, the feature map at a skipped step $t-k$ is extrapolated as
\begin{equation}
\begin{split}
\Delta^i F_t^{\ell} &= \sum_{j=0}^i (-1)^{i-j} \binom{i}{j} F_{t+j \cdot k^{\text{refresh}}}^{\ell}, \\
\tilde{F}_{t-k}^{\ell} &= F_t^{\ell} + \sum_{i=1}^{m} \frac{\Delta^i F_t^{\ell}}{i! \, (k^{\text{refresh}})^i} (-k)^i, \\
\end{split}
\label{eq:taylor_seer}
\end{equation}
where $\Delta^i F_t^{\ell}$ denotes the $i$-th order finite difference derived from the cache history. This formulation enables the model to rectify temporal drift and maintain generation fidelity across extended skipping intervals by accurately modeling the non-linear evolution of the feature manifold.

\hspace*{-\parindent}\textbf{$\blacktriangleright$ AdaCache.} AdaCache~\cite{https://doi.org/10.48550/arxiv.2411.02397} exemplifies the partial refresh paradigm by isolating and caching the residual updates of specific, computationally intensive sub-modules, such as Attention or multilayer perceptrons (MLPs), rather than bypassing entire transformer blocks. By maintaining the forward flow of the primary latent stream and only skipping the heavy sub-network evaluations, it preserves finer structural integrity during skipped steps. To determine the refresh intervals dynamically, AdaCache evaluates the rate of change in these isolated residual computations across recent timesteps. Furthermore, to prevent visual degradation in highly dynamic video scenes, the framework integrates a motion regularization scheme. This explicitly penalizes the reuse duration by scaling the feature distance metric according to a latent motion score $m_t$ and its temporal gradient $mg_t$, forcing more frequent sub-module refreshes for complex motions.

\hspace*{-\parindent}\textbf{$\blacktriangleright$ DiCache.} DiCache~\cite{bu2025dicacheletdiffusionmodel} introduces an adaptive error-compensated reuse framework that predicts the current feature by extrapolating along a feature evolution trend line. Recognizing that directly copying cached features causes temporal drift, DiCache treats feature evolution as a continuous trajectory. It leverages an online ``shallow-layer probe'' (evaluating only the first $m$ layers) at every step to gauge the current generation progress. Instead of static reuse, DiCache calculates a trajectory parameter $\hat{\gamma}_t$, essentially a progress scaling factor, derived from the relative distances of the probe's outputs. Using this parameter, the method fits a trend line between the two most recently fully computed network residuals (from past steps $t_\alpha$ and $t_\beta$) and extrapolates the current full-network residual $r_t$. The compensated residual is formulated as:
\begin{equation}
\begin{split}
\hat{\gamma}_t &= \frac{\|r_t^m - r_{t_\beta}^m\|_1}{\|r_{t_\alpha}^m - r_{t_\beta}^m\|_1}, \\
r_t &= r_{t_\beta} + \hat{\gamma}_t (r_{t_\alpha} - r_{t_\beta}),
\end{split}
\label{eq:dicache_trajectory}
\end{equation}
where $r_t^m = F_t^m - x_t$, $r_{t_\beta}^m = F_{t_\beta}^m - x_{t_\beta}$, and $r_{t_\alpha}^m = F_{t_\alpha}^m - x_{t_\alpha}$ represent the residuals of the shallow $m$-layer probe, while $r_{t_\alpha}$ and $r_{t_\beta}$ denote the fully computed network residuals cached from previous steps. By aligning the predicted state with the observed trend of the probe, DiCache effectively mitigates error accumulation without relying on predefined offline heuristics.

\subsubsection{KV Cache}

This subsubsection highlights representative methods under KV cache.

\hspace*{-\parindent}\textbf{$\blacktriangleright$ ViD-GPT.} ViD-GPT~\cite{https://doi.org/10.48550/arxiv.2406.10981} exemplifies the budget window and eviction paradigm by transforming the standard bidirectional temporal attention of diffusion models into a causal autoregressive process akin to GPT. Instead of maintaining an infinitely growing history, ViD-GPT enforces a strict maximum cache length budget $C^{\text{KV}}_\text{size}$. As new chunks are generated, the framework employs a FIFO queue mechanism to actively dequeue the oldest cached states once the temporal context exceeds $C^{\text{KV}}_\text{size}$. Coupled with cyclic positional embeddings, this sliding window approach ensures constant memory complexity. The causal attention output $A_i$ for the $i$-th chunk is formulated as
\begin{equation}
\begin{split}
\tilde{K}_{\text{cache}} &= \{K_j \mid \max(0, i-C^{\text{KV}}_\text{size}) \le j < i\}, \\
\tilde{V}_{\text{cache}} &= \{V_j \mid \max(0, i-C^{\text{KV}}_\text{size}) \le j < i\}, \\
A_i &= \text{Softmax}\left(\frac{Q_i [\tilde{K}_{\text{cache}}; K_i]^\top}{\sqrt{d}} + M_{\text{causal}}\right) [\tilde{V}_{\text{cache}}; V_i],
\end{split}
\label{eq:vid_gpt}
\end{equation}
where $[\cdot ; \cdot]$ denotes temporal concatenation, $M_{\text{causal}}$ is the autoregressive mask preventing attention to future frames, and $\tilde{K}_{\text{cache}}$ and $\tilde{V}_{\text{cache}}$ represent the constrained memory buffers strictly managed through the continuous eviction of stale states.

\hspace*{-\parindent}\textbf{$\blacktriangleright$ PFP.} PFP~\cite{zhang2026pretrainingframepreservationautoregressive} exemplifies the compressed memory KV paradigm by encoding extensive historical video frames into a compact, fixed-size context representation. To resolve the typical trade-off between the compression rate and the loss of high-frequency visual details, PFP introduces a dedicated memory compression network $\phi^\text{mc}(\cdot)$ that is explicitly pretrained on a random frame retrieval task. During pretraining, the generator is tasked with reconstructing randomly selected target frames by attending solely to the compressed history context. Once pretrained, this compact context acts as the bounded key and value states $\{\tilde{K}_{\text{cache}}, \tilde{V}_{\text{cache}}\}$ for subsequent autoregressive generation steps. The compressed KV states are formulated as
\begin{equation}
\begin{split}
\{\tilde{K}_{\text{cache}}, \tilde{V}_{\text{cache}}\} &= \text{Proj}(\phi^\text{mc}(z_0^{<i})), \quad \text{where } |\tilde{K}_{\text{cache}}| \le C^{\text{KV}}_\text{size}, \\
\end{split}
\label{eq:pfp}
\end{equation}
where $z_0^{<i}$ and $\text{Proj}(\cdot)$ denote the full historical clean latents prior to the current chunk $i$ and the linear projection operation, respectively. Furthermore, $\phi^\text{mc}(\cdot)$ condenses the history to fit the strict memory budget $C^{\text{KV}}_\text{size}$.

\hspace*{-\parindent}\textbf{$\blacktriangleright$ MemFlow.} MemFlow~\cite{ji2025memflowflowingadaptivememory} exemplifies the selective KV paradigm by dynamically retrieving and activating historical contexts tailored to streaming narrative prompts. To handle arbitrary scene transitions and new element insertions during interactive autoregressive generation, MemFlow introduces a narrative adaptive memory. This mechanism continually evaluates the semantic relevance between the textual query $Q_c$ (derived from the conditioning prompt $c$) and the historical visual keys to maintain a semantically aligned global memory bank. To further alleviate the computational bottleneck of attention, it employs sparse memory activation, which computes a visual relevance score between the pooled query of the current chunk $\bar{Q}_i$ and the pooled keys of the memory frames $\bar{K}_j$. Only the top $k$ most relevant historical KV states, denoted by the index set $\mathcal{I}_k$, are activated for the attention computation across chunks:
\begin{equation}
\begin{split}
\mathcal{I}_k &= \mathop{\text{arg\,max}}\limits_{\mathcal{I}, |\mathcal{I}|=k} \sum_{j \in \mathcal{I}} \bar{Q}_i \bar{K}_j^\top, \\
A_i &= \text{Softmax}\left(\frac{Q_i [K_{\mathcal{I}_k}; K_i]^\top}{\sqrt{D}}\right) [V_{\mathcal{I}_k}; V_i],
\end{split}
\label{eq:memflow}
\end{equation}
where $\bar{Q}_i$ and $\bar{K}_j$ represent the condensed token descriptors, via mean pooling for instance, for current chunk $i$ and historical frame $j$, respectively. This selection based on relevance ensures that the model strictly attends to semantically and visually pertinent history, preserving narrative coherence over the long term while significantly curtailing memory and computational overhead.

\section{Supplementary Trend Analysis}\label{app:trend_analysis}

This appendix provides an extended interpretation of Fig.~\ref{fig:dmd-survey-trend}. For the left panel, ``Step Distillation'' receives the broadest attention because it directly compresses denoising depth from approximately $2 \times 50$ passes (with CFG) to few-step regimes (typically around 4 steps), yielding the largest first-order latency reduction. ``Efficient Attention'' is the second-largest branch because it usually provides practical 1--3$\times$ acceleration with relatively mild quality degradation. ``Model Compression'' can also reach comparable acceleration ranges, but quality retention and recovery costs make its deployment trade-offs more sensitive. ``Feature Cache'' is popular due to low entry cost and training-free integration, yet its speedup ceiling is usually lower in practice.

For the middle panel, publication volume increases in nearly all categories, with the most visible post-2024 surge in distribution-distillation-related lines and streaming-associated methods. Within step distillation, ``Distribution Distillation'' now clearly outpaces ``Consistency Distillation'', while ``Adversarial Distillation'' remains comparatively small and is often used as a quality-compensation component. Within efficient attention, sparse methods continue to dominate linear-attention variants because video generation generally requires stronger preservation of motion-critical token interactions than aggressive global compression allows. In model compression, quantization and pruning remain active but secondary compared with distillation and sparse attention in total adoption.

For the right panel, the image domain still has much larger cumulative acceleration literature, whereas the video domain shows a steeper recent growth slope. This pattern is consistent with an emerging transfer-then-specialize cycle: many ideas originate in image diffusion, but video-native constraints (temporal coherence, long-context memory pressure, and cross-step error accumulation) increasingly require dedicated redesign and evaluation protocols.

\section{Supplementary Categorization Notes}\label{app:categorization_supplement}

Due to the manuscript page limit, the main manuscript retains a condensed categorization table that only indexes methods explicitly discussed there. This appendix therefore provides the complete categorization and full supplementary coverage used in our survey. Given the substantial inference overhead of video diffusion models, accelerated sampling algorithms are indispensable for efficient video generation. Table~\ref{tab:video-accel-categorization-full} organizes these methodologies into four paradigms: ``Step Distillation'', ``Efficient Attention'', ``Model Compression'', and ``Cache/Trajectory Optimization''.

\tikzset{
    taxRoot/.style={rectangle, draw=hidden-draw, rounded corners=3pt, text opacity=1, minimum height=2.2em, inner xsep=5pt, inner ysep=4pt, align=center, fill=hidden-draw, text=white, line width=1.0pt, font=\bfseries\small},
    taxL1/.style={rectangle, draw=hidden-draw, rounded corners=3pt, text opacity=1, minimum height=1.9em, inner xsep=4pt, inner ysep=4pt, align=center, fill=hidden-draw!92, text=white, line width=0.95pt, font=\bfseries\normalsize},
    taxL2/.style={rectangle, draw=hidden-draw!85, rounded corners=3pt, text opacity=1, minimum height=1.8em, inner xsep=4pt, inner ysep=4pt, align=center, fill=black!3, text=black, line width=0.85pt, font=\normalsize},
    taxLeaf/.style={rectangle, draw=hidden-draw!80, rounded corners=3pt, text opacity=1, minimum height=2.0em, inner xsep=4pt, inner ysep=4pt, align=left, fill=hidden-pink!55, text=black, line width=0.8pt, font=\footnotesize}
}

\providecommand{\taxleaf}[2]{#1\\[-1pt]{\scriptsize #2}}
\providecommand{\taxbadge}[1]{\textcolor{hidden-draw!95}{\raisebox{0.05ex}{\scriptsize #1}}}

\providecommand{\Description}[1]{}

\begin{table*}[h]
    \centering
    \resizebox{0.99\linewidth}{!}{
    \begin{forest}
        forked edges,
        for tree={
            grow=east,
            reversed=true,
            anchor=base west,
            parent anchor=east,
            child anchor=west,
            base=center,
            font=\footnotesize,
            edge+={darkgray!80, line width=0.9pt},
            s sep=4pt,
            l sep=7pt,
            inner xsep=2pt,
            inner ysep=3pt,
        },
        where level=0{taxRoot, text width=8.8em}{},
        where level=1{taxL1, text width=13.8em}{},
        where level=2{taxL2, text width=11.8em}{},
        where level=3{taxLeaf, text width=16.5em}{},
        [
            Accelerated Sampling for\\Video Diffusion Models
            [
                \taxbadge{\textbullet}~Step Distillation
                [
                    Distribution Distillation
                    [
    Streaming Distillation,
                        content={\taxleaf{Streaming Distillation}{[1--44]}}
                    ]
                    [
    Non-Streaming Distillation,
                        content={\taxleaf{Non-Streaming Distillation}{[45--58]}}
                    ]
                ]
                [
    Consistency Distillation,
                        content={\taxleaf{Consistency Distillation}{[59--81]}}
                ]
                [
                    Adversarial Distillation
                    [
    Combined Distillation,
                        content={\taxleaf{Combined Distillation}{[82--89]}}
                    ]
                    [
    Independent Distillation,
                        content={\taxleaf{Independent Distillation}{[90--92]}}
                    ]
                ]
            ]
            [
                \taxbadge{\ensuremath{\square}}~Efficient Attention
                [
                    Sparse Attention
                    [
    Dynamic Sparsity,
                        content={\taxleaf{Dynamic Sparsity}{[93--119]}}
                    ]
                    [
    Static Sparsity,
                        content={\taxleaf{Static Sparsity}{[120--137]}}
                    ]
                ]
                [
                    Linear Attention
                    [
    Training-Based,
                        content={\taxleaf{Training-Based}{[138--148]}}
                    ]
                ]
            ]
            [
                \taxbadge{\ensuremath{\triangle}}~Model Compression
                [
                    Quantization
                    [
    Quantization-Aware Training,
                        content={\taxleaf{Quantization-Aware Training}{[149--155]}}
                    ]
                    [
    Post-Training Quantization,
                        content={\taxleaf{Post-Training Quantization}{[156--167]}}
                    ]
                ]
                [
    VAE Compression,
                        content={\taxleaf{VAE Compression}{[168--175]}}
                ]
                [
                    Pruning
                    [
    Token Pruning,
                        content={\taxleaf{Token Pruning}{[176--185]}}
                    ]
                    [
    Channel Pruning,
                        content={\taxleaf{Channel Pruning}{[186]}}
                    ]
                    [
    Model Pruning,
                        content={\taxleaf{Model Pruning}{[187--193]}}
                    ]
                ]
            ]
            [
                \taxbadge{\ensuremath{\diamond}}~Cache and Trajectory Optimization
                [
                    Cache
                    [
    Feature Cache,
                        content={\taxleaf{Feature Cache}{[194--228]}}
                    ]
                    [
    KV Cache,
                        content={\taxleaf{KV Cache}{[229--248]}}
                    ]
                ]
                [
                    Latent Trajectory Tricks
                    [
    Noise and State Modification,
                        content={\taxleaf{Noise and State Modification}{[249--253]}}
                    ]
                    [
    Trajectory Modification,
                        content={\taxleaf{Trajectory Modification}{[69, 254--275]}}
                    ]
                    [
    Parallel Computation,
                        content={\taxleaf{Parallel Computation}{[111, 276--280]}}
                    ]
                ]
                [
                    Other Efficiency Methods,
                    content={\taxleaf{Other Efficiency Methods}{[128, 281--285]}}
                ]
            ]
        ]
    \end{forest}

    }
\nocite{https://doi.org/10.48550/arxiv.2509.25161,https://doi.org/10.48550/arxiv.2506.03099,Yin_2025,yin2024slow,https://doi.org/10.48550/arxiv.2511.01266,Zhou_2025,https://doi.org/10.48550/arxiv.2510.08131,chen2024diffusion,https://doi.org/10.48550/arxiv.2405.11473,https://doi.org/10.48550/arxiv.2412.14169,https://doi.org/10.48550/arxiv.2506.09350,https://doi.org/10.48550/arxiv.2508.13009,https://doi.org/10.48550/arxiv.2510.02283,lu2025reward,Henschel_2025,https://doi.org/10.48550/arxiv.2506.08009,https://doi.org/10.48550/arxiv.2408.14837,https://doi.org/10.48550/arxiv.2504.13074,Sun_2025,Sun_2025_ARDiffusion,huang2025live,https://doi.org/10.48550/arxiv.2510.03198,https://doi.org/10.48550/arxiv.2506.01380,https://doi.org/10.48550/arxiv.2509.22622,yi2025deep,wang2025restdiffusionbasedrealtimeendtoend,kodaira2025streamditrealtimestreamingtexttovideo,wang2025longdwmcrossgranularitydistillationbuilding,yi2025magicinfinitegeneratinginfinitetalking,guo2025endtoendtrainingautoregressivevideo,yu2025videossmautoregressivelongvideo,zhu2025memorizeandgeneratelongtermconsistencyrealtime,sun2025streamavatarstreamingdiffusionmodels,xiao2025knotforcingtamingautoregressive,ryu2025shragaframeworkforcombininghumaninspired,zhang2025infvsrbreakinglengthlimits,chen2024streamingvideodiffusiononline,du2025raprealtimeaudiodrivenportrait,zhang2025generativepretrainedautoregressivediffusion,zhang2025blockvidblockdiffusionhighquality,yuan2025lumos1autoregressivevideogeneration,liang2025lookingbackwardstreamingvideotovideo,yu2025autorefinerimprovingautoregressivevideo,zhu2026causal,sun2025worldplay,gu2025blade,karnewar2025neodragon,https://doi.org/10.48550/arxiv.2503.06674,https://doi.org/10.48550/arxiv.2412.05899,https://doi.org/10.48550/arxiv.2510.12747,https://doi.org/10.48550/arxiv.2502.07701,https://doi.org/10.48550/arxiv.2211.11018,shao2025magicdistillationweaktostrongvideodistillation,nie2026transitionmatchingdistillationfast,liu2025equivariancefastsamplingvideo,gao2025seedance10exploringboundaries,bedel2023dreamrdiffusiondrivencounterfactualexplanation,chen2023diffusiontalkerpersonalizationaccelerationspeechdriven,teng2025gfixperceptuallyenhancedgaussian,https://doi.org/10.48550/arxiv.2406.06890,zhang2025mobilei2v,lv2025dcm,https://doi.org/10.48550/arxiv.2504.11143,https://doi.org/10.48550/arxiv.2312.09109,https://doi.org/10.48550/arxiv.2505.16239,Mao_2025,https://doi.org/10.48550/arxiv.2508.06082,https://doi.org/10.48550/arxiv.2410.05677,Xu_2025,Zhang_2025_AdaDiff,Zhang_2025_FastVideoEdit,https://doi.org/10.48550/arxiv.2412.15689,https://doi.org/10.48550/arxiv.2405.18750,wang2024phased,Liu_2025_UltraVSR,Liu_2025_TimestepCache,zhang2025vividfacehighqualityefficientonestep,hu2024efficienttextdrivenmotiongeneration,yin2024lm2dlyricsmusicdrivendance,dai2024motionlcmrealtimecontrollablemotion,dao2025improvedtrainingtechniquelatent,yu2025lliaenablinglowlatency,zhang2025turbodiffusionacceleratingvideodiffusion,rCM,https://doi.org/10.48550/arxiv.2507.18569,https://doi.org/10.48550/arxiv.2503.19462,cheng2025pose,https://doi.org/10.48550/arxiv.2412.06578,https://doi.org/10.48550/arxiv.2403.12706,guo2024realtimeonestepdiffusionbasedexpressive,jiang2025motionpcmrealtimemotionsynthesis,xue2025moganimprovingmotionquality,https://doi.org/10.48550/arxiv.2501.08316,https://doi.org/10.48550/arxiv.2509.16507,https://doi.org/10.48550/arxiv.2406.04324,https://doi.org/10.48550/arxiv.2505.18809,wu2025usv,https://doi.org/10.48550/arxiv.2510.02617,https://doi.org/10.48550/arxiv.2506.23858,https://doi.org/10.48550/arxiv.2502.01776,shmilovich2025liteattention,https://doi.org/10.48550/arxiv.2509.01085,https://doi.org/10.48550/arxiv.2505.14708,https://doi.org/10.48550/arxiv.2510.18692,https://doi.org/10.48550/arxiv.2508.21058,https://doi.org/10.48550/arxiv.2505.18875,https://doi.org/10.48550/arxiv.2509.16518,https://doi.org/10.48550/arxiv.2502.21079,https://doi.org/10.48550/arxiv.2505.13389,sparge_attn,https://doi.org/10.48550/arxiv.2412.20404,https://doi.org/10.48550/arxiv.2505.22918,https://doi.org/10.48550/arxiv.2504.12027,https://doi.org/10.48550/arxiv.2305.13077,kahatapitiya2024objectcentricdiffusionefficientvideo,zhang2025trainingfreeefficientvideogeneration,qiao2025flashomniunifiedsparseattention,chen2025rainfusion20temporalspatialawarenesshardwareefficient,xu2025xattentionblocksparseattention,jabri2023scalableadaptivecomputationiterative,gao2025ditvrzeroshotdiffusiontransformer,https://doi.org/10.48550/arxiv.2312.06662,https://doi.org/10.48550/arxiv.2506.19852,https://doi.org/10.48550/arxiv.2506.16054,https://doi.org/10.48550/arxiv.2502.06155,https://doi.org/10.48550/arxiv.2506.03065,Wu_2023,Ruan_2023,Lin_2022,Shi_2024_MotionI2V,Shrivastava_2024,ghafoorian2025attention,li2025radialattentiononlogn,hassani2025generalizedneighborhoodattentionmultidimensional,meng2025holocineholisticgenerationcinematic,team2025longcat,lin2024opensoraplanopensourcelarge,xu2024mscmultiscalespatiotemporalcausal,wang2024qihoot2xefficientproxytokenizeddiffusion,ren2025groupingfirstattendingsmartly,https://doi.org/10.48550/arxiv.2509.24695,huang2025linvideo,https://doi.org/10.48550/arxiv.2509.24006,wang2024unianimatetamingunifiedvideo,hu2024zigmaditstylezigzagmamba,ma2025consistentcontrollableimageanimation,ghafoorian2026rehyatrecurrenthybridattention,zhang2025efficientlongdurationtalkingvideo,zatsarynna2025mantadiffusionmambaefficient,wang2024opticalflowrepresentationalignment,mo2024scalingdiffusionmambabidirectional,https://doi.org/10.48550/arxiv.2509.23681,https://doi.org/10.48550/arxiv.2505.11497,https://doi.org/10.48550/arxiv.2506.04648,https://doi.org/10.48550/arxiv.2505.22167,feng2025s,feng2025pmqveprogressivemultiframequantization,cao2025simplelowbitquantizationframework,https://doi.org/10.48550/arxiv.2505.18663,Chen_2025,https://doi.org/10.48550/arxiv.2406.02540,zhang2025sageattentionaccurate8bitattention,zhang2026sageattention3microscalingfp4attention,zhang2025sageattention2efficientattentionthorough,yang2025lrqditlogrotationposttrainingquantization,feng2025s2qvditaccuratequantizedvideo,liu2024taqdittimeawarequantizationdiffusion,liu2025clqcrosslayerguidedorthogonalbased,chai2025quantvsrlowbitposttrainingquantization,https://doi.org/10.48550/arxiv.2503.06564,zhao2024cvvaecompatiblevideovae,wu2025improvedvideovaelatentvideo,chen2024odvaeomnidimensionalvideocompressor,li2024wfvaeenhancingvideovae,cheng2025leanvaeultraefficientreconstructionvae,wang2024vidtwinvideovaedecoupled,liu2025hivaeefficientvideoautoencoding,chen2025dcvideogenefficientvideogeneration,https://doi.org/10.48550/arxiv.2412.11706,Piergiovanni_2023,he2025fulldit2efficientincontextconditioning,zhang2025framecontextpackingdrift,liu2025astraeatokenwiseaccelerationframework,yuan2025vgdfrdiffusionbasedvideogeneration,gu2025longcontextautoregressivevideomodeling,bi2025efficientdiffusionbased3dhuman,yang2025scriptgraphstructuredqueryconditionedsemantic,shen2025fastviddynamicdensitypruning,https://doi.org/10.48550/arxiv.2412.07583,wu2025individualcontentmotiondynamics,kim2025vip,wu2025tamingdiffusiontransformerefficient,yan2024animatedstickersbringingstickers,zou2025turbovaedfaststabletransfer,li2025asymmetricvaeonestepvideo,f3_pruning,https://doi.org/10.48550/arxiv.2510.05367,https://doi.org/10.48550/arxiv.2507.02860,https://doi.org/10.48550/arxiv.2408.12588,Ceylan_2023,https://doi.org/10.48550/arxiv.2411.02397,https://doi.org/10.48550/arxiv.2410.19355,sun2024unveilingredundancydiffusiontransformers,zheng2025letfeaturesdecidesolvers,fan2025taocachestructuremaintainedvideogeneration,wang2025drivegen3dboostingfeedforwarddriving,liu2025fastcachefastcachingdiffusion,huang2025enerverseenvisioningembodiedfuture,zou2025acceleratingdiffusiontransformerstokenwise,zheng2025dynamictryontamingvideo,zou2025rethinkingtokenwisefeaturecaching,wu2025quantcacheadaptiveimportanceguidedquantization,liu2025reusingforecastingacceleratingdiffusion,zheng2025forecastcalibratefeaturecaching,feng2025hicachetrainingfreeaccelerationdiffusion,song2025herohierarchicalextrapolationrefresh,cui2025bwcacheacceleratingvideodiffusion,bu2025dicacheletdiffusionmodel,peng2025ertacacheerrorrectificationtimesteps,wei2025mixcachemixtureofcachevideodiffusion,ma2025magcachefastvideogeneration,chen2026sortblocksimilarityawarefeaturereuse,ji2025blockwiseadaptivecachingaccelerating,yang2025evctrlefficientcontroladapter,sun2024flexcacheflexibleapproximatecache,ye2025supergenefficientultrahighresolutionvideo,adnan2025foresightadaptivelayerreuse,yang2025rethinkingvideotokenizationconditioned,kong2025tamingflowbasedi2vmodels,shi2025aquariusfamilyindustrylevelvideo,li2025selfguidanceboostingflowdiffusion,https://doi.org/10.48550/arxiv.2411.16375,https://doi.org/10.48550/arxiv.2406.10981,jiang2025lovicefficientlongvideo,guo2025longcontexttuningvideo,xing2024live2difflivestreamtranslation,wang2025liftvsrliftingimagediffusion,zhang2025egolcdegocentricvideogeneration,zhang2026pretrainingframepreservationautoregressive,ye2025yanfoundationalinteractivevideo,li2026packcachetrainingfreeaccelerationmethod,feng2025vidarcembodiedvideodiffusion,wen2025dvladiffusionvisionlanguageactionmodel,song2025physicalautoregressivemodelrobotic,kong2025causnvsautoregressivemultiviewdiffusion,ji2025memflowflowingadaptivememory,yu2025videomarautoregressivevideogeneratio,he2025posegenincontextlorafinetuning,pfp,li2024dicodediffusioncompresseddeeptokens,po2025longcontextstatespacevideoworld,https://doi.org/10.48550/arxiv.2310.15169,https://doi.org/10.48550/arxiv.2307.10373,https://doi.org/10.48550/arxiv.2304.08477,hoeg2024streamingdiffusionpolicyfast,liu2025free4dtuningfree4dscene,https://doi.org/10.48550/arxiv.2403.14148,https://doi.org/10.48550/arxiv.2503.18940,Esser_2023,https://doi.org/10.48550/arxiv.2410.05954,https://doi.org/10.48550/arxiv.2502.05179,zhu2025acceleratingdiffusionsamplingexploiting,liu2025gesturelsmlatentshortcutbased,kraljusic2025searchbasedrobotmotionplanning,sabour2024alignstepsoptimizingsampling,zhuang2025timestepmasterasymmetricalmixture,zhou2025unitenhancementguidanceframework,chen2026efficientcameracontrolledvideogeneration,ling2024motionclonetrainingfreemotioncloning,liu2025multimotion,wu2025attentionscalephasealignedrotary,wu2025ditpainterefficientvideoinpainting,li2025arlonboostingdiffusiontransformers,kim2025ondevicesoraenablingtrainingfree,cheng2025srdiffusionacceleratevideodiffusion,oshima2025inferencetimetexttovideoalignmentdiffusion,cheng2025adaptivebeginofvideotokensautoregressive,cao2025video,Zhang_2023,bandyopadhyay2025blockcascadingtrainingfree,chen2025dbspacceleratingsparseattention,wang2025pipeditacceleratingdiffusiontransformers,fang2024xditinferenceenginediffusion}
\caption{A complete categorization of accelerated sampling algorithms for video diffusion models. \textit{Note: Bracketed IDs inside nodes are local categorization indices for compact display.}}
    \Description{A tree summarizing step distillation families, attention efficiency, model compression, caching/trajectory techniques, and related efficiency tricks for video diffusion models.}
    \label{tab:video-accel-categorization-full}
\end{table*}

Accelerated sampling algorithms~\cite{lichen2025zigzag,bai2026weaktostrong,shao2026core2,bai2026optimizing,zhou2025golden} can be categorized into three distinct classes based on their optimization objective: those reducing NFE, those minimizing per-step overhead, and hybrid approaches that address both simultaneously.

\hspace*{-\parindent}\textbf{Minimizing Number of Function Evaluations.} Given the iterative sampling nature of diffusion models, reducing NFE constitutes a primary strategy for acceleration. Algorithms in this category include ``Distribution Distillation'', ``Consistency Distillation'', ``Adversarial Distillation'', and various ``Latent Trajectory Tricks''. While all distillation methods aim to align the quality of few-step generation with that of multi-step baselines, ``Distribution Distillation'' exhibits superior scaling behavior. The ``Latent Trajectory Tricks'' category encompasses advanced solvers and approximation schemes for CFG. Prominent examples adaptable to both image and video domains include DDIM~\cite{ddim} and DPM-Solver~\cite{dpm_solver}. Related image-domain studies further explore self-reflection, weak-to-strong guidance, adaptive distillation, and learned noise design for accelerated sampling.

\hspace*{-\parindent}\textbf{Reducing Per-Step Overhead.} Strategies for mitigating per-step computational overhead include ``Sparse Attention'', ``Linear Attention'', ``Quantization'', and ``Pruning''. Among these, ``Sparse Attention'' and ``Linear Attention'' accelerate inference by reducing the algorithmic complexity of the attention mechanism. In contrast, ``Quantization'' enhances efficiency by compressing weights and activations into low-bit representations, while ``Pruning'' reduces model size by eliminating redundant components, such as specific blocks or individual weights. Related image-domain work also suggests that structured sparsity remains promising for efficient diffusion transformers~\cite{li2026pisa}.

\hspace*{-\parindent}\textbf{Joint Optimization of NFE and Overhead.} Approaches for jointly minimizing NFE and per-step overhead are primarily classified into two paradigms: ``Feature Cache'' and hybrid integration strategies. ``Feature Cache'' primarily mitigates per-step overhead by bypassing redundant computations in specific model layers and does not directly change nominal NFE in its standard form. Some hybrid systems additionally combine cache reuse with step-skipping or distilled trajectories, which can further reduce effective sampling cost. Furthermore, combining distinct acceleration algorithms has proven highly effective, most notably the integration of ``Step Distillation'' with ``Efficient Attention'' or ``Quantization''. For instance, the open-source FastVideo~\cite{fastvideo} project integrates DMD~\cite{dmd,dmd2} and VSA~\cite{https://doi.org/10.48550/arxiv.2505.13389} to achieve substantial acceleration. Related video-generation work has also begun to jointly reduce sampling steps and model size within a single acceleration framework~\cite{shitong2026fastlightgen}. SageAttention 3~\cite{zhang2026sageattention3microscalingfp4attention} likewise combines ``Efficient Attention'' and ``Quantization'' to maximize computational efficiency.

\section{Supplementary Evaluation and Dataset Analysis}\label{app:evaluation_dataset_supplement}

\subsection{Acceleration and Quality Metrics}
\begin{table*}[th]
    \centering
    \small
    \caption{Acceleration and quality metrics.}
    \label{tab:accel_quality_metrics}
    \begin{tabular}{@{}p{0.24\textwidth} p{0.16\textwidth} p{0.52\textwidth}@{}}
        \toprule
        Metric & Type & Meaning \\
        \midrule
        Speedup & Acceleration & Relative runtime reduction against a baseline. \\
        Inference Time/Latency & Acceleration & Wall-clock time per video, or per denoising step (ms/s). \\
        Time-to-first-output & Acceleration & Time until the first token/frame is produced. \\
        Throughput & Acceleration & Samples or frames generated per second. \\
        NFE & Acceleration & Number of function evaluations during sampling. \\
        FLOPs/MACs/OPS & Acceleration & Computational cost per inference or per step. \\
        Model Size (\#Params) & Acceleration & Parameter count or storage size of the model. \\
        Memory / VRAM & Acceleration & Peak memory footprint during inference. \\
        KV Cache Size   & Acceleration & Cache size for KV states. \\
        Sparsity / Keep Ratio & Acceleration & Report whether this denotes pruned ratio (sparsity) or retained ratio (keep ratio). \\
        W$b_w$A$b_a$ & Acceleration & Activation values use $b_a$ bits, and weights use $b_w$ bits in quantization. \\
        Sequence Length & Acceleration & Sequence length used in evaluation or scaling. \\
        \midrule
        VBench~\cite{vbench,vbench++} & Quality & Video benchmark suite (includes T2V/I2V). \\
        FVD~\cite{fvd} & Quality & Fr\'echet Video Distance for distributional video fidelity. \\
        Frame-wise FID~\cite{fid} & Quality & Fr\'echet Inception Distance for frame/image fidelity. \\
        Frame-wise IS~\cite{is} & Quality & Inception Score for quality/diversity. \\
        Frame-wise CLIPScore & Quality & Text--video/image alignment measured by CLIP. \\
        Frame-wise LPIPS~\cite{LPIPS} & Quality & Perceptual similarity in deep feature space. \\
        Frame-wise PSNR & Quality & Signal fidelity metric (higher is better). \\
        Frame-wise SSIM & Quality & Structural similarity index. \\
        MSE / L2 & Quality & Pixel-wise reconstruction error. \\
        DOVER~\cite{dover_score} & Quality & Video quality assessment. \\
        Frame-wise HPSv2~\cite{HPSV2} & Quality & Human Preference Score. \\
        Frame-wise AES~\cite{AES} & Quality & Aesthetic quality predictor score. \\
        ImageReward~\cite{Imagereward} & Quality & Validate human preferences for each frame. \\
        T2V-CompBench~\cite{t2vcompbench} & Quality & Compositional Text-to-video Generation benchmark. \\
        FasterVQA~\cite{fastvqa} & Quality & Video quality assessment with fragment sampling. \\
        FAD~\cite{fad} & Quality & Fr\'echet Audio Distance for audio-related fidelity. \\
        Sync-C~\cite{sync_c} & Quality & Evaluate lip motion synchronization; higher is better. \\
        Sync-D~\cite{sync_c} & Quality & Evaluate lip motion synchronization; lower is better. \\
        \bottomrule
    \end{tabular}
\end{table*}

\textbf{Acceleration Metrics.} Metrics for evaluating the generation speed and efficiency of video diffusion models are largely adapted from established machine learning benchmarks. As summarized in Table~\ref{tab:accel_quality_metrics}, these indicators can be categorized into two groups. The first comprises hardware-dependent metrics that necessitate empirical benchmarking on specific graphics processing unit (GPU) clusters, including throughput, inference time, time-to-first-output, speedup, and peak video random-access memory (VRAM) utilization. The second group consists of hardware-agnostic, theoretical properties derived from the model architecture, such as NFE, FLOPs/MACs, model size, KV cache size, and sequence length. Furthermore, domain-specific metrics are increasingly utilized to characterize specialized optimizations, such as W$b_w$A$b_a$ in model quantization~\cite{https://doi.org/10.48550/arxiv.2505.18663,Chen_2025,https://doi.org/10.48550/arxiv.2406.02540,pan2025diga3dcoarsetofinediffusionalpropagation,zhang2025sageattentionaccurate8bitattention,zhang2026sageattention3microscalingfp4attention} and sparsity in efficient attention mechanisms~\cite{https://doi.org/10.48550/arxiv.2505.18809,wu2025usv,https://doi.org/10.48550/arxiv.2510.02617,https://doi.org/10.48550/arxiv.2506.23858,https://doi.org/10.48550/arxiv.2502.01776,shmilovich2025liteattention,https://doi.org/10.48550/arxiv.2509.01085}.

\hspace*{-\parindent}\textbf{Quality Metrics.} The evaluation of sampling acceleration in video diffusion models generally leverages the diverse set of metrics established for video generation. As summarized in Table~\ref{tab:accel_quality_metrics}, early research primarily relied on traditional benchmarks such as Fr\'echet Video Distance (FVD)~\cite{fvd} and various frame-wise methodologies. These approaches simplify video evaluation by treating the sequence as a collection of independent images, averaging individual frame scores, such as Fr\'echet Inception Distance (FID)~\cite{fid}, Inception Score (IS)~\cite{is}, Learned Perceptual Image Patch Similarity (LPIPS)~\cite{LPIPS}, peak signal-to-noise ratio (PSNR), structural similarity index (SSIM), HPSv2~\cite{HPSV2}, AES~\cite{AES}, and ImageReward~\cite{Imagereward}, to represent overall quality. However, these static metrics often overlook temporal coherence and motion dynamics. To address these limitations, motion-aware benchmarks like DOVER~\cite{dover_score} and FasterVQA~\cite{fastvqa} were developed to provide a more holistic assessment of both visual fidelity and temporal consistency. Currently, VBench~\cite{vbench,vbench++} stands as the most comprehensive evaluation suite, utilizing a multi-dimensional set of sub-metrics to evaluate video quality, whereas T2V-CompBench~\cite{t2vcompbench} is utilized less frequently. Finally, for models supporting multimodal audio-video synthesis, specialized metrics such as Fr\'echet Audio Distance (FAD)~\cite{fad} and lip-synchronization scores (e.g., Sync-C~\cite{sync_c} and Sync-D~\cite{sync_c}) are employed to quantify cross-modal alignment.

\subsection{Evaluation for Accelerated Sampling of Video Diffusion Models}
\begin{table*}[th]
    \centering
    \small
    \caption{Metric evaluation for accelerated sampling of video diffusion models.}
    \label{tab:quality_metric_usage_popular_methods}
    \begin{tabular}{@{}p{0.24\textwidth} p{0.54\textwidth} p{0.10\textwidth}@{}}
    \toprule
    Metric & Example Methods & Percentage \\
    \midrule
    VBench & MagicDistillation~\cite{shao2025magicdistillationweaktostrongvideodistillation}; StreamDiT~\cite{kodaira2025streamditrealtimestreamingtexttovideo}; LongDWM~\cite{wang2025longdwmcrossgranularitydistillationbuilding}; ViP~\cite{kim2025vip}; EquiVDM~\cite{liu2025equivariancefastsamplingvideo} & 36.0\% \\
    FVD & OSA-LCM~\cite{guo2024realtimeonestepdiffusionbasedexpressive}; REST~\cite{wang2025restdiffusionbasedrealtimeendtoend}; MagicDistillation~\cite{shao2025magicdistillationweaktostrongvideodistillation}; LongDWM~\cite{wang2025longdwmcrossgranularitydistillationbuilding}; MagicInfinite~\cite{yi2025magicinfinitegeneratinginfinitetalking} & 28.2\% \\
    Frame-wise FID & OSA-LCM~\cite{guo2024realtimeonestepdiffusionbasedexpressive}; REST~\cite{wang2025restdiffusionbasedrealtimeendtoend}; MagicDistillation~\cite{shao2025magicdistillationweaktostrongvideodistillation}; LongDWM~\cite{wang2025longdwmcrossgranularitydistillationbuilding}; MagicInfinite~\cite{yi2025magicinfinitegeneratinginfinitetalking} & 27.3\% \\
    Frame-wise IS & AYS~\cite{sabour2024alignstepsoptimizingsampling}; TimeStep Master~\cite{zhuang2025timestepmasterasymmetricalmixture}; Qihoo-t2x~\cite{wang2024qihoot2xefficientproxytokenizeddiffusion}; Audiogen-omni~\cite{wang2025audiogenomniunifiedmultimodaldiffusion}; TaQ-DiT~\cite{liu2024taqdittimeawarequantizationdiffusion} & 10.4\% \\
    Frame-wise CLIPScore & EquiVDM~\cite{liu2025equivariancefastsamplingvideo}; Im-3d~\cite{melaskyriazi2024im3diterativemultiviewdiffusion}; LFDOS~\cite{zheng2025letfeaturesdecidesolvers}; MPQ-DMv2~\cite{pan2025diga3dcoarsetofinediffusionalpropagation}; FullDiT2~\cite{he2025fulldit2efficientincontextconditioning} & 24.9\% \\
    Frame-wise LPIPS & VividFace~\cite{zhang2025vividfacehighqualityefficientonestep}; Lyra~\cite{bahmani2025lyragenerative3dscene}; Seaweed-7b~\cite{seawead2025seaweed7bcosteffectivetrainingvideo}; Memorize-and-Generate~\cite{zhu2025memorizeandgeneratelongtermconsistencyrealtime}; LFDOS~\cite{zheng2025letfeaturesdecidesolvers} & 26.1\% \\
    Frame-wise PSNR & VividFace~\cite{zhang2025vividfacehighqualityefficientonestep}; Lyra~\cite{bahmani2025lyragenerative3dscene}; EquiVDM~\cite{liu2025equivariancefastsamplingvideo}; Seaweed-7b~\cite{seawead2025seaweed7bcosteffectivetrainingvideo}; Memorize-and-Generate~\cite{zhu2025memorizeandgeneratelongtermconsistencyrealtime} & 29.4\% \\
    Frame-wise SSIM & VividFace~\cite{zhang2025vividfacehighqualityefficientonestep}; Lyra~\cite{bahmani2025lyragenerative3dscene}; EquiVDM~\cite{liu2025equivariancefastsamplingvideo}; Memorize-and-Generate~\cite{zhu2025memorizeandgeneratelongtermconsistencyrealtime}; LFDOS~\cite{zheng2025letfeaturesdecidesolvers} & 25.6\% \\
    MSE / L2 & Lyra~\cite{bahmani2025lyragenerative3dscene}; MotionLCM~\cite{dai2024motionlcmrealtimecontrollablemotion}; VideoSSM~\cite{yu2025videossmautoregressivelongvideo}; Memorize-and-Generate~\cite{zhu2025memorizeandgeneratelongtermconsistencyrealtime}; LFDOS~\cite{zheng2025letfeaturesdecidesolvers} & 16.6\% \\
    ImageReward & LFDOS~\cite{zheng2025letfeaturesdecidesolvers}; SpargeAttention~\cite{sparge_attn}; SageAttention~\cite{zhang2025sageattentionaccurate8bitattention}; SageAttention3~\cite{zhang2026sageattention3microscalingfp4attention}; SageAttention2~\cite{zhang2025sageattention2efficientattentionthorough} & 6.2\% \\
    DOVER & InfVSR~\cite{zhang2025infvsrbreakinglengthlimits}; LiftVSR~\cite{wang2025liftvsrliftingimagediffusion}; QuantVSR~\cite{chai2025quantvsrlowbitposttrainingquantization}; QuantCache~\cite{wu2025quantcacheadaptiveimportanceguidedquantization}; Foresight~\cite{adnan2025foresightadaptivelayerreuse} & 4.5\% \\
    Frame-wise AES & FullDiT2~\cite{he2025fulldit2efficientincontextconditioning}; StreamAvatar~\cite{sun2025streamavatarstreamingdiffusionmodels}; Self-Guidance~\cite{li2025selfguidanceboostingflowdiffusion} & 4.0\% \\
    Frame-wise HPSv2 & Self-Guidance~\cite{li2025selfguidanceboostingflowdiffusion}; AutoRefiner~\cite{yu2025autorefinerimprovingautoregressivevideo} & 2.4\% \\
    Sync-C & REST~\cite{wang2025restdiffusionbasedrealtimeendtoend}; MagicInfinite~\cite{yi2025magicinfinitegeneratinginfinitetalking}; StreamAvatar~\cite{sun2025streamavatarstreamingdiffusionmodels}; RAP~\cite{du2025raprealtimeaudiodrivenportrait} & 2.1\% \\
    Sync-D & REST~\cite{wang2025restdiffusionbasedrealtimeendtoend}; MagicInfinite~\cite{yi2025magicinfinitegeneratinginfinitetalking}; StreamAvatar~\cite{sun2025streamavatarstreamingdiffusionmodels}; RAP~\cite{du2025raprealtimeaudiodrivenportrait} & 1.9\% \\
    FasterVQA & VividFace~\cite{zhang2025vividfacehighqualityefficientonestep} & 0.7\% \\
    FAD & ProAV-DiT~\cite{sun2025proavditprojectedlatentdiffusion}; Audiogen-omni~\cite{wang2025audiogenomniunifiedmultimodaldiffusion} & 0.7\% \\
    \bottomrule
    \end{tabular}
\end{table*}

The evaluation of accelerated sampling methods for video diffusion models can be further informed by an analysis of metric adoption patterns. Based on the categorization in Table~\ref{tab:accel_quality_metrics}, Table~\ref{tab:quality_metric_usage_popular_methods} illustrates distinct usage trends in current research. Our survey statistics identify VBench~\cite{vbench,vbench++} as the most prevalent benchmark suite (36.0\%), signaling a methodological shift from single-score proxies toward multi-dimensional evaluation. Nevertheless, traditional quality metrics remain ubiquitous, including frame-wise PSNR (29.4\%), FVD~\cite{fvd} (28.2\%), frame-wise FID~\cite{fid} (27.3\%), frame-wise LPIPS~\cite{LPIPS} (26.1\%), frame-wise SSIM (25.6\%), and frame-wise CLIPScore (24.9\%). This distribution underscores the importance of VBench~\cite{vbench,vbench++} for comprehensive video assessment while reflecting the continued utility of FVD~\cite{fvd} in verifying data distribution alignment. Specialized metrics are reserved for niche subdomains, such as DOVER~\cite{dover_score} (4.5\%) and FasterVQA~\cite{fastvqa} (0.7\%) for video quality assessment, ImageReward~\cite{Imagereward} (6.2\%), HPSv2~\cite{HPSV2} (2.4\%), and AES~\cite{AES} (4.0\%) for preference alignment, and FAD~\cite{fad} (0.7\%) alongside Sync-C~\cite{sync_c}/Sync-D~\cite{sync_c} (2.1\%/1.9\%) for audio-visual or talking-head synchronization. Such patterns suggest that while task-specific evaluation remains fragmented, acceleration research still relies heavily on general-purpose image and video metrics.

From a methodological standpoint, these metrics serve as complementary but imperfect probes of generation quality. FVD~\cite{fvd} assesses video-level distributional similarity and remains a standard for temporal realism. However, it is sensitive to feature extractor domain mismatches and may deviate from human preference in editing or stylized generation tasks. Frame-wise FID~\cite{fid} and IS~\cite{is} maintain historical significance for benchmarking against prior work, yet they overlook temporal continuity and may overestimate quality when motion consistency degrades. Similarly, frame-wise LPIPS~\cite{LPIPS}, PSNR, SSIM, and MSE/L2 are effective in reference-based settings like restoration or controllable generation, where they quantify perceptual or pixel fidelity, but they offer limited utility for open-ended text-to-video generation lacking ground truth. While frame-wise CLIPScore, HPSv2~\cite{HPSV2}, AES~\cite{AES}, and ImageReward~\cite{Imagereward} capture semantic alignment and aesthetic quality, they fail to evaluate motion dynamics or the instruction-following capabilities of text-and-image-to-video (TI2V) models. VBench~\cite{vbench,vbench++} provides a practical and comprehensive assessment by aggregating consistency, motion, and aesthetics. Nevertheless, its multi-submetric design can introduce reporting heterogeneity. Furthermore, because different sub-metrics vary in sensitivity, the average score may be disproportionately influenced by certain dimensions. For instance, the dynamic degree metric is highly sensitive, and substantial gains in VBench~\cite{vbench,vbench++} scores can be achieved simply by increasing motion, even if other dimensions show negligible improvement. Finally, although DOVER~\cite{dover_score} and FasterVQA~\cite{fastvqa} offer superior assessment compared to frame-averaging methods, their task-dependency limits their frequency in general acceleration benchmarks. Consequently, a rigorous evaluation should combine a holistic video-level metric, such as VBench~\cite{vbench,vbench++}, with relevant task-specific scores.

\subsection{Dataset for Accelerated Sampling of Video Diffusion Models}
\begin{table*}[th]
    \centering
    \small
    \caption{Training dataset usage in training-based acceleration methods for video diffusion models.}
    \label{tab:training_dataset_usage_popular_methods}
    \resizebox{1.0\textwidth}{!}{
        \begin{tabular}{@{}p{0.17\textwidth} p{0.39\textwidth} p{0.08\textwidth} p{0.10\textwidth} p{0.11\textwidth} p{0.11\textwidth}@{}}
        \toprule
        Dataset & Example Methods & Percentage & Quantity & Frames & Resolution \\
        \midrule
        Customized Dataset & MCM~\cite{https://doi.org/10.48550/arxiv.2406.06890}; Neodragon~\cite{karnewar2025neodragon}; Open-Sora~\cite{https://doi.org/10.48550/arxiv.2412.20404}; TalkingMachines~\cite{https://doi.org/10.48550/arxiv.2506.03099}; MotionStream~\cite{https://doi.org/10.48550/arxiv.2511.01266} & 40.0\% & - & - & - \\
        WebVid-10M~\cite{webvid10m} & Open-Sora~\cite{https://doi.org/10.48550/arxiv.2412.20404}; VideoLCM~\cite{https://doi.org/10.48550/arxiv.2312.09109}; Latent-Shift~\cite{https://doi.org/10.48550/arxiv.2304.08477}; MagicVideo~\cite{https://doi.org/10.48550/arxiv.2211.11018}; AnimateDiff-Lightning~\cite{https://doi.org/10.48550/arxiv.2403.12706} & 7.5\% & 10.7M & $\sim$375 & 596$\times$336 \\
        OpenVid-1M~\cite{nan2024openvid} & OSV~\cite{Mao_2025}; MotionStream~\cite{https://doi.org/10.48550/arxiv.2511.01266}; SwiftVideo~\cite{https://doi.org/10.48550/arxiv.2508.06082} & 6.2\% & 1.0M & $\sim$114 & >512$\times$512 \\
        UCF101~\cite{soomro2012ucf101} & PVDM~\cite{Yu_2023}; TR-DQ~\cite{https://doi.org/10.48550/arxiv.2503.06564}; LVDM~\cite{https://doi.org/10.48550/arxiv.2211.13221} & 6.2\% & 13.3K+ & $\sim$75 & 320$\times$240 \\
        Mixkit~\cite{Mixkit} & Open-Sora~\cite{https://doi.org/10.48550/arxiv.2412.20404}; VORTA~\cite{https://doi.org/10.48550/arxiv.2505.18809}; Efficient-vDiT~\cite{https://doi.org/10.48550/arxiv.2502.06155} & 3.8\% & 46K+ & $\sim$462 & 1080$\times$1920 \\
        SkyTimelapse~\cite{SkyTimelapse} & PVDM~\cite{Yu_2023}; Ca2-VDM~\cite{https://doi.org/10.48550/arxiv.2411.16375}; LVDM~\cite{https://doi.org/10.48550/arxiv.2211.13221} & 3.8\% & 35K & $\sim$32 & 640$\times$360 \\
        \bottomrule
        \end{tabular}
    }
\end{table*}

Datasets are a critical component for accelerating video diffusion models whenever the acceleration pipeline requires training, because the model must preserve visual fidelity and temporal consistency after modifications to the sampling trajectory, attention pattern, precision, or architecture. In practice, most training-based acceleration methods depend on training data for fine-tuning, distillation, or adaptation, while only a small subset of approaches are genuinely data-free. This dependence is particularly evident in step distillation methods, training-based efficient attention methods, most model-pruning pipelines, and QAT methods, all of which typically require task-adaptive data to absorb acceleration-induced bias. As summarized in Table~\ref{tab:training_dataset_usage_popular_methods}, customized datasets account for the largest share (40.0\%) among training-based acceleration methods. This indicates that many works rely on self-collected or internally curated video corpora tailored to their target domain, such as talking-head generation, long-video streaming, mobile deployment, or domain-specific motion patterns. While such customization can improve task fit, it also weakens reproducibility and highlights a broader privatization trend in video generation data.

According to our analysis, WebVid-10M~\cite{webvid10m}, OpenVid-1M~\cite{nan2024openvid}, UCF101~\cite{soomro2012ucf101}, Mixkit~\cite{Mixkit}, and SkyTimelapse~\cite{SkyTimelapse} emerge as the most frequently used open video datasets. WebVid-10M~\cite{webvid10m} (7.5\%) remains a prevalent large-scale training source, containing 10.7M text-video pairs and 52K total video hours, adopted by methods such as Open-Sora~\cite{https://doi.org/10.48550/arxiv.2412.20404}, VideoLCM~\cite{https://doi.org/10.48550/arxiv.2312.09109}, Latent-Shift~\cite{https://doi.org/10.48550/arxiv.2304.08477}, MagicVideo~\cite{https://doi.org/10.48550/arxiv.2211.11018}, and AnimateDiff-Lightning~\cite{https://doi.org/10.48550/arxiv.2403.12706}. However, it suffers from an abundance of low-quality, watermarked, and low-resolution videos. OpenVid-1M (6.2\%), employed by OSV, MotionStream, and SwiftVideo, serves as a high-quality alternative featuring a rigorous filtering pipeline and expressive captions. It comprises 1,019,957 video clips totaling 2,051 hours, with an average duration of 7.2 seconds and a minimum resolution of 512×512, alongside a curated 1080p subset of 433K videos (OpenVidHD-0.4M). UCF101~\cite{soomro2012ucf101} (6.2\%), utilized by PVDM~\cite{Yu_2023}, TR-DQ~\cite{https://doi.org/10.48550/arxiv.2503.06564}, and LVDM~\cite{https://doi.org/10.48550/arxiv.2211.13221}, is a classic action recognition dataset consisting of 13,320 videos across 101 categories. Although valuable for early research on video diffusion models, its narrow action-centric semantics, simplistic labels, and lower resolutions limit its applicability compared to modern T2V corpora. SkyTimelapse (3.8\%), adopted by PVDM~\cite{Yu_2023}, Ca2-VDM~\cite{https://doi.org/10.48550/arxiv.2411.16375}, and LVDM~\cite{https://doi.org/10.48550/arxiv.2211.13221}, is a domain-specific dataset reported to contain 35K uncaptioned video clips at 640×360 resolution~\cite{nan2024openvid}; while beneficial for studying long-range temporal dynamics, it lacks semantic diversity. Finally, Mixkit (3.8\%), used by Open-Sora~\cite{https://doi.org/10.48550/arxiv.2412.20404}, VORTA~\cite{https://doi.org/10.48550/arxiv.2505.18809}, and Efficient-vDiT~\cite{https://doi.org/10.48550/arxiv.2502.06155}, functions more as a curated stock video library than a standard academic benchmark. Although frequently utilized to supplement visually clean training data, the lack of standardization across papers regarding specific subsets, frame sampling policies, and effective training resolutions complicates direct performance comparisons across methods.

{\renewcommand{\refname}{Appendix References}\putbib}
\end{bibunit}

\end{document}